\documentclass[a4paper,final]{article}
\usepackage{hyperref}		% PDF hyperreferences??
\usepackage{listings}

\usepackage[utf8]{inputenc}
\usepackage[english,activeacute]{babel}
\usepackage{tabularx,wrapfig}
\usepackage{graphics}
\usepackage{graphicx}
\usepackage{epsfig}
\usepackage{amssymb,amsmath,amsthm,amsfonts}
\usepackage[all]{xy}
\usepackage{url}
\usepackage{subfigure}
\usepackage{chngpage}
\usepackage[figuresright]{rotating}
\usepackage{booktabs}
\usepackage{algorithm}
\usepackage{algorithmic}
\usepackage{comment}
\usepackage{morefloats}
\usepackage[shortlabels]{enumitem}
\newcommand{\meg}{\textbf}

\newtheorem{definition}{Definition}

\newcommand{\conecta}[3]{#1 \overset{#2}{\rightsquigarrow} #3}
\begin{document}

\title{Semantic Preserving Embeddings for Generalized Graphs}
\author{Pedro Almagro-Blanco and Fernando Sancho-Caparrini}
\maketitle

\section{Introduction}

In this work we present a new approach to the treatment of property graphs using neural encoding techniques derived from machine learning. Specifically, we will deal with the problem of embedding property graphs in vector spaces.

Throughout this paper we will use the term \emph{embedding} as an operation that allows to consider a mathematical structure, $ X $, inside another structure $ Y $, through a function, $ f: X \rightarrow Y $. We are interested on embeddings capable of capturing, within the characteristics of a vector space (distance, linearity, clustering, etc.), the interesting features of the graph.

For example, it would be interesting to get embeddings that, when projecting the nodes of the graph into points of a vector space, keep edges with the same type of the graph into the same vectors. In this way, we can interpret that the semantic associated to the relation has been captured by the embedding. Another option is to check if the embedding verifies clustering properties with respect to the types of nodes, types of edges, properties, or some of the metrics that can be measured on the graph.

Subsequently, we will use these good embedding features to try to obtain prediction / classification / discovery tools on the original graph.

This paper is structured as follows: we will start by giving some preliminary definitions necessary for the presentation of our proposal and a brief introduction to the use of artificial neural networks as \emph{encoding machines}. After this review, we will present our embedding proposal based on neural encoders, and we will verify if the topological and semantic characteristics of the original graph have been maintained in the new representation. After evaluating the properties of the new representation, it will be used to carry out machine learning and discovery tasks on real databases. Finally, we will present some conclusions and future work proposals that have arisen during the implementation of this work.

\section{Previous Definitions}

\subsection{Generalized Graphs}

The definition of \emph{Generalized Graph} that we present below unifies different graph definitions that can be found in the literature, and allows to have a general framework to support the data structures necessary for our proposal. More information about generalized graphs can be found in \cite{2017arXiv170803734A}.

\begin{definition}
A \emph{Generalized Graph} is a tuple $ G = (V, E, \mu) $ where:
\begin{itemize}
\item $ V $ and $ E $ are sets, called, respectively, \emph{set of nodes} and \emph{set of edges} of $ G $.
\item $ \mu $ is a relation (usually functional, but not necessarily) that associates each node or edge in the graph with a set of properties, that is, $ \mu: (V \cup E) \times R \rightarrow S $, where $ R $ represents the set of possible \emph{keys} for available properties, and $ S $ the set of possible \emph{values} associated.
\end{itemize} 

Usually, for each $\alpha \in R$ and $x\in V\cup E$, we write $ \alpha (x) = \mu (x,\alpha) $.

In addition, we require the existence of a special key for the edges of the graph, called \emph{incidences} and denoted by $ \gamma $, which associates to each edge of the graph a tuple, ordered or not, of vertices of the graph.
\end{definition}

Although the definition that we have presented here is more general than those from related literature, we will also call them \emph{Property Graphs}, since they are a natural extension of this type of graphs.

It should be noted that in generalized graphs, unlike traditional definitions, the elements in $ E $ are symbols representing the edges, and not pairs of elements of $ V $ and $ \gamma $ is the function that associates to each edge the set of vertices that it relates. Generalized graphs represent a simple and powerful generalization for most existing graph definitions and allow for working with broader concepts, such as hypergraphs, in a natural way.

\subsection{Encoding Neural Networks}

Prediction-related tasks represent the most common application of feedforward neural networks. In this section we present this kind of networks from a diferent perspective, using them in a way that will be (and has been) of fundamental importance for the new results that have been obtained with them.

\begin{figure}[htb]
  \centering
  \includegraphics[scale=0.25]{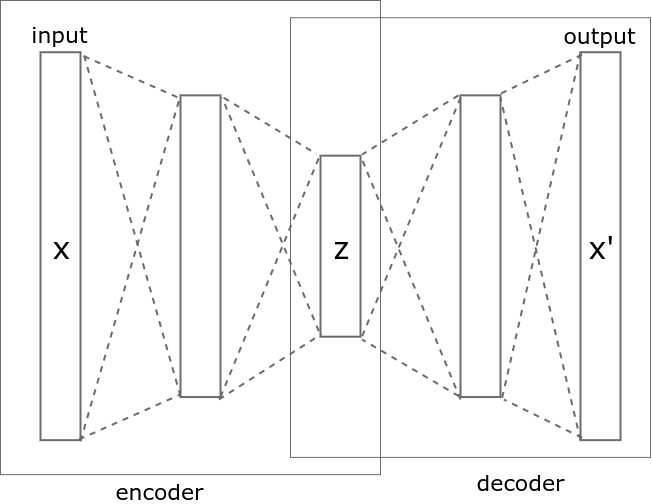}
  \caption{Neural Encoder}
  \label{encoder}
\end{figure}

A neural encoder is a neural network used for learning codings for datasets. Note that when a feedforward neural network has hidden layers, all the communication that occurs between input and output layer passes through each of the hidden layers. Thus, if we are trying to approximate a function by means of a feedforward network, after setting the parameters of the network we can interpret that a given hidden layer keeps the information required from the input data for the calculation of the function. Therefore, always from the point of view of the function that calculates the network, we can say that the section of the network from the input layer to the hidden layer encodes the input data, and the weights (and bias) of this section of the network define the encoding function between both spaces \cite {Deeplearning}. Similarly, we can understand that the part of the original network that goes from this hidden layer to the output layer defines a decoding function (see Figure \ref{encoder}).

If we leave aside the posterior layers (including the original output layer) and the associated parameters, we obtain a new neural network that produces as output a representation of the input space into a specific dimension vector space (the number of neurons in the hidden layer). We must remember that this representation is achieved as partial application of a complete feedforward network that approximates a prefixed function and, consequently, this coding is relative to this function (and, of course, to the approximation process).

An \emph{autoencoder} is a specific neural encoder where the function to learn is the identity function and, consequently, the input and output layers have the same number of neurons. As with normal encoders, when the network reaches an acceptable state (it is able to show an output similar enough to the input), the activations in the units of the hidden layers represent the encoding of the original data presented in the input layer \cite{bengio2009learning}.

If the number of units in the hidden layer differs from the number of units in the input layer (and output) we are also making a dimensional change when performing the encoding. In fact, this is one of the available methods to perform dimensionality changes by maintaining the structural characteristics (eg, proximity or similarity relations) of the training sets.

In this work we use neural encoders as a tool to perform generalized graph embeddings into vector spaces. Neural networks trained on adequate functions are used in order to verify to what extent semantic structures of the graph are conserved in the new vector space representation. 

\section{Related Works}

The application of neural encoders to texts has provided very interesting results. In 2013, T. Mikolov et al. \cite{eff} presented two new architectures, under the generic name of \emph{word2vec}, to learn vector representations of words trying to minimize computational complexity while maintaining the grammatical properties present in the texts from which they are extracted: \emph{Continuous bag-of-words} (CBOW) and \emph{Skip-gram}. In this task the \emph{context of a word} in a text is defined as the set of words that appear in adjacent positions  to it. The two architectures presented in \cite{eff} consist of feedforward artificial neural networks with $3$ layers: an input layer, a hidden layer (encoding layer) and an output layer, but they differ in the objective function they try to approximate. On the one hand, neural encoders with CBOW architecture receive the context of a given word as input and try to predict the word in its output. By contrast, encoders with Skip-gram architecture receive the word as input and try to predict the context associated with it. The main objective of the work of Mikolov et al. is to reduce the complexity in the neural model allowing the system to learn from a large volume of textual data. Until the arrival of word2vec, none of the available architectures had been able to train with more than a few million words. Through the relationship established between vocabulary words and their contexts, the model captures different types of similarity \cite{lreg}, both functional and structural, and provides an embedding of words in vector space that reflects these similarities.

In recent years different methods that try to learn vector representations of entities and relations in knowledge databases have been developed \cite{semantic, transe, nt}. All of them represent the entities as elements of a given vector space and the relationships as a combination of the representations of the entities that participate in it.

In \cite{transe}, embeddings of multi-relational data in vector space trying to verify some additional properties are proposed. Specifically, they look for a projection of nodes and types of edges, $ \pi $, in vector space with two goals:

\begin{enumerate}
\item To minimize the distance $d(\pi(s)+\pi(l),\pi(t))$ from each existing (observed) relation $ (s \xrightarrow{l} t)$ in the dataset, where $ s $ represents the source element of the relation, $ l $ represents the relationship type, and $ t $ represents the target element of the relation.

\item To maximize the distances $ d (\pi(s') + \pi(l), \pi(t)) $, $d(\pi(s)+\pi(l'),\pi(t))$ and $d(\pi(s)+\pi(l),\pi(t'))$, where $ s' $ and $ t '$ represent graph nodes, and $ l' $ represents a type of relationship of the graph, for which the relations $ s' \xrightarrow {l} t $, $ s \xrightarrow {l} t '$ and $ s \xrightarrow {l'} t $ do not exist (unobserved relationships) in the graph.
\end{enumerate}

To improve the efficiency of the algorithm, the authors randomly sample the original graph for both existing and non-existing relationships. In \cite {semantic}, and in order to achieve better results in the projection, the authors follow a similar procedure but making use of a Siamese Neural Network \footnote{A Siamese Neural Network is a type of comparative neural network composed by two networks that share weights and architecture (each one receives a data to be compared) and whose outputs are compared by a distance function.}. In \cite{embedding} some of these techniques are grouped together on the same general theoretical framework making it possible to compare the complexity of the models and the obtained results.

Despite the relationship between these works and our approach, the requirement to maximize unobserved relationships (essential for the results they obtain) works against one of the objectives we pursue, since we do not assume that the original graph has complete information and, in our context, random unobserved links creation is not a good idea. Moreover, the prediction of this kind of links is one of the task that we pursue.

In other cases, works that perform vector embedding of entities and relations in knowledge databases learns representation of the entities using the prediction of unobserved links (\emph{Link Prediction}) as objective function, conditioning the embedding with a supervised learning. In our case, the encoding is only conditioned by the similarity of the contexts in which the entities are immersed, opening the possibility of using these representations to a wider range of tasks.

DeepWalk \cite{deepwalk} is a recent methodology that uses neural encoders to obtain vector representations  nodes in uni-relational graphs using a very similar idea to word2vec. In this work the uni-relational graph is \emph{linearized} using truncated random paths, interpreting the obtained paths as sentences. Subsequently, and completely equivalent to word2vec, they use these paths to train a neural encoder with Skip-gram architecture and to obtain an embedding of the nodes of the uni-relational graph in a vector space. This method does not allow to work with \textit{large} uni-relational graphs efficiently, nor with multi-relational graphs.

LINE \cite{Tang:2015:LLI:2736277.2741093} is able to learn $d$-dimensional representations of uni-relational graph nodes through two phases: first it learns $ d / 2 $ dimensions generating random paths in Breath-First Search mode, then it learns the remaining $ d / 2 $ dimensions by sampling the nodes that are strictly at distance 2 from the embedding node.

Node2vec \cite{grover2016node2vec} groups and extends the ideas presented in DeepWalk and LINE. Specifically, the authors develop a flexible algorithm that, through two hyperparameters, allows to modify the generation of random paths that explore the environment of the nodes and that become their contexts. Based on two standard search strategies, Breath-First Sampling (BFS) and Depth-First Sampling (DFS), the two hyperparameters allow to control whether the random paths tend to a BFS or DFS strategy. In particular, they assert that a sampling guided by a BFS strategy results in embeddings reflecting structural equivalence between entities and that a sampling driven by a DFS strategy results in a embedding in which the \textit{homophilia} is reflected. In this sense, DeepWalk is a specific case of this model. Node2vec have been evaluated in \textit{Multi-Label Classification} and \textit{Link Prediction} tasks. 

In recent years, some works that use convolutional neural networks (CNN) to create vector representations of the nodes of uni-relational graphs have been published. In \cite{kipf2016semi}, the objective is to learn a function that encodes nodes by constructing descriptions of them. Outputs that represent the complete graph can also be obtained by applying some \textit{pooling} operation \cite{NIPS2015:5954}. In \cite{DBLP:journals/corr/DefferrardBV16}, the authors work with the convolution operator in the Fourier field, and generalize the convolutional networks to go from their original definition in low-dimensional regular euclidean spaces (where we work with images, videos or audio) to be able to work with high-dimensional irregular domains (multi-relational graphs obtained from social networks or biological phenomena). In \cite{schlichtkrull2017modeling} an extension of the Graph Convolutional Network \cite{kipf2016semi} called the Relational Graph Convolutional Network is presented, which allows learning through the convolution and pooling operations typical of convolutional networks on multi-relational graphs.

In \cite{DBLP:journals/tnn/ScarselliGTHM09}, the Graph Neural Network Model is defined, which converts the data graph into a recurrent neural network, and each node in a multi-layer feedforward network. The combination of these structures allows supervised learning where many of the weights of the network are shared, reducing the learning cost.

In Heterogeneous Network Embedding \cite{Chang:2015:HNE:2783258.2783296}, a framework to perform network embeddings connecting data of different types in low-dimensional spaces is presented. As the model perform unsupervised learning, the new representation is adequate for any prediction task since it has not be conditioned. Using these representations, in \cite{Jacob:2014:LLR:2556195.2556225} they face the task of automatically assign tags to nodes of different types in a heterogeneous network that has no types in the edges. The algorithm is designed to learn dependencies between node tags and to infer them exploiting the properties of the global graph and the characteristics of the neighbours of nodes. They impose two objectives: (1) grouping nodes of the same type that are connected (with less intensity as longer is the path connecting them), and (2) grouping nodes of different types if they share contexts. When working with property graphs, properties are represented as nodes.

As we have shown, there are numerous methodologies that allows to perform vector embedding of graphs, some of them are limited to working with uni-relational graphs, others are conditioned through the generation of unobserved relationships or do not capture the global semantic characteristics of a property graph. Our proposal, which we detail below, aims to obtain embeddings that are not affected by these limitations.

\section{Generalized Graph Embeddings}
\label{sec:met}

The selected architecture for our neural encoder is CBOW because, despite its simplicity and the low computational training cost, it obtains good results capturing both \textit{syntactical} and \textit{semantic} relations \cite{eff}.

In a first approximation, and in order to evaluate to what extent the semantic structure given by the edges is maintained, we will make a projection on the vector space using only the set of nodes, $V$. In this way, following the analogy offered by the word2vec algorithm, our vocabulary will be the set of nodes of the graph and their associated properties values, $S$.

A context, $ C $, associated to a node $ n \in V $ is obtained by randomly selecting a number of neighbouring nodes to $ n $ and their properties (selecting elements from $\mathcal{N}(n) \cup \mu(n,\cdot)$), regardless of the edges that connects them and of the type of property. The number of selected nodes/properties determines the \emph{selection window size}.

Following a similar methodology to those from the previous section, we will generate a training set consisting of pairs $(n, C)$, where $ n \in V $ and $ C $ is one of its associated contexts. The neural encoder is trained using this training set and then the activations of the hidden layer of the neural network are used as vector representation of each node. Algorithm \ref{alg:graph2vec}, GG2Vec, shows the followed procedure.

\begin{algorithm}
\begin{algorithmic}[1]
\STATE {$training\_set = \{\}$}
\FOR {each $0<i\leq N$}
\STATE {$n$ = randomly selected element from $V_G$}
\STATE {$C = \{\}$}
\FOR {each $0<i\leq ws$}
\STATE {$e$ = randomly selected element from $\mathcal{N}(n) \cup \mu_G(n)$}
\STATE {$C = C \cup \{e\}$}
\ENDFOR
\STATE {$training\_set = training\_set \cup \{(v,C)\}$}
\ENDFOR
\STATE {Train a $CBOW$-like architecture with $D$ neurons in hidden layer using $training\_set$}
\RETURN The resulting encoding for each element in $V$
\end{algorithmic}
\caption{GG2Vec($G$,$N$,$ws$,$D$)}\label{alg:graph2vec}
\end{algorithm}

\begin{figure}[htb]
  \centering
  \includegraphics[width=\textwidth]{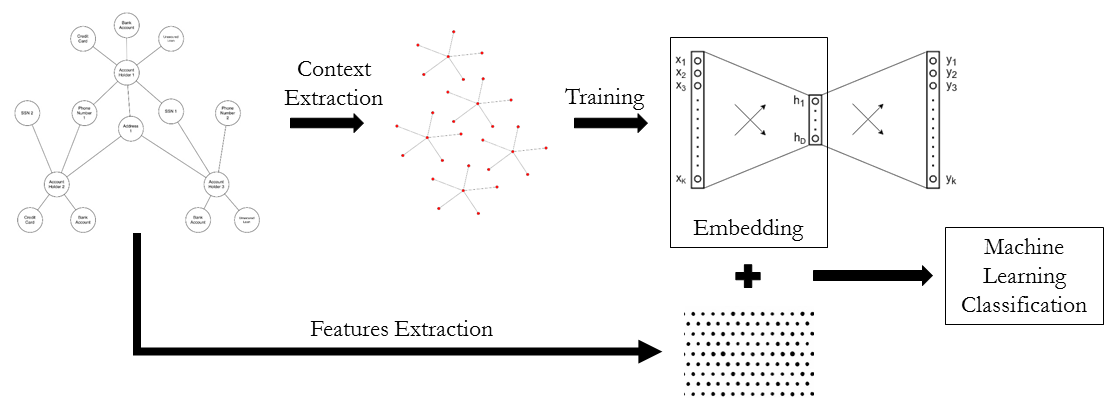}
  \caption{Proposed Methodology Representation}
  \label{diagram}
\end{figure}

We will use these vector representations trying to solve some classification and discovery tasks in the original graph. The results of these tasks will provide a measure of reliability on the achieved embedding (Fig. \ref{diagram}). 

In the embedding procedure the \emph{free parameters} of the model, which will have to be adjusted in the various experiments to analyze its effectiveness and viability, are:
\begin{itemize}
\item $D$, number of neurons in the hidden layer, determines the dimension of the vector space where we will embed the elements of the graph.
\item $N$, size of the training set, number of pairs $ (n, C) $ used to train the encoder.
\item $ws$, selection window size, number of neighbours and properties considered to construct the contexts of nodes in $ V $.
\end{itemize}

In what follows, we will note by $ \pi: V \to \mathbb{R}^D $ the embedding obtained from the trained neural encoder.

Despite generalized graphs allow hypergraph definition, next we present the embedding procedure for binary links because the databases that we use in the experiments represent binary relational data. After obtaining an embedding of the nodes of the graph, an embedding of the edges in the same vector space is induced (which will also be noted by $ \pi $) in the following way:

\begin{definition}
If $G=(V,E,\tau, \mu)$ is a binary Generalized Graph, and $\pi:V\to \mathbb{R}^D$ is a node embedding, we extend $\pi:E\to \mathbb{R}^D$ by:

$$e\in E,\ s\overset{e}{\rightarrow}t\mbox{, then } \pi(e)=\overrightarrow{\pi(s) \pi(t)}$$
\end{definition}

Since the usual operations in vector spaces are widely used in current computation units (processors and GPUs), this new representation can be used to analyze, repair and extract information from multi-relational datasets efficiently. Some tasks that can improve with this type of embeddings are:
\begin{itemize}
\item Clusters formed by nodes/edges in the new vector space can be used to induce missing properties in the elements of the graph (making use of distance, linearity or clustering relationships, for example).
\item Vector representations of the elements of a graph can help to obtain measures of similarity between them.
\item Analysis of vectors associated with the different families of relations (those sharing a common type, or verifying similar properties, for example) can help to detect missing relationships in the original dataset that in the new representation become evident. If the position of two nodes complies with the \emph{representative vector} of some type of relation, maybe those nodes should be connected by an edge with this type although that relation does not appear in the graph.
\item The representation of \emph{graph paths} in the new space can help to develop more efficient ways to perform multi-relational queries.
\end{itemize}

\section{Empirical Evaluation}

Let us perform some empirical evaluations of our method with two differentiated objectives:
\begin{enumerate}
	\item To analyse that the obtained vectorial representations maintain semantic characteristics of the original graphs.
	\item To perform classification and discovery tasks making use of the resulting embeddings.
\end{enumerate}

We will say that an embedding respects the semantics of a property graph if, from the new representation, it is possible to obtain the types associated with nodes and edges despite them have not being present during the embedding process. The type of each node or edge will be determined by a key $ \tau \in R $. In order to perform this verification, the several embeddings we will calculate will not receive information about types of nodes or edges in the original the graph (formally, they will not receive information about $ \tau $). Hence, contexts associated with nodes of the graph, which are used to create the training set, are generated by randomly selecting a number ($ws$) of neighbouring nodes and values of their different properties in $ \mu $ excluding $ \tau $.

\subsection{Implementation details and experiments}

Python has been chosen as programming language to perform the experimental evaluation \footnote{\url {https://github.com/palmagro/gg2vec}}. CBOW architecture implementation of \textit{Gensim} toolkit \footnote{\url{https://radimrehurek.com/gensim}} (version 0.12.4) has been used. In addition, Neo4j \footnote {\url{http://neo4j.com}} has been used as a persistence system.

Each embedding experiment, has been repeated 10 times, obtaining a standard deviation smaller than $ 2 \% $ in type prediction experiments. In the case of tasks related to \textit{Entity Retrieval} these deviation is bounded by $ 8 \% $ and in the case of obtaining the target nodes of a typed path is bounded by $ 8.9 \% $.

Machine learning models used to learn from the new data representations are$ k $-NN, Random Forest and Neural Networks. For the general classification tests, and unless otherwise is indicated, $ k $-NN with $ k = 3 $ has been used as base comparison model.

\subsection{Datasets}
\label{sec:ds}

The experiments were carried out in 3 different graph databases, two of them widely known by the data analysis community: WordNet and TheMovieDB. The third is a data set about the ecuadorian intangible cultural heritage.

Datasets have been partially manipulated to reduce their size and complexity. Below we give some details about each of these graphs in order to contextualize the characteristics that we will find in the results.

\begin{figure}[htb]
	\centering
	\includegraphics[scale=0.3]{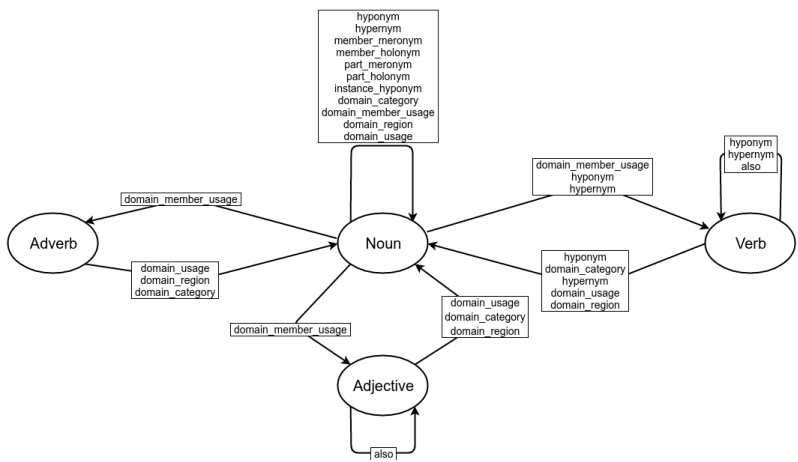}
	\caption{WordNet Schema}
	\label{wn-schema}
\end{figure}

\begin{figure}[htb]

     \begin{center}
        \subfigure[Node types distribution]{%
            \label{wordnet-nodes}
            \includegraphics[scale=0.15]{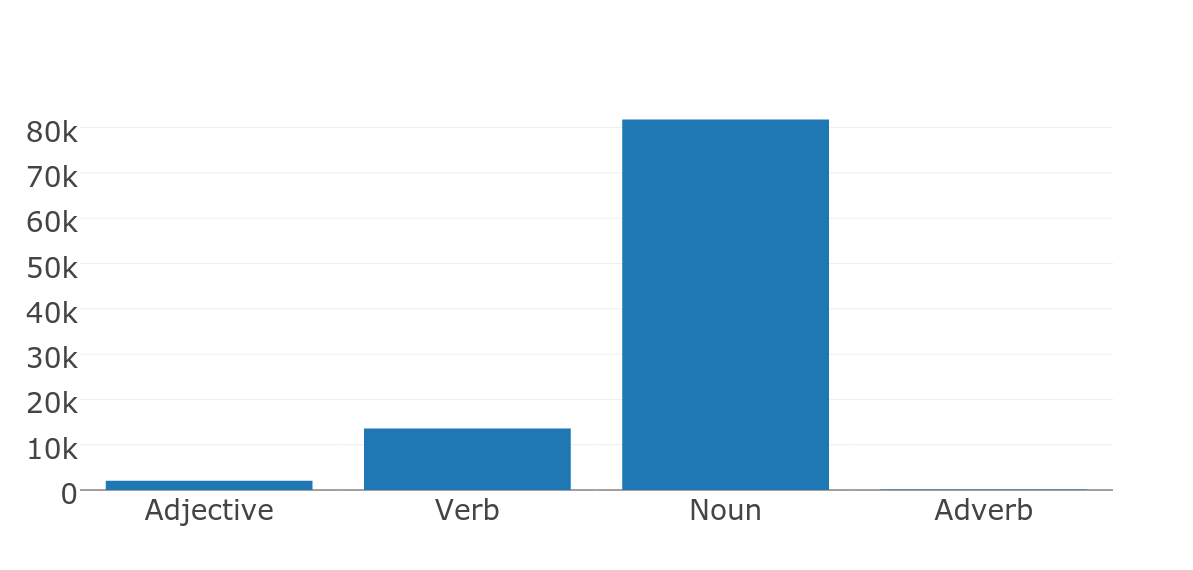}
        }%
        \subfigure[Edge types distribution]{%
            \label{wordnet-links}
            \includegraphics[scale=0.15]{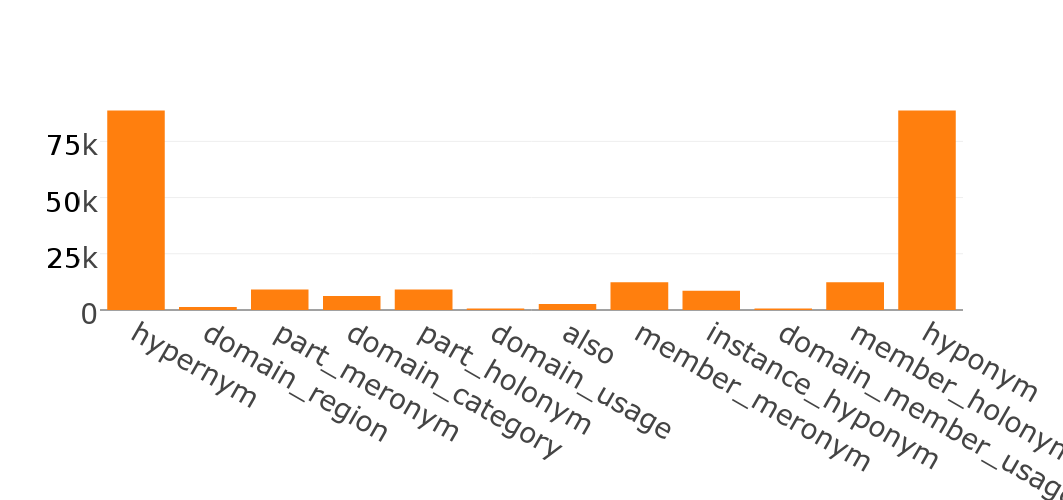}
        }%
    \end{center}
    \caption{%
       Nodes and edges distribution in WordNet
     }%
     \label{wn1}
\end{figure}

\textbf{WordNet®} \cite{wn} is a database of english nouns, verbs, adjectives and adverbs. It is one of the most important resources in the area of computational linguistics, and has been built as a combination of dictionary and thesaurus, designed to be intuitive. Each element in the database represents a list of synonymous words (which they call \textit {synset}), and the relationships that are established between the elements occur both at lexical and semantic level. In this work we have used a section of the 3.0 version, considering only entities and relations that are shown in Figure \ref{wn-schema} (in a similar way to \cite{semantic}), obtaining a graph with 97,593 nodes and 240,485 relations, with the distribution of types shown in Figure \ref{wn1}.

\begin{figure}[htb]
	\centering
	\includegraphics[scale=0.40]{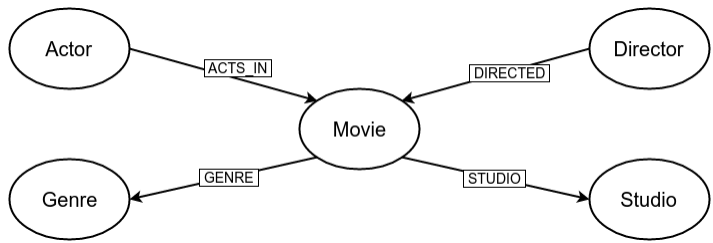}
	\caption{TMDb schema}
	\label{tmdb-schema}
\end{figure}

\begin{figure}[htb]
	\begin{center}
		\subfigure[Node types distribution]{%
			\label{cine-nodes}
			\includegraphics[scale=0.14]{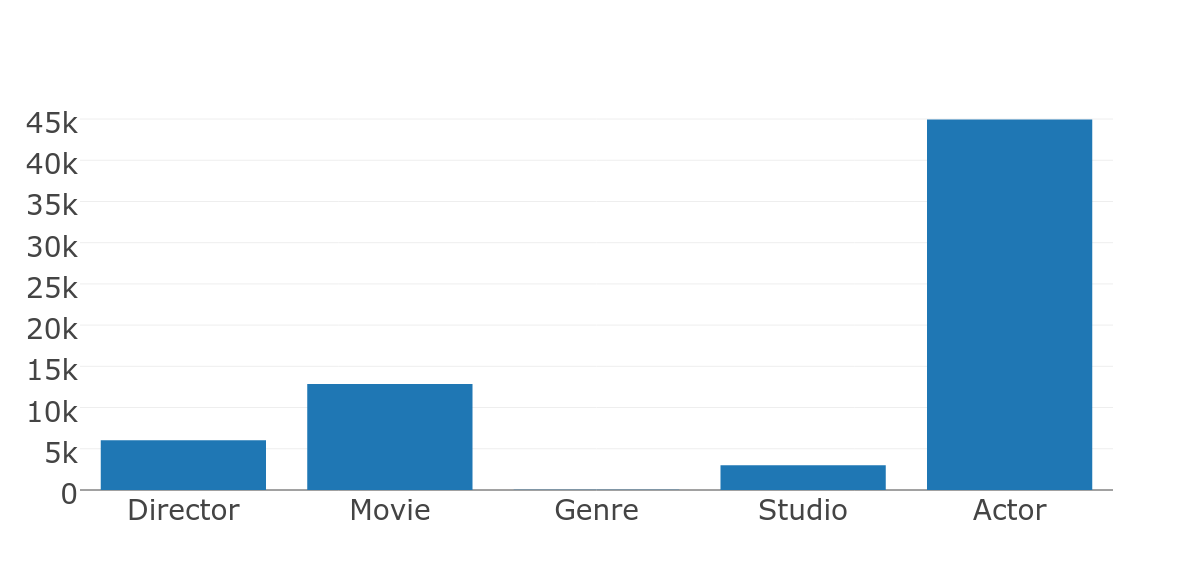}
		}%
		\subfigure[Edge types distribution]{%
			\label{cine-links}
			\includegraphics[scale=0.14]{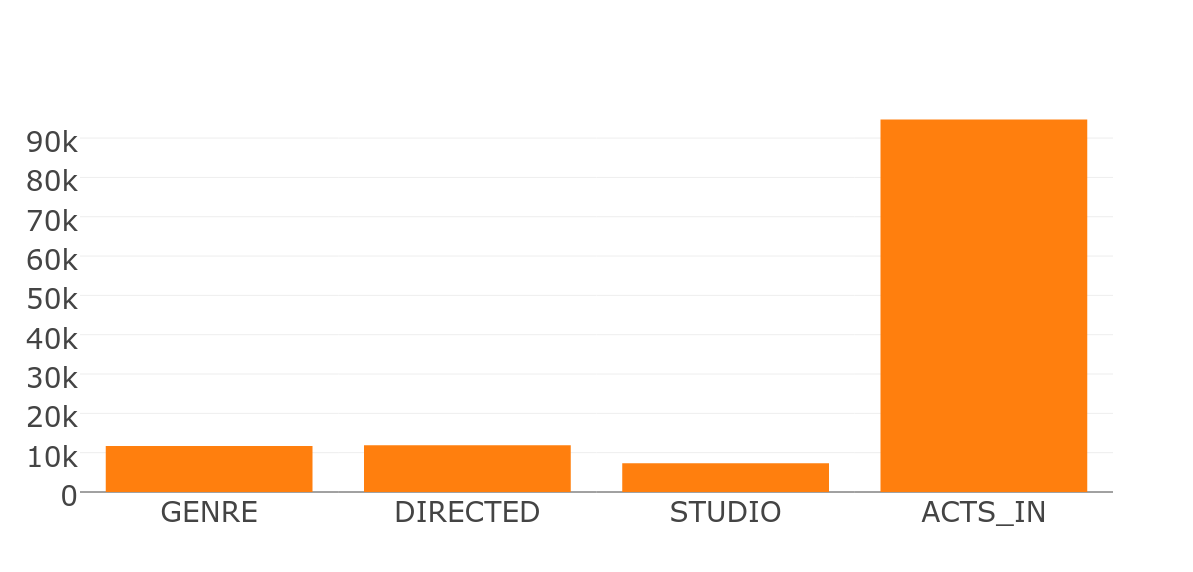}
		}%
	\end{center}
	\caption{%
		Nodes and edges distribution in TMDb.
	}%
	\label{cine2}
\end{figure}

\textbf{TheMovieDB} (TMDb) \footnote{Available at \url{https://www.themoviedb.org}} is a dataset with information about actors, movies and television content. For our experiments we have considered all the TMDb entities that are connected by relations belonging to the types \texttt{acts\textunderscore in, directed, genre} and \texttt{studio}, obtaining a graph with 66,020 nodes and 125,624 relations. Figure \ref{tmdb-schema} shows a graphical representation of the data schema, and Figure \ref{cine2} shows the distribution by types of nodes and edges of this dataset.

It should be noted that \texttt{Actor} and \texttt{Director} types are overlapping, specifically in our dataset there are 44,097 nodes associated only to \texttt{Actor} type, 5,191 nodes associated only to \texttt{Director} type, and 846 nodes with both types simultaneously (multi-type nodes).

\begin{figure}[htb]
	\centering
	\includegraphics[scale=0.30]{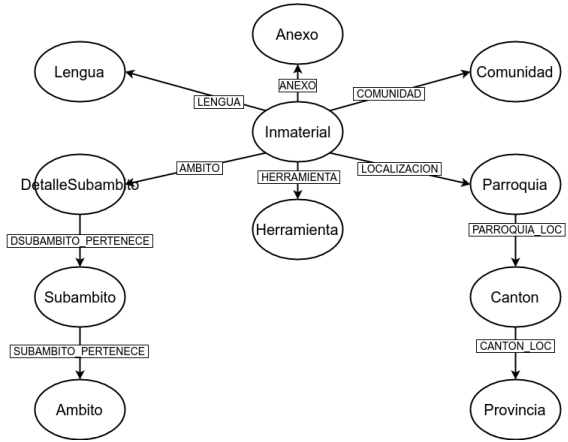}
	\caption{EICH schema}
	\label{eich-schema}
\end{figure}

\begin{figure}[htb]
	\begin{center}
		\subfigure[Node types distribution]{%
			\label{eich-nodes}
			\includegraphics[scale=0.14]{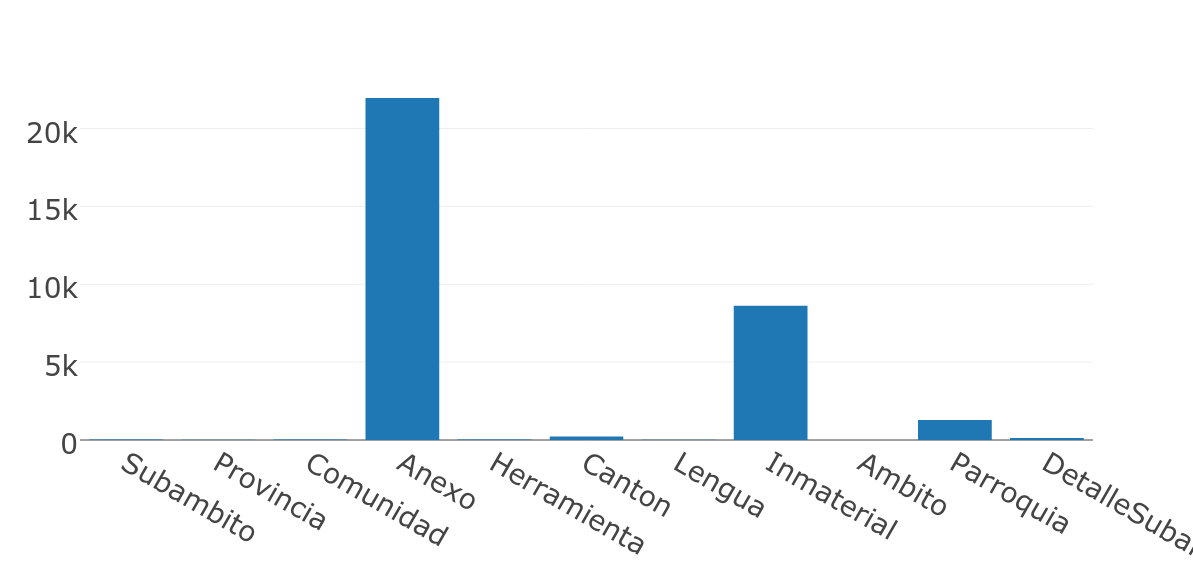}
		}%
		\subfigure[Edge types distribution]{%
			\label{eich-links}
			\includegraphics[scale=0.24]{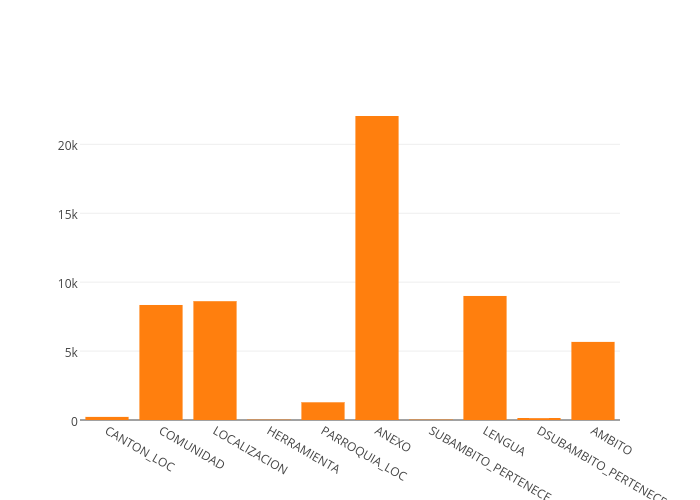}
		}%
	\end{center}
	\caption{%
		Nodes and edges distribution in EICH
	}%
	\label{eich2}
\end{figure}

\textbf{Ecuadorian Intangible Cultural Heritage} (EICH Database of Ecuador) corresponds to a section of the National Institute of Ecuadorian Cultural Heritage database \footnote{Accessible from \url{http://www.inpc.gob.ec}} which contains 38,990 nodes and 55,358 relations distributed through 11 types of nodes and 10 types of edges, with information about the intangible cultural heritage of Ecuador. This database is the most heterogeneous of the three analysed, presenting more typology in both nodes and edges, and also its elements have more properties than the other two considered datasets. Figures \ref{eich-schema} and \ref{eich2} show the schema and distribution of nodes and edges in this graph, respectively.

\begin{figure}[htb]
     \begin{center}
        \subfigure[WordNet]{%
            \label{ntp-ns}
            \includegraphics[scale=0.17]{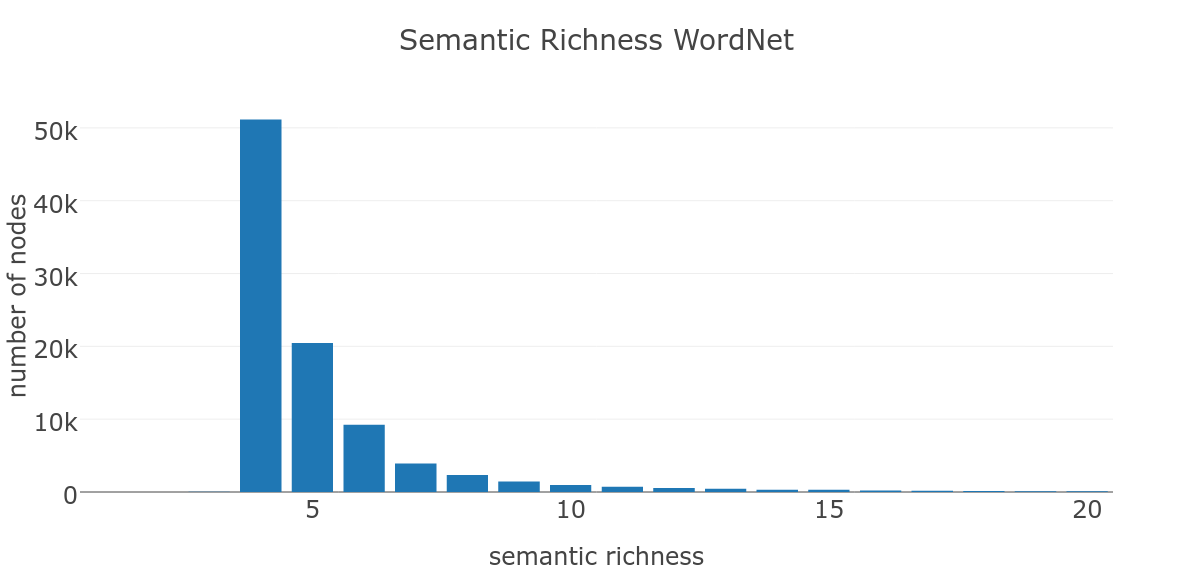}
        }%
     
        \subfigure[TMDb]{%
            \label{ntp-ndim}
            \includegraphics[scale=0.17]{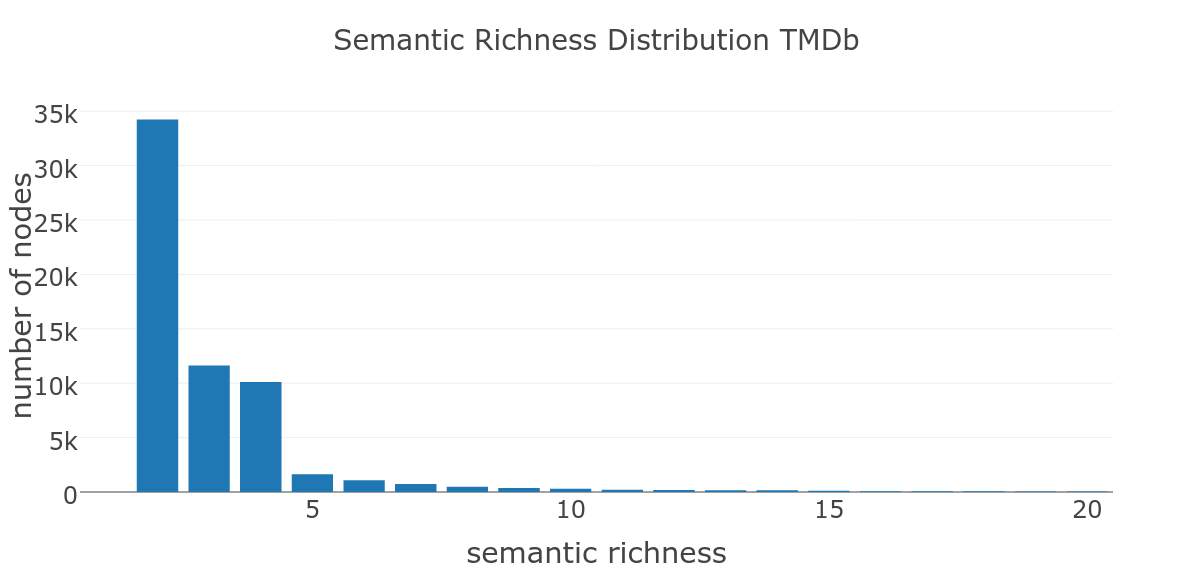}
        }\\ %  ------- End of the first row ----------------------%
     
        \subfigure[EICH]{%
            \label{ntp-l}
		    \includegraphics[scale=0.17]{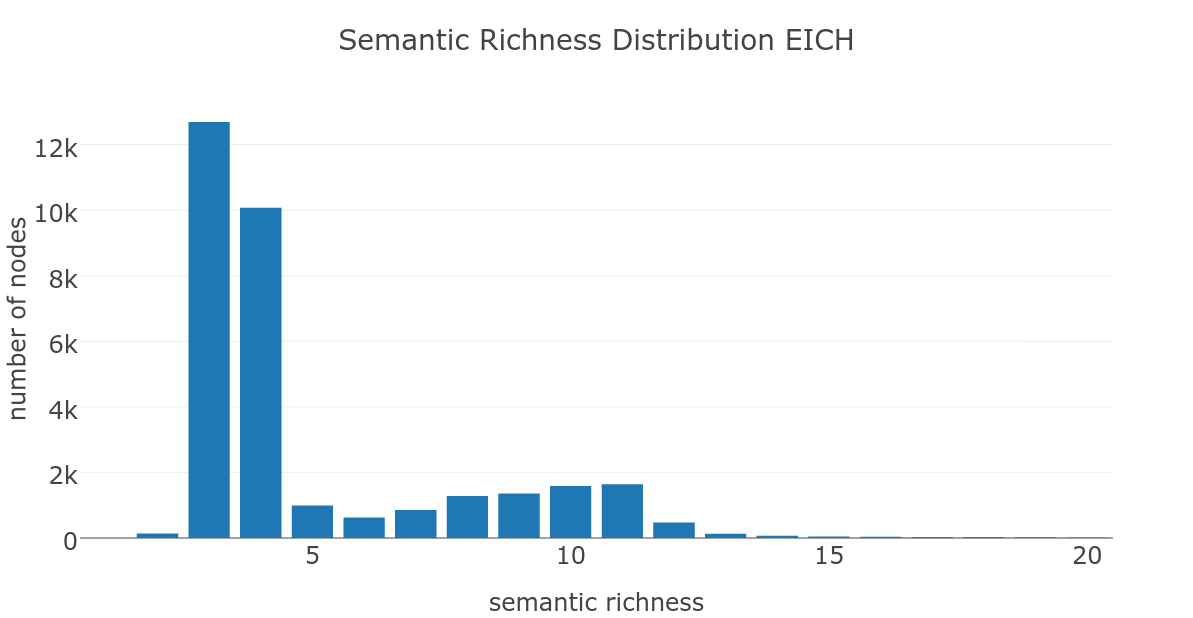}
        }%
    \end{center}
    \caption{%
       Semantic Richness (less than 20)
     }%
   \label{rsd}
\end{figure}

If we define the \emph{semantic richness} of a node as the sum of the number of relations it participates in and the number of properties it possesses, we can construct the histogram of semantic richness for each dataset (Fig. \ref{rsd}). In the case of WordNet, the average semantic richness is 5.56, in the case of TMDb it is 3.21, and in the case of EICH it is 7.86. The different behaviour of this distribution in the studied cases may be help to understand the results.

\subsection{Node Types Prediction} 

Our first experiment aims to predict $ \tau $ function. In a similar way, we could try to predict any property of $ \mu $, always being careful that it is not used during training.

A first intuition about how the achieved embeddings maintain the semantic structures can be obtained by analysing how the various types are distributed in the vector space.

\begin{figure}[htb]
	\begin{center}
		\subfigure{%
			\label{cine-nsample1}
			\includegraphics[scale=0.28]{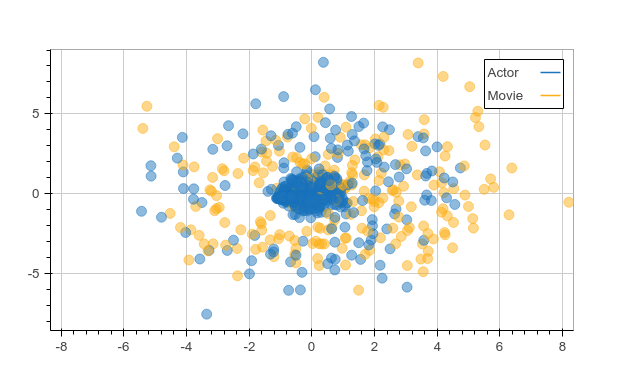}
		}%
		\subfigure{%
			\label{cine-nsample2}
			\includegraphics[scale=0.28]{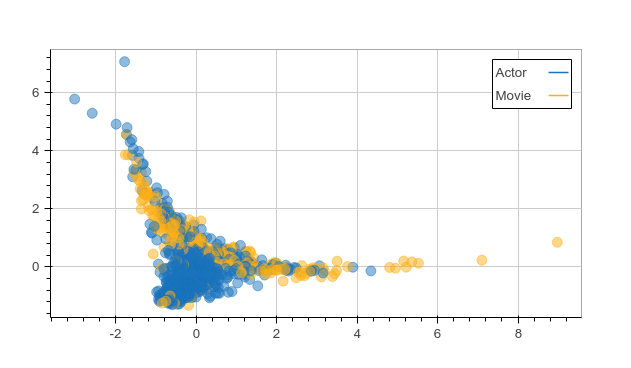}
		}%
	\end{center}
	\caption{%
		2D representations of \texttt{Movie} and \texttt{Actor} nodes in TMDb
	}%
	\label{cine-nsample}
\end{figure}

Figure \ref{cine-nsample} shows two projections of a section of TMDb graph embedding. The representation on the left shows a random selection of Movie and Actor nodes using a embedding in a space of 200 dimensions (which has been projected later on a two-dimensional space using Multi-Dimensional Scaling \cite{mds}), while the representation on the right shows the same section of the graph making use of an embedding on a 2 dimension space. Although the dimensionality reduction considered is clearly excessive (but necessary to visualize the data in these pages), both representations show that the TMDb nodes are not randomly distributed with respect to the types, which shows that, the embedding maintains information relative to the type of nodes, and we consider that the embedding process captures the semantic associated to node types.

In addition to the freedom of choice over the parameters involved in the encoding, we find some additional degrees of freedom when deciding which machine learning methiod will be used to learn from the vector representation of the nodes in a graph. As a first approximation, an exhaustive study of the free parameters of the model has been made using the classification method $k$-NN. The reason for choosing this model focuses on two fundamental aspects: it depends only on its own parameter (the value of $ k $, which is known to work relatively well for $ k = 3 $) and, in spite of its simplicity, provides robust results that serve as a comparative basis for other more sophisticated classification models.

\begin{table}[tb]
	\centering
	\caption{Embedding optimal parameters}
	\label{ntp-optimos}
	\begin{tabular}{lcccc}
		& \meg{ $N$ } & \meg{ $D$}   & \meg{ Selection Window Size}  &\meg{Prediction} \\ \hline
		\meg{ TMDb} & 400.000    & 150         & 3       &$\simeq 72\%$  \\ \hline
		\meg{ WordNet}      & 1.000.000            & 50 & 8        &$\simeq 96\%$  \\ \hline
		\meg{ EICH}      & 300.000             & 20         & 2 & $\simeq 83\%$         \\ \hline
	\end{tabular}
\end{table}

\begin{figure}[htb]
	\begin{center}
		\subfigure[Depending on the the training set size]{%
			\label{ntp-ns2}
			\includegraphics[scale=0.35]{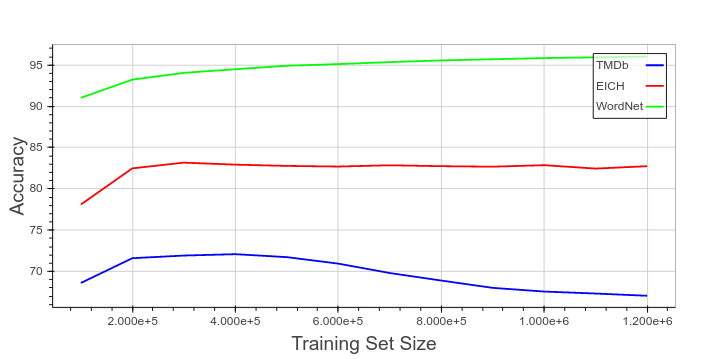}
		}\\ %
		\subfigure[Depending on the number of dimensions]{%
			\label{ntp-ndim2}
			\includegraphics[scale=0.35]{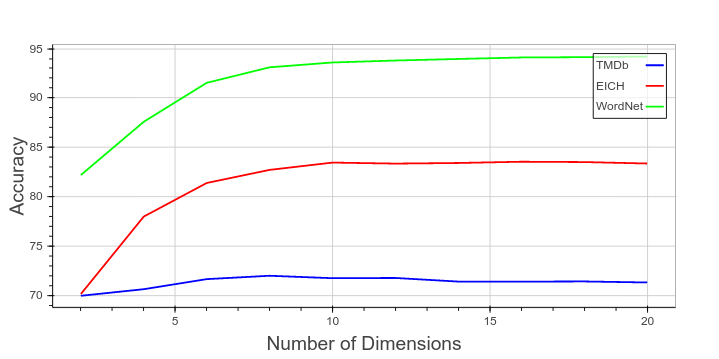}
		}\\ %  ------- End of the first row ----------------------%
		\subfigure[Depending on the \textit{selection window size}]{%
			\label{ntp-l2}
			\includegraphics[scale=0.35]{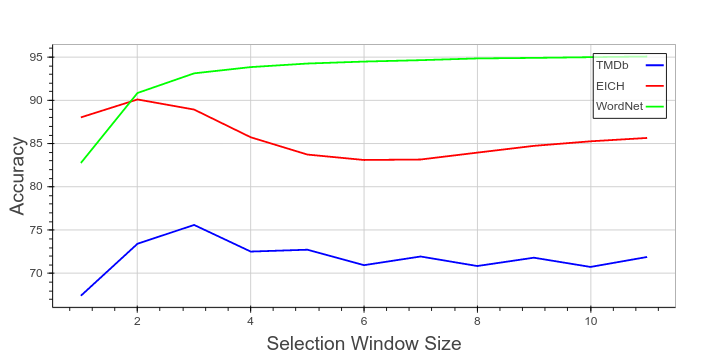}
		}%
	\end{center}
	\caption{%
		Immersion analysis (node type prediction)
	}%
	\label{ntp}
\end{figure}

Table \ref{ntp-optimos} shows optimal values of free parameters of the embedding using $ k $-NN as a later learning model (with $ k = 3 $). Figure \ref{ntp} shows the results of this analysis for the three considered graphs, where prediction rates above 70 \% are obtained for all of them.

In all cases, the optimal training set size to perform the automatic prediction of node types is proportional to the number of nodes in the graph. Both EICH and TMDb (for WordNet we do not know the optimal value, because it is increasing in the analysed range) show a reduction in the prediction rate from the optimum value, this can be due to an overfitting related to the existence of nodes of different types with the same label.

Regarding the vector space dimension, no big changes can be observed when we increase $ D $ above 10-15, a relatively low dimension, but it is almost imperceptible. We can interpret that around 10-15 dimensions are enough to capture the complexity of the analysed graphs.

The study of the influence of the selection window size ($ws$) shows that small values of this parameter are required to obtain good results in the prediction of node types. It is important to note that the best prediction is not achieved in any case with $ ws = 1 $, since this would mean that the system does not need to receive node-context pairs as elements of the training set but would suffice to show instances of the relations present in each node.

\subsubsection{Using other prediction models}

Once the parameters of the embedding shown in Table \ref{ntp-optimos} are set, we proceed to compare the predictive capacity with some other classification methods on the same task. Specifically, we will compare the results of $k$-NN with those obtained through Random Forest and Feedforward Neural Networks.

\begin{figure}[htb]
     \begin{center}
        \subfigure[k-Nearest Neighbor]{%
            \label{ntp-ns3}
            \includegraphics[scale=0.35]{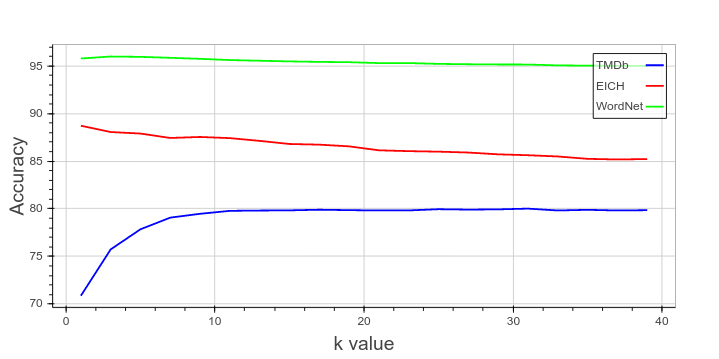}
        }\\ %
        \subfigure[Random Forest]{%
            \label{ntp-ndim3}
            \includegraphics[scale=0.35]{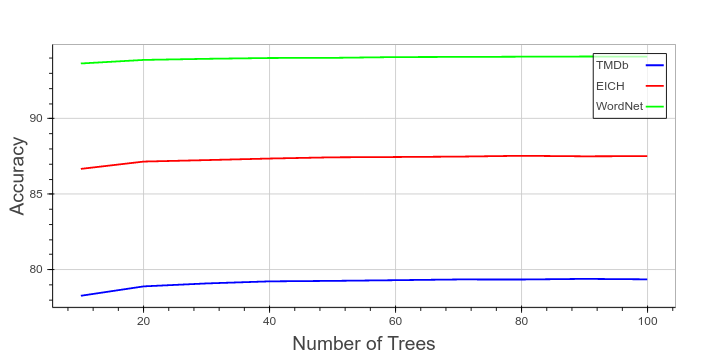}
        }\\ %  ------- End of the first row ----------------------%
        \subfigure[Feddforward Neural Network]{%
            \label{ntp-l3}
		    \includegraphics[scale=0.35]{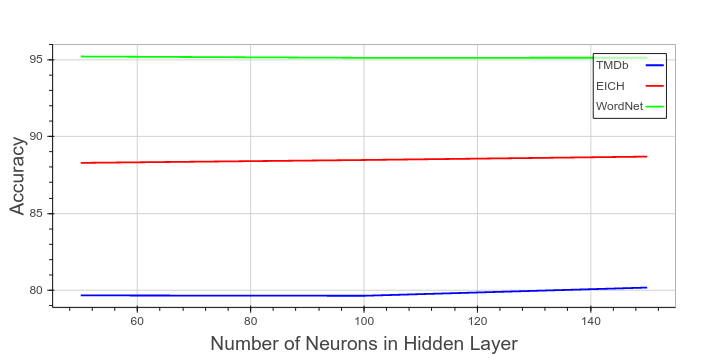}
        }%

    \end{center}
    \caption{%
       Analysis of the node types classification using different Machine Learning methods
     }%
   \label{ntp-methods}
\end{figure}

Figure \ref{ntp-methods} shows obtained results. (a) Shows the variation of the results provided by $k$-NN depending on $ k $. In (b), results obtained varying the number of trees using Random Forest as a later machine learning are shown. Finally, in (c) the results of the neural network are shown when the number of neurons in the hidden layer is modified. 

We also present averaged confusion matrices after performing 10 experiments using the optimal parameters indicated above and $k$-NN: WordNet (Table \ref{wordnet_ntcm}), TMDb (Table \ref{tmdb_ntcm}), and EICH (Table \ref{eich_ntcm}). These matrices capture the semantic similarities between node types. In EICH, for example, \texttt{Canton}, \texttt{Parroquia} and \texttt{Provincia} show overlapping behaviour because they all represent highly correlated geospatial information. In TMDb something similar occurs, \texttt{ACTOR} and \texttt{DIRECTOR} nodes appear related because, as we mentioned, there are numerous nodes in this database having both types simultaneously.

\begin{table}[tb]
\centering
\caption{Confusion Matrix: Node types prediction (WordNet)}
\label{wordnet_ntcm}
\begin{tabular}{lcccc}
                  & \meg{ adjective} & \meg{ verb}   & \meg{ noun}   & \meg{ adverb} \\ \hline
 \meg{ adjective} & \meg{ 90.01\%}    & 1.75\%         & 8.2\%          & 0.05\%         \\ \hline
 \meg{ verb}      & 0.44\%            & \meg{ 88.36\%} & 11.19\%        & 0.0\%          \\ \hline
 \meg{ noun}      & 0.2\%             & 1.56\%         & \meg{ 98.23\%} & 0.01\%         \\ \hline
 \meg{ adverb}    & 10.16\%           & 1.63\%         & 29.27\%        & \meg{ 58.94\%} \\ \hline
\end{tabular}

\end{table}

\begin{table}[tb]
\centering
\caption{Confusion Matrix: Node types prediction (TMDb)}
\label{tmdb_ntcm}
\begin{tabular}{lccccc}
                 & \meg{ Director} & \meg{ Movie} & \meg{ Genre} & \meg{ Studio} & \meg{ Actor}  \\ \hline
 \meg{ Director} & \meg{ 11.45\%}   & 9.54\%        & 0.02\%        & 1.48\%         & 77.51\%        \\ \hline
 \meg{ Movie}    & 6.51\%           & \meg{ 65.8\%} & 0.02\%        & 0.29\%         & 27.37\%        \\ \hline
 \meg{ Genre}    & 9.66\%           & 33.79\%       & \meg{ 2.07\%} & 3.45\%         & 51.03\%        \\ \hline
 \meg{ Studio}   & 9.66\%           & 8.49\%        & 0.01\%        & \meg{ 1.23\%}  & 80.61\%        \\ \hline
 \meg{ Actor}    & 5.77\%           & 7.87\%        & 0.0\%         & 0.7\%          & \meg{ 85.66\%} \\ \hline

\end{tabular}

\end{table}

\subsection{Edge Types Prediction} 

The second experiment aims to determine the goodness of the embedding in edge types prediction task.

\begin{figure}[H]
	\centering
	\includegraphics[scale=0.3]{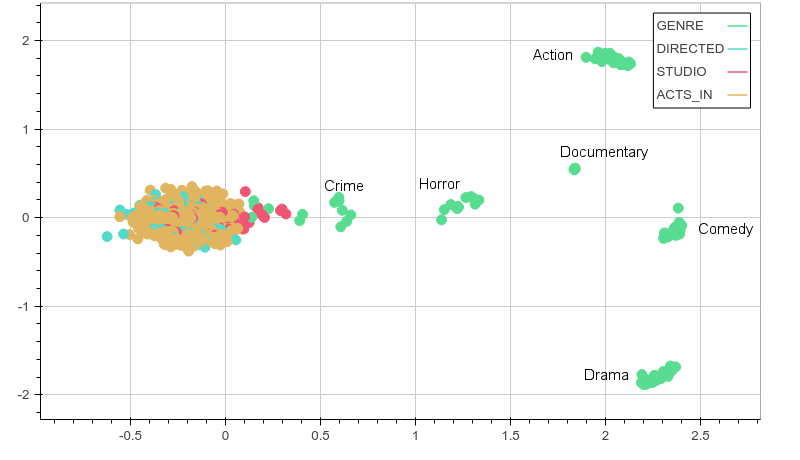}
	\caption{2D Representation of edges in TMDb}
	\label{cine-lsample}
\end{figure}

Figure \ref{cine-lsample} shows a two-dimensional projection from a randomly selected set of edges of the TMDb dataset. It can be observed that \texttt{Genre} edges do not form a single cluster, but a collection of peripheral ones that corresponding to values \texttt{Action, Comedy, Drama, Documentary, Horror} and \texttt{Crime}, showing a different semantic behaviour respect other edge types. This might indicate that \texttt{Genre} may not form a semantically unique edge type, and that its behaviour reflects some heterogeneity in design decisions when constructing the original database model. It is here where this type of analysis shows unique characteristics that can make it suitable to be used as an additional normalizer to databases covering also semantic and not only structural information. 

\begin{figure}[htb]
     \begin{center}
        \subfigure[Depending on the training set size]{%
            \label{ltp-ns}
            \includegraphics[scale=0.35]{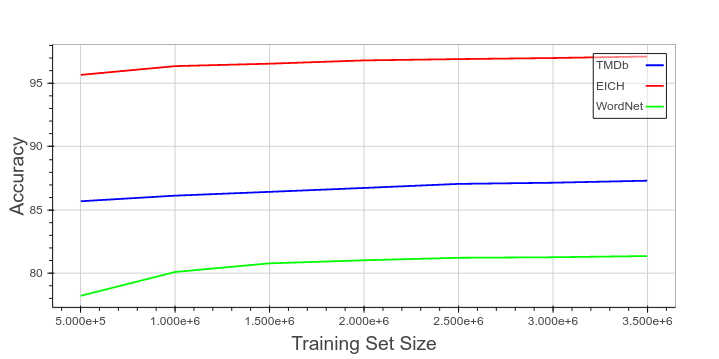}
        }\\ %
        \subfigure[Depending on the number of dimensions]{%
            \label{ltp-ndim}
            \includegraphics[scale=0.35]{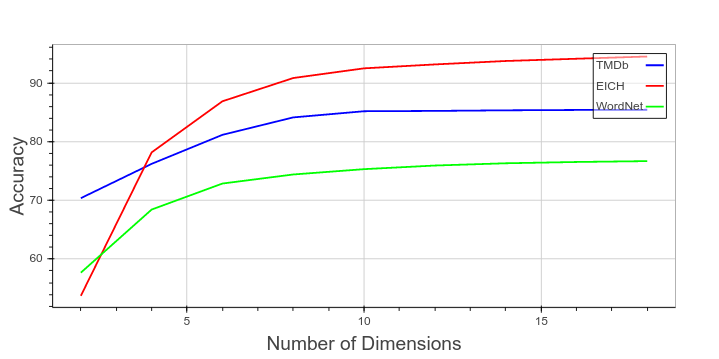}
        } \\ %  ------- End of the first row ----------------------%
        \subfigure[Depending on the \textit{selection window size}]{%
            \label{ltp-l}
		    \includegraphics[scale=0.35]{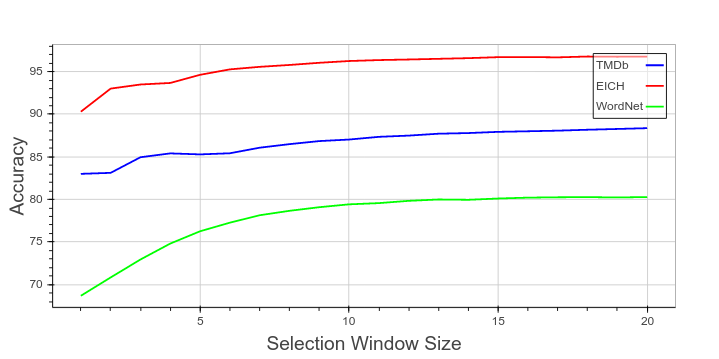}
        }%
    \end{center}
    \caption{%
       Embedding Analysis (link types prediction)
     }%
   \label{ltp}
\end{figure}

In Figure \ref{ltp}, results of edge type prediction using $k$-NN method are shown. Taking into account that the percentage of correctness is above 80 \% in all the studied datasets (even over 95 \% in some of them), we can conclude that our embedding methodology maintains the semantic properties of the edges.

In general, we can observe that the training set size necessary to obtain good results when predicting edge types is superior to that required to make a good prediction of the node types for the three analysed graphs. In addition, although WordNet is the dataset with best results when predicting node types, in the case of edge types the best results are obtained for EICH.

As in the previous case, embeddings require a relatively low dimension, $ D \simeq 15$.

The behaviour of edge types prediction tasks according to $ ws $ shows values higher than those required for nodes. In any case, it is important to note that, again, the best prediction is not achieved in any case with $ ws = 1 $.

\subsubsection{Using other prediction models}

\begin{figure}[htb]
     \begin{center}
        \subfigure[k-Nearest Neighbor]{%
            \label{ntp-ns4}
            \includegraphics[scale=0.35]{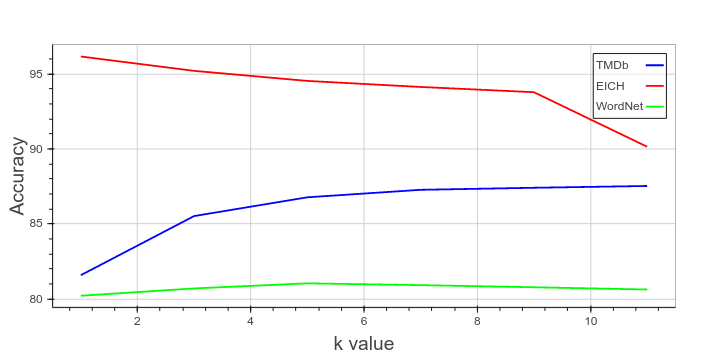}
        }\\%
        \subfigure[Random Forest]{%
            \label{ntp-ndim4}
            \includegraphics[scale=0.35]{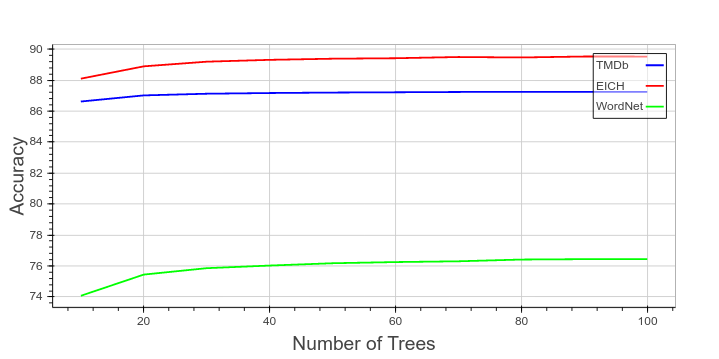}
        }\\ %  ------- End of the first row ----------------------%
        \subfigure[Feedforward Neural Network]{%
            \label{ntp-l4}
		    \includegraphics[scale=0.5]{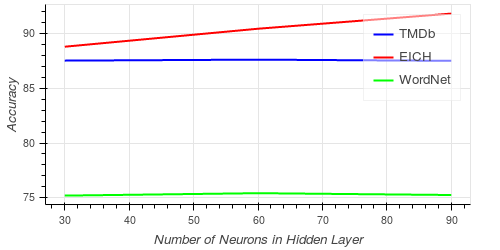}
        }%
    \end{center}
    \caption{%
Edge types classification analysis using different Machine Learning methods
     }%
   \label{ltp-methods}
\end{figure}

Following the same methodology as for node types, Figure \ref{ltp-methods} shows results obtained by the three same automatic classification methods in the edge types prediction case. These classification tasks were performed with embeddings using the parameters presented in Table \ref{ntp-optimos}.

Confusion matrices after averaging 10 experiments and using $k$ are shown in tables \ref{tmdb_ltcm}, \ref{wordnet_ltcm}, and \ref{eich_ltcm}. Results show that the embeddings capture similarities between different types of edges. In the case of EICH, edge types related to geospatial information show an overlapping behaviour with edges of type \texttt{LENGUA}, because there is a correlation between the languages and the territories where they are spoken. WordNet shows similar behaviour between \texttt{hypernym} and \texttt{hyponim} types. In the case of TMDb, as expected, \texttt{DIRECTED} edges are confused with \texttt{ACTED\textunderscore IN} edges due to the overlapping between \texttt{ACTOR} and \texttt{DIRECTOR} nodes.

\begin{table}[tb]
\centering
\caption{Confusion Matrix: Edge types prediction(TMDb)}
\label{tmdb_ltcm}
\begin{tabular}{lcccc}

                 & \meg{ GENRE}  & \meg{ DIRECTED} & \meg{ STUDIO} & \meg{ ACTS\_IN} \\ \hline
 \meg{ GENRE}    & \meg{ 99.51\%} & 0.02\%           & 0.21\%         & 0.26\%          \\ \hline
 \meg{ DIRECTED} & 0.01\%         & \meg{ 15.28\%}   & 2.04\%         & 82.67\%         \\ \hline
 \meg{ STUDIO}   & 0.13\%         & 7.22\%           & \meg{ 62.87\%} & 29.79\%         \\ \hline
 \meg{ ACTS\_IN}  & 0.01\%         & 4.75\%           & 0.94\%         & \meg{ 94.3\%}   \\ \hline

\end{tabular}

\end{table}

%\newgeometry{top=0cm,bottom=1cm}

Experimental results show that for the analysed datasets and with the proposed methodology, the obtained embedding preserves the semantics associated with edge types, and it is able to detect semantic similarities between them.

It should be noted that, since there may be edges of different types between the same pair of nodes, the results in edge types prediction may have been affected. This fact has not been taken into account in our experiments, so the results in edge types prediction could be improved. Undoubtedly, using this method we can never obtain absolute reliability about the results, but it can be taken into account for additional data normalization tasks, aproximate solution method or as a filtering method for other options.

\subsection{Entity Retrieval}

In order to show the goodness and usefulness of our embeddings in other prediction tasks, we will try to recover \textit{missing relations}. Let us consider a subset of edges, $ E' \in E $, that belong to the original graph $ G $ and consider the subgraph $ G'= (V, E \setminus E', \tau, \mu) $, that will be used to learn an embedding, $\pi$. We will try to obtain the target node associated with each edge in $ E'$ using only its source node and $ \pi $.

Formally, given an edge $e = (s, t) \in E'$ of type $ \tau(e) $, which has been eliminated before the embedding process, we will try to obtain $ t$ from $ \pi $, $ \tau (e) $ and $ s $. This task is known as \textit {Entity Retrieval} \cite{typed}.

To obtain the target node of the missing relations we will use the \textit {representative vector} of edge type which we define as:

\begin{definition}
Given a property graph $G=(V,E,\tau,\mu)$, the \emph{representative vector}, $\pi(\omega)$, associated to an edge type $\omega\in \tau(E)$,  is the mean vector of all vectors representing edges of type $\omega$. 

If we denote $E_\omega= \tau^{-1}(\{\omega\}) =\{e\in E:\ \tau(e)=\omega\}$, then:

$$\pi(\omega) = \frac{1}{\#(E_\omega)} \sum_{e\in E_\omega} \pi(e)$$
\end{definition}

A candidate of the target of a relation $ e $ from the source node can be obtained through the \textit{representative vector} by:

$$	\pi(t_e) = \pi(s) + \pi(\tau(e))$$

Then, we obtain a \textit {ranking} of the nodes of the graph by using the distances to $\pi (t_e)$.

Table \ref{ER} shows the first ten ranked results after applying this method to query nodes related by \texttt{hypernym} type to different source nodes in WordNet graph. The results are filtered to show only nodes of type \texttt{NOUN}.

\begin{table}[tb]
\setlength{\tabcolsep}{1pt}
\centering 
\caption{Entity Retrieval ranking using \texttt{hypernym} relation in WordNet}
\label{ER}
\small
\begin{tabular}{lcccc}
\hline
            & \multicolumn{1}{c}{\textit{foam}} & \multicolumn{1}{c}{\textit{spasm}} & \multicolumn{1}{c}{\textit{justification}} & \multicolumn{1}{c}{\textit{neconservatism}} \\ \hline
\textbf{1}    & hydrazine                                          & ejection & reading                             & pruritus \\ \hline
\textbf{2} & pasteboard                                          & rescue                                     & explanation                             & conservatism                                \\ \hline
\textbf{3}        & silicon dioxide                                          & putting to death                                      & analysis                           & sight                               \\ \hline
\textbf{4}        & humate                                          & sexual activity                                      & proposition                             & hawkishness                                \\ \hline
\textbf{5}        & cellulose ester                                          & behavior modification                                     & religious doctrine                             & coma                                \\ \hline
\textbf{6}        & synthetic substance                                         & disturbance & accusation                               & scientific method                               \\ \hline
\textbf{7}        & silver nitrate                                          & mastectomy                                     & assay                             & autocracy                             \\ \hline
\textbf{8}        & cast iron                                          & sales event                                     & confession                             & judiciousness                                \\ \hline
\textbf{9}        & sulfide                                          & instruction                                      & research                             & reverie                                \\ \hline
\textbf{10}        & antihemorrhagic factor & debasement                            & discouragement  & racism\\ \hline

\end{tabular}
\end{table}

It is possible that in some cases the source node of the relation has not been sampled furing the embedding process and, therefore, we can not construct its vector representation. In these cases the edge can not be evaluated and it will not be considered.

To evaluate the goodness of the embedding with respect to this task we will use the {Mean Reciprocal Rank} metric, a usual metric in \textit{Information Retrieval} \cite{typed,learning}.

\begin{definition}
The \emph{Reciprocal Rank} associated with a particular result with a list of possible answers given a query, is the inverse of the position that the correct result occupies in that list. The Mean Reciprocal Rank (MRR) is the average of the Reciprocal Ranks for a set of queries, $Q$: $$MRR = \frac{1}{|Q|} \sum_{i=1}^{|Q|} \frac{1}{\text{rank}_i}$$
where $rank_i$ is the position of the correct answer in each ranking.
\end{definition}

\begin{figure}[H]
	\centering
	\includegraphics[scale=0.35]{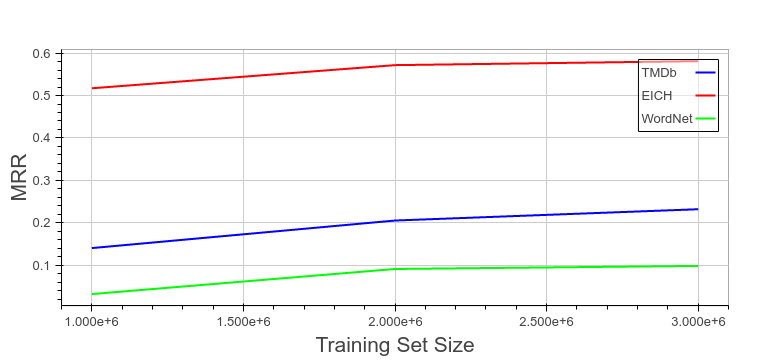}
	\caption{%
		MRR analysis for \textit{Entity Retrieval}
	}%
	\label{ntp-ns5}
\end{figure}

In Figure \ref{ntp-ns5} results obtained using MRR metric on EICH, TMDb and WordNet are shown (after removing from the ranking the nodes with wrong type). As it can be seen, our method produces excellent results that are improved when we increase the size of the training set used to perform the embedding.

\subsection{Typed Paths Prediction}
\label{cyclic}

Finally, to show the possibilities offered by a generalized graph embedding, we present a technique to obtain the target node of a given typed path, using the type of the path and the source node of the same.

A \emph{typed path} is a sequence of node and edge types that correspond to one path in the graph (in some  context, those typed paths are known as \textit{traversals}):

\begin{definition}
A \emph{typed path} of a generalized graph $G=(V,E,\tau,\mu)$ is a sequence 
$$T=t_1\overset{r_1}{\rightarrow}t_2\overset{r_2}{\rightarrow} \ldots \overset{r_q}{\rightarrow}t_{q+1}$$
where $t_i \in \tau(V)$ and $r_i \in \tau(E)$. We denote $Tp(G)$ the set of typed paths in $G$.
\end{definition}

\begin{definition}
We define an application, $Tp$, that associates to each possible typed path in $G$ the set of paths that verify it, such that if $T=t_1\overset{r_1}{\rightarrow}t_2\overset{r_2}{\rightarrow} \ldots \overset{r_q}{\rightarrow}t_{q+1}$, then for each $\rho\in Tp(T)$, $(t_1,\dots,t_{q+1})$ is the ordered sequence of node types in $\rho$, and $(r_1\dots,r_q )$ is the ordered sequence of edge types in $\rho$.
\end{definition}

Our goal is to obtain the target node of a path, just from the source node representation and the representative vector of the type that such path verifies. In this case we are not removing the paths before performing the embedding because we only pursue a new way to perform long distance queries in graph databases (not prediction tasks). It should be noted that this kind of queries in the current persistence systems are computationally expensive and that, with methods like the presented here, better performance can be reached by sacrificing optimality.

We will define the representative vector of a path as we did before for edges (indeed, edges can be seen as a particular case of types paths with length 1).

\begin{definition}
The \emph{representative vector} of a path, $\conecta{n_1}{\rho}{n_k}$, is the vector separating the representations of the source node of the path, $\pi(n_1)$,  and the target node of the path, $\pi(n_k)$. Then:
$$\pi(\rho)=\overrightarrow{\pi(n_1) \pi(n_k)} = \pi(n_k)-\pi(n_1)$$

The \emph{representative vector} of a typed path, $T$, is the average vector of all representative vectors of paths of type $T$:
$$\pi(T)=\frac{1}{|Tp(T)|}\sum_{\rho\in Tp(T)}\pi(\rho)$$
\end{definition}

Similarly as in the case of edge types, it is possible that some source nodes have not been sampled by the embedding process and thus their representation do not exist. In those cases, such path will not be evaluated.

We have performed typed path entity retrieval experiments using a similar approach as in the edge case. We have filtered target nodes depending on the type of the last element in the node sequence of the path and we have used the MRR metric again. Experiments have been performed on EICH dataset since this dataset represents a more complex schema and allows more complex typed paths queries.

\begin{figure}[H]
	\centering
	\includegraphics[scale=0.5]{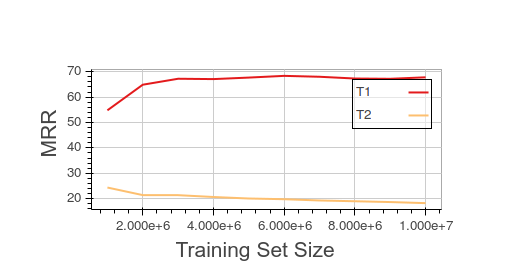}
	\caption{Typed paths MRR Analysis}
	\label{traversals_EICH}
\end{figure}

Figure \ref{traversals_EICH} shows some results obtained using the following typed paths (edge types are omitted as they can be directly inferred from the schema in Figure \ref{eich-schema}):

\begin{enumerate}

\item $T1=(Inmaterial\overset{r_1}{\rightarrow} DetSubambito\overset{r_2}{\rightarrow}Subambito\overset{r_3}{\rightarrow}Ambito)$

It is associated to paths of length 3 and contains information about \texttt{Ambito} nodes (there are 5 of this type in EICH) associated with \texttt{Inmaterial} nodes in the graph.

\item $T2=(Inmaterial\overset{r_1}{\rightarrow}Parroquia\overset{r_2}{\rightarrow}Canton\overset{r_3}{\rightarrow}Provincia)$

It is associated to paths of length 3 and contains information about \texttt{Provin\-cia} nodes (there are 24 of this type in EICH) associated with \texttt{Inmaterial} nodes in the graph.
\end{enumerate}

As Figure \ref{traversals_EICH} shows, this methodology performs well in obtaining the target node of $T1$ paths, obtaining results near to $70\%$ in MRR metric for a training set with more than 3 million pairs. In the case of $T2$, its results tends to be worst when we increase the training set size over a million pairs. The problem associated to $T_2$ is more complex, because of the number of \texttt{Provincia} nodes in the graph and the big \textit{confusion} with this kind of nodes. 

We need to perform more tests to validate it, but this application shows that it can be used to approximate long distance queries in databases, a task specially inefficient in classic persistence systems.

\section{Conclusions and Future Work}

The main goal of this work has been to allow \textit{traditional} machine learning algorithms to learn from multi-relational data through vector space embedding of generealized graphs, performing automatic \emph{feature extraction}, and maintaining semantic structures in the obtained representation. We have analysed the different options that a semantic preserving embedding for property graphs (generalized graphs) in vector spaces allows. 

If there exists an element (a subgraph) that is immersed in a graph database (a generalized graph), the task of manual feature extraction to characterize it can be very complex. The approximation presented in this work automatically obtains a vectorial representation of such relational data through a sampling of the network information. In this way, this work aliviate the process of feature extraction in multi-relational data and takes into account the global information available during the embedding process.

There are not many works that use neural encoders to perform multi-relational graph embeddings in vector spaces. Our methodology uses simple architectures to obtain vector representations of nodes and edges that maintain (to some extend) the structural and semantic characteristics of the original graph. In addition, it has been experimentally demonstrated that unobserved semantic connections in the original graph (due to lack of information or inconsistency) can be recovered and that it is possible to perform normalization tasks in graph databases or to optimize some kind of queries using our methodology.

Evaluation tests have shown that the accuracy and precision of machine learning algorithms on the new vector representation can inform about the quality of the semantic structure of the dataset. For example, confusion between some nodes / edges in classification tasks can inform about some adjustments to be done in the dataset in order to reflect the semantic characteristics correctly (and to improve classification or prediction tasks). A detailed report about how different node and edge types clusters are overlapping in the new space would be very useful when normalizing graph database data schemas. 

The training set and selection window size positively influence the applicability of the obtained embedding, but these facts must be studied in greater depth, since they can be crucial in the automatic tunning of the encoding paremeters.

We have explored how vector structures can be used to retrieve information from generalized graphs, as shown by Entity Retrieval and typed paths experiments. It is likely that looking for complex structures in the projected space will be simpler than in the original one. In fact, the use of a second layer of learning models after neural encoding can improve the results of various tasks related to information retrieval tasks in semantic graphs. Results show that it is worth considering this line of research. Although not enough experiments have been carried out on typed path queries, the results obtained show that executing query time can be reduced dramatically by sacrificing optimality. This type of queries are very expensive in classical databases, and although graph databases help to reduce their computational cost they still present hard efficiency problems when the path to query is longer than 3.

Compared to other approaches in the same direction, this paper presents the novelty of working with more general semantic contexts, and not only with random paths, which suppose a linearisation of the original graph structure. But these are not the only options to carry out generalized graph encodings through neural networks, as future work, we could achieve vectorial encodings by using neural autoencoders, so that the neural encoder will learn the identity function for the elements of the graph, avoiding the bias imposed by the function that relates the elements to their context.

It should be noted that during the revision of this document new tools optimizing \textit{word2vec} related learning procedures have been published \cite{bojanowski2016enriching}. In spite of the probable improvement that these tools would suppose in our methodology, we have decided not to take them into account since they do not modify the essence of our proposal, although it would alleviate the computational costs associated to the performed experiments.

Efficiency improvements in long-distance queries shown in section \ref{cyclic} deserve to be evaluated in greater depth and compared with other similar methods. Some results related to the semantic analysis of generalized graphs have not been presented in this paper although they are expected to be in later works. Options such as sampling the context of edges, performing their embedding and infer from it an embedding for nodes have not been taken into account and can offer interesting results. Even embeddings of the both sets simultaneously should be considered.

Also as a future line of work, to analyse the characteristics of the embeddings should be considered. A first step is about how to construct the training set to be consumed by the neural encoder. In a first approximation the construction of the training set has been totally random, ie, all nodes have the same probability of being sampled, as well as all their properties and neighbours. This may not be the most appropriate way depending on the type of activity to be performed with the obtained embedding. For example, it may be beneficial that nodes with a greater semantic richness are more likely to be in the training set, this option may contribute to explore regions initially less likely to be considered.

It should also be noted that the possibility of working with continuous properties in nodes and edges is open, and should be considered to expand the capacity of our methodology. There are direct mechanisms to include the presence of continuous properties, it remains as work to begin by testing them and to measure later to what extent other approaches can be taken into account. 

Similarly, it would be interesting to think about neural encoders that make use of recurrent neural networks to analyse the behaviour of dynamic relational information, an area practically unexplored today.

\section*{Acknowledgement}

We thank the "Instituto Nacional de Patrimonio Cultural" of Ecuador for the information related to the Intangible Cultural Heritage of Ecuador. This work has been partially supported by TIC-6064 Excellence Project of the Junta de Andalucía and TIN2013-41086-P from Spanish Ministry of Economy and Competitiveness (cofinanced with FEDER funds) and by Research and Graduate Studies Head Department of Central University of Ecuador.

\bibliographystyle{plain}	% or "siam", or "alpha", etc.
\bibliography{biblio}

\begin{sidewaystable}
	\setlength{\tabcolsep}{1pt}
	
	\tiny
	\caption{Confusion matrix: Link types prediction (WordNet)}
	\label{wordnet_ltcm}
	\begin{tabular}{lcccccccccccc}
		
		& \meg{ hyper} & \meg{ dom\_reg} & \meg{ part\_mero} & \meg{ dom\_cat} & \meg{ part\_holo} & \meg{ dom\_usage} & \meg{ also}  & \meg{ memb\_mero} & \meg{ inst\_hypo} & \meg{ dom\_memb\_usage} & \meg{ memb\_holo} & \meg{ hypo} \\ \hline
		\meg{ hyper}            & \meg{ 83.67\%}   & 0.01\%                & 1.14\%               & 0.76\%                  & 0.43\%               & 0.02\%               & 0.56\%        & 1.96\%                 & 0.13\%                   & 0.02\%                      & 2.77\%                 & 8.52\%          \\ \hline
		\meg{ dom\_reg}       & 2.08\%           & \meg{ 65.58\%}        & 25.0\%               & 0.33\%                  & 0.93\%               & 0.0\%                & 0.21\%        & 3.27\%                 & 0.24\%                   & 0.0\%                       & 1.12\%                 & 1.25\%          \\ \hline
		\meg{ part\_mero}        & 29.21\%          & 0.74\%                & \meg{ 44.47\%}       & 1.1\%                   & 3.54\%               & 0.01\%               & 0.79\%        & 4.12\%                 & 0.34\%                   & 0.0\%                       & 4.47\%                 & 11.2\%          \\ \hline
		\meg{ dom\_cat}     & 14.88\%          & 0.02\%                & 1.19\%               & \meg{ 78.91\%}          & 0.11\%               & 0.01\%               & 0.44\%        & 0.18\%                 & 0.04\%                   & 0.0\%                       & 0.25\%                 & 3.97\%          \\ \hline
		\meg{ part\_holo}        & 15.53\%          & 0.09\%                & 3.28\%               & 0.34\%                  & \meg{ 45.36\%}       & 0.01\%               & 0.78\%        & 2.25\%                 & 1.24\%                   & 0.0\%                       & 6.67\%                 & 24.45\%         \\ \hline
		\meg{ dom\_usage}        & 4.28\%           & 0.0\%                 & 0.06\%               & 0.39\%                  & 0.03\%               & \meg{ 93.41\%}       & 0.99\%        & 0.06\%                 & 0.0\%                    & 0.0\%                       & 0.11\%                 & 0.68\%          \\ \hline
		\meg{ also}                & 8.0\%            & 0.0\%                 & 0.38\%               & 0.43\%                  & 0.19\%               & 0.03\%               & \meg{ 78.0\%} & 1.92\%                 & 0.07\%                   & 0.01\%                      & 4.64\%                 & 6.35\%          \\ \hline
		\meg{ memb\_mero}      & 11.7\%           & 0.18\%                & 1.76\%               & 0.2\%                   & 0.77\%               & 0.0\%                & 0.93\%        & \meg{ 50.72\%}         & 0.21\%                   & 0.0\%                       & 24.67\%                & 8.87\%          \\ \hline
		\meg{ inst\_hypo}    & 2.47\%           & 0.06\%                & 0.87\%               & 0.05\%                  & 1.81\%               & 0.0\%                & 0.2\%         & 0.53\%                 & \meg{ 80.42\%}           & 0.0\%                       & 1.56\%                 & 12.02\%         \\ \hline
		\meg{ dom\_memb\_usage} & 1.1\%            & 0.0\%                 & 0.03\%               & 0.03\%                  & 0.14\%               & 0.0\%                & 1.13\%        & 0.09\%                 & 0.06\%                   & \meg{ 93.62\%}              & 0.09\%                 & 3.73\%          \\ \hline
		\meg{ memb\_holo}      & 11.24\%          & 0.05\%                & 0.67\%               & 0.09\%                  & 1.83\%               & 0.0\%                & 0.91\%        & 14.01\%                & 0.21\%                   & 0.0\%                       & \meg{ 62.37\%}         & 8.61\%          \\ \hline
		\meg{ hypo}             & 10.52\%          & 0.01\%                & 0.42\%               & 0.31\%                  & 1.18\%               & 0.02\%               & 0.57\%        & 1.23\%                 & 1.18\%                   & 0.01\%                      & 3.67\%                 & \meg{ 80.88\%}  \\ \hline
		
	\end{tabular}
	
	\tiny
	\caption{Confusion matrix: Node types prediction (EICH)}
	\label{eich_ntcm}
	\begin{tabular}{lccccccccccc}
		
		& \meg{ Subambito} & \meg{ Provincia} & \meg{ Comunidad} & \meg{ Anexos} & \meg{ Herramienta} & \meg{ Canton} & \meg{ Lengua} & \meg{ Inmaterial} & \meg{ Ambito} & \meg{ Parroquia} & \meg{ DetalleSubambito} \\ \hline
		\meg{ Subambito}        & \meg{ 14.53\%}    & 0.0\%             & 0.0\%             & 2.56\%         & 0.0\%               & 0.0\%          & 0.0\%          & 47.86\%            & 0.0\%          & 0.85\%            & 34.19\%                  \\ \hline
		\meg{ Provincia}        & 0.0\%             & \meg{ 7.14\%}     & 2.04\%            & 0.0\%          & 0.0\%               & 69.39\%        & 5.1\%          & 1.02\%             & 0.0\%          & 15.31\%           & 0.0\%                    \\ \hline
		\meg{ Comunidad}        & 0.0\%             & 0.0\%             & \meg{ 0.0\%}      & 7.91\%         & 0.0\%               & 1.44\%         & 0.0\%          & 25.18\%            & 0.0\%          & 65.47\%           & 0.0\%                    \\ \hline
		\meg{ Anexos}           & 0.0\%             & 0.0\%             & 0.0\%             & \meg{ 81.16\%} & 0.0\%               & 0.0\%          & 0.0\%          & 18.63\%            & 0.0\%          & 0.21\%            & 0.0\%                    \\ \hline
		\meg{ Herramienta}      & 0.0\%             & 0.0\%             & 0.0\%             & 0.68\%         & \meg{ 36.99\%}      & 0.0\%          & 0.0\%          & 62.33\%            & 0.0\%          & 0.0\%             & 0.0\%                    \\ \hline
		\meg{ Canton}           & 0.0\%             & 3.74\%            & 0.1\%             & 5.27\%         & 0.0\%               & \meg{ 12.18\%} & 0.0\%          & 24.26\%            & 0.0\%          & 54.27\%           & 0.19\%                    \\ \hline
		\meg{ Lengua}           & 0.0\%             & 0.0\%             & 0.0\%             & 6.25\%         & 0.0\%               & 0.0\%          & \meg{ 0.0\%}   & 21.25\%            & 0.0\%          & 72.5\%            & 0.0\%                    \\ \hline
		\meg{ Inmaterial}       & 0.01\%            & 0.0\%             & 0.0\%             & 9.44\%         & 0.19\%              & 0.0\%          & 0.01\%         & \meg{ 89.77\%}     & 0.0\%          & 0.56\%            & 0.01\%                   \\ \hline
		\meg{ Ambito}           & 44.0\%            & 0.0\%             & 0.0\%             & 0.0\%          & 0.0\%               & 0.0\%          & 0.0\%          & 28.0\%             & \meg{ 0.0\%}   & 0.0\%             & 28.0\%                   \\ \hline
		\meg{ Parroquia}        & 0.02\%            & 0.42\%            & 0.08\%            & 2.34\%         & 0.02\%              & 2.11\%         & 0.35\%         & 29.67\%            & 0.0\%          & \meg{ 64.82\%}    & 0.18\%                   \\ \hline
		\meg{ DetalleSubambito} & 1.63\%            & 0.0\%             & 0.0\%             & 5.42\%         & 0.0\%               & 0.18\%         & 0.0\%          & 49.37\%            & 0.0\%          & 4.52\%            & \meg{ 38.88\%}           \\ \hline
		
	\end{tabular}

	\tiny
	\caption{Confusion matrix: Link types prediction (EICH)}
	\label{eich_ltcm}
	\begin{tabular}{lcccccccccc}
		
		& \meg{ CANTON\_L} & \meg{ COM} & \meg{ LOC} & \meg{ HERRAM} & \meg{ PARROQ\_L} & \meg{ ANEXO}  & \meg{ SUBAMBITO\_P} & \meg{ LENGUA} & \meg{ AMBITO} & \meg{ DSUBAMBITO\_P} \\ \hline
		\meg{ CANTON\_L}           & \meg{ 25.05\%}     & 1.94\%            & 12.62\%              & 0.0\%               & 26.8\%                & 6.99\%         & 0.0\%                       & 22.91\%        & 3.69\%         & 0.0\%                        \\ \hline
		\meg{ COM}            & 0.0\%              & \meg{ 97.92\%}    & 0.18\%               & 0.0\%               & 0.09\%                & 0.37\%         & 0.0\%                       & 1.4\%          & 0.04\%         & 0.0\%                        \\ \hline
		\meg{ LOC}         & 0.0\%              & 0.12\%            & \meg{ 96.77\%}       & 0.04\%              & 1.08\%                & 1.51\%         & 0.0\%                       & 0.33\%         & 0.15\%         & 0.0\%                        \\ \hline
		\meg{ HERRAM}          & 0.0\%              & 0.0\%             & 1.49\%               & \meg{ 44.03\%}      & 2.24\%                & 48.51\%        & 0.0\%                       & 3.73\%         & 0.0\%          & 0.0\%                        \\ \hline
		\meg{ PARROQ\_L}        & 0.89\%             & 0.99\%            & 13.95\%              & 0.02\%              & \meg{ 59.11\%}        & 3.34\%         & 0.0\%                       & 19.2\%         & 2.45\%         & 0.05\%                       \\ \hline
		\meg{ ANEXO}                & 0.14\%             & 0.14\%            & 0.49\%               & 0.02\%              & 2.46\%                & \meg{ 95.87\%} & 0.0\%                       & 0.73\%         & 0.11\%         & 0.05\%                       \\ \hline
		\meg{ SUBAMBITO\_P}  & 0.81\%             & 0.0\%             & 10.57\%              & 0.0\%               & 8.13\%                & 9.76\%         & \meg{ 30.89\%}              & 14.63\%        & 5.69\%         & 19.51\%                      \\ \hline
		\meg{ LENGUA}               & 0.0\%              & 1.06\%            & 0.09\%               & 0.0\%               & 0.01\%                & 0.39\%         & 0.0\%                       & \meg{ 98.34\%} & 0.1\%          & 0.0\%                        \\ \hline
		\meg{ AMBITO}               & 0.01\%             & 0.04\%            & 0.28\%               & 0.01\%              & 0.04\%                & 1.9\%          & 0.0\%                       & 2.38\%         & \meg{ 95.33\%} & 0.02\%                       \\ \hline
		\meg{ DSUBAMBITO\_P} & 0.18\%             & 0.18\%            & 1.96\%               & 0.0\%               & 2.67\%                & 3.21\%         & 0.89\%                      & 22.46\%        & 7.66\%         & \meg{ 60.78\%}               \\ \hline
		
	\end{tabular}

\end{sidewaystable}

\end{document}

\title{Inmersión Semántica de Grafos Generalizados en Espacios Vectoriales}
\author{Pedro Almagro-Blanco, Fernando Sancho-Caparrini}
\maketitle

\section{Introducción}

En este trabajo presentamos una nueva aproximación al tratamiento de Grafos con Propiedades, en este caso haciendo uso de técnicas derivadas del aprendizaje automático por medio de redes neuronales codificadoras. Concretamente, trataremos aquí el problema de la inmersión de Grafos con Propiedades en espacios vectoriales.

A lo largo de este artículo utilizaremos el término inmersión como una operación que permite considerar una estructura matemática, $X$, dentro de otra estructura $Y$, y que vendrá dada por una función, $f : X \rightarrow Y$. En el caso que nos ocupa no nos interesa cualquier inmersión, ya que la mayoría de ellas no mantendrán las estructuras topológicas (dadas por las relaciones existentes en el grafo) y semánticas (dadas por los tipos asociados a nodos y aristas) que confieren su interés a los grafos que estudiamos, sino solo aquellas que sean capaces de reflejar, dentro de las características propias de un espacio vectorial (relaciones de distancia, linealidad, clusterización, etc.), las características interesantes del grafo.

Por ejemplo, sería interesante conseguir inmersiones que, al proyectar los nodos del grafo en puntos de un espacio vectorial, hagan que las relaciones existentes queden proyectadas automáticamente en vectores que mantengan los tipos. De esta forma, interpretamos que la carga semántica asociada a la relación ha sido capturada por la inmersión del grafo en el espacio vectorial. Otra opción es comprobar si la inmersión verifica propiedades de clusterización respecto a los tipos de nodos, sus propiedades, o algunas de las métricas que podemos definir sobre el grafo.

Posteriormente, haremos uso de estas buenas propiedades de inmersión para tratar de obtener herramientas de predicción / clasificación / descubrimiento sobre el grafo original.

Este artículo se estructura como sigue: comenzaremos dando algunas definiciones previas necesarias para la presentación de nuestra propuesta y una breve introducción al uso de redes neuronales artificiales como \emph{máquinas codificadoras}. Tras hacer un repaso por algunos trabajos relacionados que pueden ser vistos como precedentes del que presentamos aquí, pasaremos a presentar nuestra propuesta de inmersión por medio de un codificador neuronal que hace uso de un conjunto de entrenamiento extraído de un grafo con propiedades, y verificaremos hasta qué punto se mantienen las características topológicas y semánticas del grafo original en la nueva representación. Tras evaluar las propiedades que posee la nueva representación, ésta será usada para llevar a cabo tareas de aprendizaje automático y descubrimiento sobre bases de datos reales. Por último, se presentarán las conclusiones obtenidas y las propuestas de trabajo futuro que han surgido durante la ejecución de este trabajo.

\section{Definiciones Previas}

\subsection{Grafos Generalizados}

La definición de \emph{Grafo Generalizado} que presentamos a continuación unifica diferentes variantes de grafo que se pueden encontrar en la literatura, y nos permite disponer de un marco lo suficiente general para dar soporte a las estructuras de datos necesarias para nuestra propuesta. Se puede encontrar más información acerca de grafos generalizados en \cite{2017arXiv170803734A}.

\begin{definition}
Un \emph{Grafo Generalizado} es una tupla $G = (V,E,\mu)$ donde:
\begin{itemize}
	\item $V$ y $E$ son conjuntos, que llamaremos, respectivamente, \emph{conjunto de nodos} y \emph{conjunto de aristas} de $G$.
	\item $\mu$ es una relación (habitualmente la consideraremos funcional, pero no es necesario) que asocia a cada nodo o arista en el grafo su conjunto de propiedades, es decir, $\mu : ( V \cup E) \times R \rightarrow S$, donde $R$ representa el conjunto de posibles \emph{claves} para dichas propiedades, y $S$ el conjunto de posibles \emph{valores} asociados a las mismas.
\end{itemize} 

Habitualmente, para cada $\alpha \in R$ y $x\in V\cup E$, escribiremos $\alpha(x)=\mu(x,\alpha)$.

Además, exigiremos la existencia de una clave destacada para las aristas del grafo, que llamaremos \emph{incidencias} y denotaremos por $\gamma$, que asocia a cada arista del grafo una tupla, ordenada o no, de vértices del grafo.
\end{definition}

Aunque la definición que hemos presentado aquí es más general que las que se pueden encontrar en la literatura relacionada, también los denominaremos \emph{Grafos con Propiedades}, como hacen muchos de esos trabajos, ya que suponen una extensión natural de este tipo de grafos. 

Cabe indicar que en los grafos generalizados que acabamos de mostrar, y a diferencia de las definiciones tradicionales, los elementos en $E$ son símbolos que representan a las aristas, y no pares de elementos de $V$, y es $\gamma$ la función que asocia a cada arista el conjunto de vértices que relaciona, pudiendo trabajar con conceptos más amplios, como el de hipergafo, de forma natural. 

\subsection{Redes Neuronales Codificadoras}

El uso más habitual de las redes neuronales feedforward ha sido como máquinas de cálculo, pero en esta sección presentamos un uso que será (y ha sido) de fundamental importancia para los nuevos resultados que se han obtenido con ellas.

\begin{figure}[htb]
  \centering
  \includegraphics[scale=0.25]{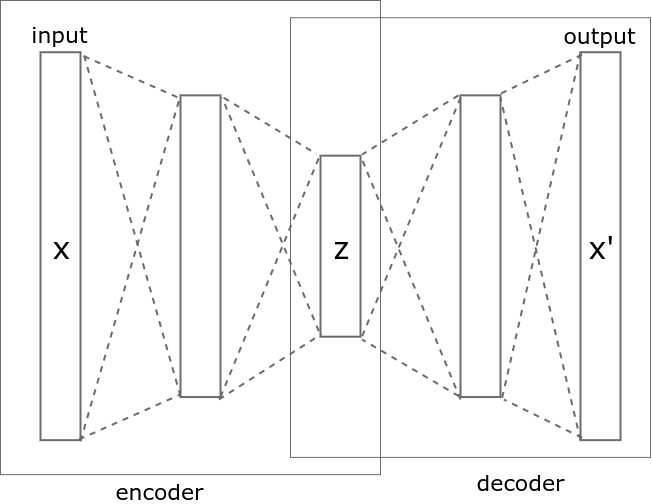}
  \caption{Codificador neuronal.}
  \label{encoder}
\end{figure}

Obsérvese que cuando una red feedforward tiene capas ocultas toda la comunicación que se produce entre la capa de entrada y la de salida pasa por cada una de las capas ocultas. De esta forma, si estamos intentando aproximar una función por medio de una red feedforward que tiene una capa oculta, tras el ajuste de los parámetros de la red (se haga por el procedimiento que se haga) podemos interpretar que la capa oculta mantiene la información necesaria de los datos de entrada que son imprescindibles para el cálculo de la función. Por ello, siempre desde el punto de vista de la función que calcula la red, podemos decir que la capa oculta codifica los datos de entrada, y los pesos (y bias) que se han usado definen la función de codificación entre ambos espacios \cite{deeplearning}. De igual forma, podemos entender que la parte de la red original que va desde la capa oculta que consideremos hasta la capa de salida define una decodificación hacia el espacio de llegada (ver Figura \ref{encoder}).

El objetivo de los codificadores neuronales es aprender una codificación a partir de un conjunto de datos. Si prescindimos de las capas posteriores (incluida la capa de salida original) a partir de una capa interior dada, obtenemos una nueva red neuronal que produce como salida una representación del espacio de entrada en un espacio de dimensión concreta (el número de neuronas en la capa oculta que se ha considerado). Debemos recordar que esta representación se consigue como aplicación parcial de una función que se ha obtenido a partir de una red feedforward completa que aproxima una función prefijada y, consecuentemente, la codificación obtenida es relativa a esta función (y, por supuesto, al proceso de aproximación).

Si el número de unidades usado en la capa oculta para la codificación difiere del número de unidades en la capa de entrada (y salida) estaremos además haciendo un cambio dimensional al realizar la codificación. De hecho, es uno de los métodos disponibles para realizar cambios de dimensionalidad manteniendo las características estructurales presentes en los conjuntos de entrenamiento (por ejemplo, las relaciones de proximidad o similitud).

Un \emph{autocodificador} es un caso concreto de codificador neuronal en el que se ha intentado aprender la función identidad y, en consecuencia, las capas de entrada y salida poseen el mismo número de neuronas. Es decir, el conjunto de muestras de entrenamiento sería  $S=\{(\vec{x}^1,\vec{x}^1),\dots,(\vec{x}^N,\vec{x}^N)\}$. Al igual que ocurre cuando trabajamos con codificadores, cuando la red alcanza un estado aceptable (es capaz de mostrar una salida suficientemente parecida a la entrada para cada uno de los ejemplos), las activaciones en las unidades de las capas ocultas capturan información del dato original $\vec{x}$ presentado en la capa de entrada \cite{bengio2009learning}.

En este trabajo usaremos codificadores neuronales como medio de inmersión de Grafos con Propiedades en espacios vectoriales. Para ello, haremos uso de redes neuronales entrenadas con funciones adecuadas con el fin de comprobar hasta qué punto las estructuras semánticas del grafo se conservan en las propiedades vectoriales de la inmersión. 

\section{Trabajos Relacionados}

El uso de codificadores neurales para la inmersión de palabras en espacios vectoriales ha proporcionado resultados muy interesantes. En 2013, T. Mikolov et al. \cite{eff} presentaron dos nuevas arquitecturas, bajo el nombre genérico de \emph{word2vec}, para aprender representaciones vectoriales de palabras tratando de minimizar la complejidad computacional y manteniendo propiedades semánticas y gramaticales presentes en los textos de los que se extraen: \emph{Continuous bag-of-words} (CBOW) y \emph{Skip-gram}. En este modelo, el \emph{contexto de una palabra} en un texto se define como el conjunto de palabras que aparecen en posiciones adyacentes a ésta. Las dos arquitecturas presentadas en \cite{eff} consisten en redes neuronales artificiales feedforward con $3$ capas: una capa de entrada, una capa oculta (capa de codificación) y una capa de salida, pero difieren en la función objetivo que intentan aproximar. Por un lado, los codificadores neuronales con arquitectura CBOW toman el contexto de una palabra como entrada y tratan de predecir la palabra en su salida. Por el contrario, los codificadores con la arquitectura Skip-gram reciben la palabra como entrada y tratan de predecir el contexto asociado a ella. El objetivo principal del trabajo de Mikolov et al. es reducir la complejidad del modelo neuronal para permitir al sistema aprender de un gran volumen de datos textuales. Hasta la llegada de word2vec, ninguna de las arquitecturas disponibles había podido entrenar con más de unos pocos millones de palabras. A través de la relación establecida entre las palabras del vocabulario y sus contextos, el modelo captura diferentes tipos de similitud, tanto funcional como estructural, y proporciona una inmersión vectorial de palabras que refleja estas similitudes.

En los últimos años se han desarrollado diferentes métodos que tratan de aprender representaciones vectoriales de entidades y relaciones en bases de conocimiento \cite{semantic,transe,nt}. Todas ellas representan las entidades como elementos de un espacio vectorial determinado y las relaciones como combinación de las representaciones de las entidades que participan en ella. 

En \cite{transe} se propone una inmersión de datos multi-relacionales en un espacio vectorial intentando verificar algunas propiedades adicionales. Concretamente, se busca una proyección de nodos y tipos de aristas, $\pi$, en el espacio vectorial con el objetivo de:
\begin{enumerate}
\item Minimizar la distancia $d(\pi(s)+\pi(l),\pi(t))$ de cada relación $(s \xrightarrow{l} t)$ existente (observada) en el conjunto de datos, donde $s$ representa el elemento origen de la relación, $l$ representa el tipo de relación, y $t$ representa el elemento destino de la relación.
\item Maximizar las distancias $d(\pi(s')+\pi(l),\pi(t))$, $d(\pi(s)+\pi(l'),\pi(t))$ y $d(\pi(s)+\pi(l),\pi(t'))$, donde $s'$ y $t'$ representan nodos del grafo, y $l'$ representa un tipo de relación del grafo, para los que las relaciones $s' \xrightarrow{l} t$, $s \xrightarrow{l}t'$ y $s\xrightarrow{l'} t$ no existen (relaciones no observadas) en el grafo.
\end{enumerate}
Para mejorar la eficiencia del algoritmo, los autores hacen un muestreo aleatorio del grafo original, tanto para las relaciones existentes como para las no existentes. En \cite{semantic}, y con el fin de conseguir mejores resultados en la proyección, los autores siguen un procedimiento similar pero haciendo uso de una red neuronal siamesa\footnote{Una red neuronal siamesa es un tipo de red neuronal comparativa compuesta por dos redes que comparten pesos y arquitectura (cada una recibe un dato a ser comparado) y cuyas salidas son comparadas mediante una función de distancia.} en vez de una red neuronal estándar. En \cite{embedding} se agrupan algunas de estas técnicas sobre un mismo marco teórico general que permite comparar la complejidad de los modelos obtenidos y de los resultados.

Pese a la relación existente entre estos trabajos y nuestra aproximación, el segundo requerimiento que imponen para maximizar las distancias en las relaciones no observadas (y que es imprescindible para los resultados que obtienen) va en contra de uno de los objetivos que perseguimos, ya que no suponemos que el grafo con propiedades original tenga información completa y, en consecuencia, las relaciones no observadas pueden deberse a una carencia informativa y no a una inexistencia real. Aún más, precisamente la predicción de este tipo de relaciones no observadas es una de las razones por las que buscamos una inmersión en un espacio que ofrezca una capacidad de análisis adicional. 

Además, la mayoría de los trabajos que realizan aprendizaje de representaciones vectoriales de entidades y relaciones en bases de conocimiento tienen como objetivo evaluar la posibilidad de existencia de determinadas relaciones (\emph{Link Prediction}), condicionando con un aprendizaje supervisado la representación de las entidades. En nuestro caso, la codificación que trataremos de aprender estará sólo condicionada por conseguir representaciones vectoriales que capturen la similitud de los contextos en los que se encuentran inmersas las entidades que representan, abriendo así la posibilidad de usar estas representaciones a un abanico más amplio de tareas.

DeepWalk \cite{deepwalk} es una metodología reciente que utiliza codificadores neuronales para representar los nodos de un grafo uni-relacional haciendo uso de una idea muy similar a la que Mikolov et al. presentaron para la inmersión de grandes conjuntos de textos en espacios vectoriales \cite{eff}. En concreto, en el citado trabajo el grafo uni-relacional es \emph{linealizado} a partir de la generación de caminos aleatorios truncados, interpretando los caminos obtenidos como frases y considerando a partir de ellos un conjunto de entrenamiento con la forma $$S=\{(n_1, C_1), \dots ,(n_N,C_N)\}$$
\noindent donde $n_i$ representa un nodo concreto del grafo y $C_i$ un contexto generado a partir de dichos caminos aleatorios truncados, es decir, el conjunto de nodos que aparecen en el mismo camino según el orden de recorrido. Posteriormente, y de forma completamente equivalente a word2vec, se usa $S$ para entrenar un codificador neuronal con arquitectura Skip-gram y obtener así una inmersión de los nodos del grafo uni-relacional en un espacio vectorial. El método propuesto no permite trabajar con grafos uni-relacionales \textit{grandes} de manera eficiente, y tampoco con grafos multi-relacionales.

En \cite{Tang:2015:LLI:2736277.2741093} se presenta la metodología LINE para aprender una representación $d$-dimensional de los elementos inmersos en un grafo uni-relacional a través de dos fases: primero se aprenden $d/2$ dimensiones generando caminos aleatorios en modo Breath-First Search, posteriormente, se aprenden las $d/2$ dimensiones restantes haciendo un muestreo de los nodos que están estrictamente a distancia 2 del nodo origen.

Node2vec \cite{grover2016node2vec} agrupa y extiende las ideas presentadas en DeepWalk y LINE ampliando las posibilidades a la hora de construir los caminos aleatorios en el grafo. Concretamente, los autores desarrollan un algoritmo flexible que, a través de dos hiperparámetros, permite modificar la generación de los caminos aleatorios que exploran el entorno de los nodos y dan lugar a su contexto. A partir de dos estrategias estándar de búsqueda, Breath-First Sampling (BFS) y Depth-First Sampling (DFS), los dos parámetros permiten controlar si el camino aleatorio tiende a una estrategia BFS o DFS. En particular, afirman que un muestreo guiado por una estrategia BFS da lugar a inmersiones que reflejan la \textit{equivalencia estructural} entre las entidades y que un muestreo guiado por una estrategia DFS da lugar a una inmersión en la que se refleja la \textit{homofilia}. Por medio de experimentos los autores evalúan su capacidad para llevar a cabo \textit{Multi-Label Classification} y \textit{Link Prediction}. En este sentido, DeepWalk sería un caso concreto en el que el valor de ambos parámetros es el mismo. 

En ninguno de los trabajos anteriores (\cite{deepwalk,Tang:2015:LLI:2736277.2741093,grover2016node2vec}) se trabaja con grafos multi-relacionales, y se limitan a detectar la \textit{homofilia} y la \textit{equivalencia estructural} en las representaciones vectoriales.

En los últimos años se han publicado algunos trabajos que hacen uso de Redes Neuronales Convolucionales (CNN) para crear representaciones vectoriales de los nodos de un grafo uni-relacional. En \cite{kipf2016semi}, el objetivo es aprender una función que codifique las características de los nodos de un grafo a partir de una descripción de las mismas (almacenadas en una matriz de descripciones) y de la matriz de adyacencias del grafo. También se pueden obtener salidas que representen el grafo completo aplicando alguna operación de tipo \textit{pooling} \cite{NIPS2015:5954}. En \cite{DBLP:journals/corr/DefferrardBV16} sin embargo, se trabaja con el operador de convolución en el campo de Fourier, y generalizan las redes convolucionales para pasar de su definición original en espacios euclídeos regulares de baja dimensión (donde se trabaja naturalmente con imágenes, vídeos o audios) a poder trabajar con dominios irregulares de alta dimensión (grafos multi-relacionales obtenidos de redes sociales o de fenómenos biológicos). En \cite{schlichtkrull2017modeling} se presenta una extensión del modelo Graph Convolutional Network \cite{kipf2016semi} denominado Relational Graph Convolutional Network, que permite realizar aprendizaje a través de las operaciones de convolución y \textit{pooling} típicas de las redes convolucionales sobre grafos multi-relacionales.

En \cite{DBLP:journals/tnn/ScarselliGTHM09} se define el modelo Graph Neural Network Model, que convierte el grafo de datos en una red neuronal recurrente, y cada nodo del mismo en una red feedforward multi-capa. La combinación de estas estructuras permite llevar a cabo un aprendizaje supervisado en el que muchos de los pesos de la red son compartidos, reduciendo así el coste en el aprendizaje.

En \cite{Chang:2015:HNE:2783258.2783296} se presenta Heterogeneous Network Embedding, un framework para hacer inmersiones de redes que conectan datos de diferentes tipos en espacios de baja dimensión utilizando una red neuronal profunda. Como el aprendizaje que se lleva a cabo es no supervisado, la nueva representación es adecuada para aplicar cualquier algoritmo de aprendizaje automático ya que no se ha condicionado el aprendizaje a una tarea determinada. En \cite{Jacob:2014:LLR:2556195.2556225} se enfrentan a la tarea de asignación automática de etiquetas a nodos de diferentes tipos en una red heterogénea que no tiene tipos en las aristas. El algoritmo que presentan está diseñado para aprender las dependencias existentes entre los conjuntos de etiquetas asociadas a los diferentes nodos y para inferir las etiquetas asociadas a un nodo explotando las propiedades del grafo global y las caraceristicas de los nodos vecinos. Para ello, imponen dos objetivos: (1) intentar agrupar nodos del mismo tipo que estén conectados (con menos intensidad cuanto más largo sea el camino que los conecta), y (2) intentar agrupar nodos de diferentes tipos si comparten contextos. Para trabajar con grafos con propiedades, representan cada propiedad como un nodo nuevo. 

Como hemos mostrado, existen numerosas metodologías para realizar inmersiones de grafos en espacios vectoriales, algunas de ellas están limitadas a trabajar con grafos uni-relacionales, otras condicionan la codificación a través de la generación de relaciones no observadas o no capturan las características semánticas típicas de un grafo con propiedades. Nuestra propuesta, que pasamos a detallar a continuación, tiene como objetivo obtener inmersiones que no estén afectadas por estas limitaciones.

\section{Inmersiones de Grafos con Propiedades}
\label{sec:met}
Entre las opciones barajadas, la arquitectura seleccionada para el codificador neuronal fue la arquitectura CBOW debido a que, a pesar de su simplicidad y el bajo coste computacional en su entrenamiento, obtiene buenos resultados en la tarea de capturar relaciones tanto \textit{sintácticas} como \textit{semánticas} entre los elementos codificados \cite{eff}.

Así pues, la metodología que presentamos a continuación hace uso de un codificador neuronal, similar al usado en la arquitectura CBOW, para codificar los elementos de un grafo con propiedades en un espacio vectorial adecuado.

Aunque un grafo con propiedades tiene muchos elementos constitutivos, en una primera aproximación, y con el fin de evaluar hasta qué punto se mantiene la estructura semántica dada por las aristas, haremos una proyección usando únicamente el conjunto de nodos sobre el espacio vectorial. De esta forma, siguiendo con la analogía que nos ofrece el algoritmo word2vec, nuestro vocabulario será el conjunto de nodos del grafo (y también sus propiedades asociadas).

Un contexto, $C$, asociado a un nodo $n\in V$ se obtiene seleccionando, aleatoriamente y con repetición, un número determinado de nodos vecinos a $n$ y propiedades suyas, independientemente del tipo de relación que los conecta y del tipo de propiedad. El número de nodos/propiedades seleccionados determina el \emph{tamaño de la ventana de selección}.

Siguiendo una metodología similar a las vistas en el apartado anterior, generaremos un conjunto de entrenamiento formado por pares $(n,C)$, donde $n\in V$ y $C$ es uno de sus contextos asociados. El conjunto de muestras es utilizado para entrenar el codificador neuronal y, a continuación, las activaciones de la capa oculta de la red neuronal se utilizan como representación vectorial de cada uno de los nodos. 

\begin{figure}[htb]
  \centering
  \includegraphics[width=\textwidth]{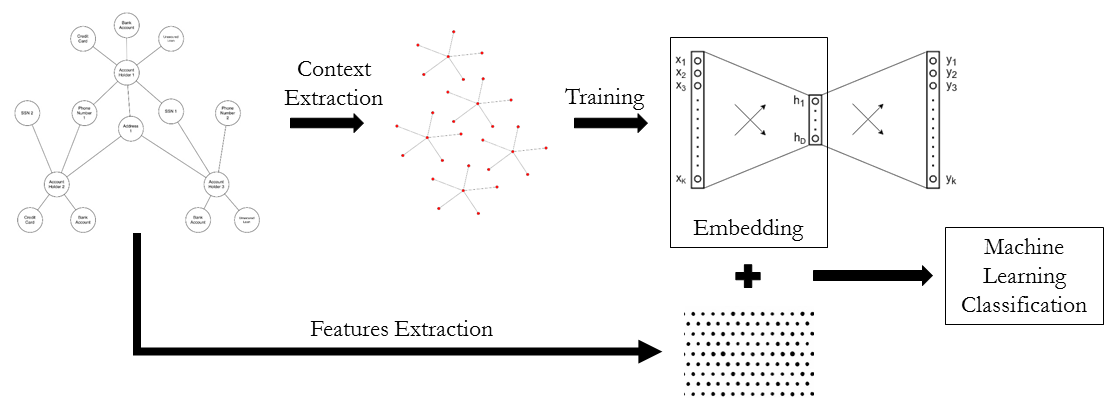}
  \caption{Representación esquemática de la metodología propuesta.}
  \label{diagram}
\end{figure}

Una vez entrenado el codificador, usaremos estas representaciones vectoriales para intentar resolver algunas tareas de clasificación y descubrimiento en el grafo original. Los resultados de estas tareas proporcionarán una medida de fiabilidad sobre las inmersiones conseguidas (Fig. \ref{diagram}). El Algoritmo \ref{alg:graph2vec} muestra el procedimiento seguido.

\begin{algorithm}
\begin{algorithmic}[1]
\STATE {$training\_set = \{\}$}
\FOR {each $0<i\leq N$}
\STATE {$n$ = randomly selected element from $V_G$}
\STATE {$C = \{\}$}
\FOR {each $0<i\leq ws$}
\STATE {$e$ = randomly selected element from $\mathcal{N}(n) \cup \mu_G(n)$}
\STATE {$C = C \cup \{e\}$}
\ENDFOR
\STATE {$training\_set = training\_set \cup \{(v,C)\}$}
\ENDFOR
\STATE {Train a $CBOW$-like architecture with $D$ neurons in hidden layer using $training\_set$}
\RETURN The resulting encoding for each element in $V$
\end{algorithmic}
\caption{GG2Vec($G$,$N$,$ws$,$D$)}\label{alg:graph2vec}
\end{algorithm}

En el procedimiento de inmersión los \emph{parámetros libres} del modelo, que habrá que ajustar en los diversos experimentos para analizar su eficacia y viabilidad, son:
\begin{itemize}
\item $D$, tamaño de la capa oculta, determina la dimensión del espacio vectorial en el que haremos la inmersión de los elementos del grafo.
\item $N$, tamaño del conjunto de entrenamiento, número de pares $(n, C)$ utilizados para entrenar el codificador.
\item $ws$, tamaño de la ventana de selección, número de vecinos y propiedades considerados para construir los contextos de nodos de $V$.
\end{itemize}

En lo que sigue, notaremos por $\pi:V \to \mathbb{R}^D$ la inmersión que hemos obtenido a partir del codificador neuronal entrenado.

Debido a que en la implementación y experimentos sobre bases de datos reales vamos a trabajar con grafos binarios con propiedades (no hipergrafos), tras haber obtenido una proyección sobre los nodos del grafo, se induce una proyección de las aristas en el mismo espacio vectorial (que notaremos también por $\pi$):
\begin{definition}
Si $G=(V,E,\tau, \mu)$ es un Grafo con Propiedades, y $\pi:V\to \mathbb{R}^D$ es una inmersión de los nodos del grafo, damos una extensión de la inmersión al conjunto de aristas, $\pi:E\to \mathbb{R}^D$ de la siguiente forma:

$$e\in E,\ s\overset{e}{\rightarrow}t\mbox{, entonces } \pi(e)=\overrightarrow{\pi(s) \pi(t)}$$
\end{definition}

Como las operaciones habituales en espacios vectoriales son de uso extendido en las actuales unidades de cálculo (procesadores y GPUs), esta nueva representación puede ayudar a desarrollar algoritmos más eficientes para analizar, reparar y extraer información de conjuntos de datos multi-relacionales, en general, y más concretamente, de bases de datos en grafos. Por ejemplo, algunas tareas que pueden verse mejoradas con este tipo de inmersiones son:
\begin{itemize}
\item Clústers formados por nodos/aristas en el nuevo espacio vectorial pueden ayudar a asignar propiedades faltantes a los elementos de un grafo (haciendo uso de relaciones de distancia, linealidad o clusterización, por ejemplo).
\item La representación vectorial de los elementos de un grafo puede ayudar a obtener medidas de similitud entre ellos.
\item El análisis de los vectores asociados a las diferentes familias de relaciones (aquellas que comparten un tipo común, o verifican propiedades similares, por ejemplo) puede ayudar a detectar relaciones faltantes en el conjunto de datos original pero que en la nueva representación se hacen evidentes (la disposición de dos nodos cumple con el \emph{vector representante} de algún tipo de relación, a pesar de que esa relación no aparece en el grafo).
\item El análisis de la representación de las aristas que forman \emph{caminos} en el grafo original puede ayudar a desarrollar formas más eficientes de detectar la existencia de dichos caminos en el grafo original.
\end{itemize}

\section{Evaluación Empírica}

Es momento ahora de realizar una evaluación empírica de nuestro método con dos objetivos claramente diferenciados: 
\begin{enumerate}
	\item Analizar que las representaciones vectoriales que se obtienen a partir de grafos con propiedades mantienen características semánticas presentes en los mismos. 
	\item Evaluar diferentes aplicaciones que hacen uso de la inmersión propuesta para realizar tareas de clasificación y descubrimiento.
\end{enumerate}

En nuestro contexto, diremos que una inmersión vectorial respeta la semántica de un grafo con propiedades si, a partir de la nueva representación, es posible obtener los tipos asociados a nodos y aristas sin que éstos hayan estado presentes durante la inmersión. El tipo de cada nodo o arista estará determinado por la clave $\tau \in R$. Como hemos comentado, para hacer estas comprobaciones, las diversas inmersiones que vamos a calcular no recibirán información acerca de los tipos de nodos o aristas que componen el grafo (formalmente, no recibirán información sobre $\tau$). Los contextos asociados a los diferentes nodos del grafo, y que son utilizados para crear el conjunto de entrenamiento, son generados seleccionando aleatoriamente un número (que viene determinado por el tamaño de la ventana de selección) de nodos vecinos y de valores de sus diferentes propiedades en $\mu$ (sin $\tau$).

%Con el objetivo de verificar que determinadas estructuras \textit{semánticas} presentes en la representación original $G = (V,E,\mu)$ se mantienen en la nueva representación del grafo analizaremos en qué medida la función $\tau$, que determina los tipos de cada nodo o arista, puede ser deducida en la nueva representación a pesar de que el proceso de inmersión de $G$ nunca recibió información sobre ella. Si es así, se verificaría que los tipos asociados a nodos y aristas pueden ser deducidos a partir de la estructura topológica original de la red y de las propiedades internas de sus elementos $(V,E,\mu)$ haciendo uso de algoritmos de aprendizaje.

\subsection{Detalles de la implementación y experimentos}

Se ha elegido Python como lenguaje de programación para llevar a cabo la evaluación experimental señalada\footnote{\url{https://github.com/palmagro/gg2vec}}. Para la implementación de la arquitectura CBOW se ha utilizado el conjunto de herramientas {\it Gensim}\footnote{\url{https://radimrehurek.com/gensim}} (versión 0.12.4). Además, se ha utilizado Neo4j\footnote{\url{http://neo4j.com}} como sistema de persistencia para los grafos con propiedades analizados.

Cada experimento de inmersión, con parámetros prefijados, se ha repetido 10 veces, valor que experimentalmente ha mostrado una desviación estándar en los resultados experimentales con respecto a la predicción de los tipos de nodos y aristas inferior al $2\%$. En el caso de las tareas relacionadas con \textit{Entity Retrieval} estas desviaciones suben hasta el $7.8\%$, y en el caso de la obtención de los nodos destino de un traversal han quedado acotadas por $8.9\%$.

En lo relativo a las tareas posteriores que hacen uso de otros modelos de aprendizaje para validar la inmersión se han considerado $k$-NN, Random Forest y Redes Neuronales. Para los test de clasificación generales, y salvo que se indique lo contrario, se ha utilizado como modelo base de comparación $k$-NN con $k = 3$. 

\subsection{Datasets}
\label{sec:ds}

Los experimentos se han llevado a cabo en 3 grafos con propiedades diferentes. Dos de ellos son ampliamente conocidos por la comunidad científica relacionada con el análisis de datos semánticos: WordNet y TheMovieDB. El tercero es un conjunto de datos desconocido para la comunidad denominado Ecuadorian Intangible Cultural Heritage. 

Los conjuntos de datos han sido parcialmente manipulados para reducir su tamaño y complejidad por motivos de eficiencia. A continuación damos algunos detalles acerca de cada uno de estos conjuntos para contextualizar las características que encontraremos en los resultados obtenidos.

\textbf{WordNet®} \cite{wn} es una base de datos de nombres, verbos, adjetivos y adverbios de la lengua inglesa. Es uno de los recursos más importantes en el área de lingüística computacional, y se ha construido como una combinación de diccionario y tesauro, ideado para que su uso sea intuitivo. Cada elemento en la base de datos representa una lista de palabras sinónimas (que denominan \textit{synset}), y las relaciones que se establecen entre los elementos se dan tanto a nivel léxico como semántico, razón por la cual esta base de datos ha sido ampliamente usada en el análisis sintáctico de textos y en entornos de extracción automática de información semántica. 

\begin{figure}[htb]
	\centering
	\includegraphics[scale=0.3]{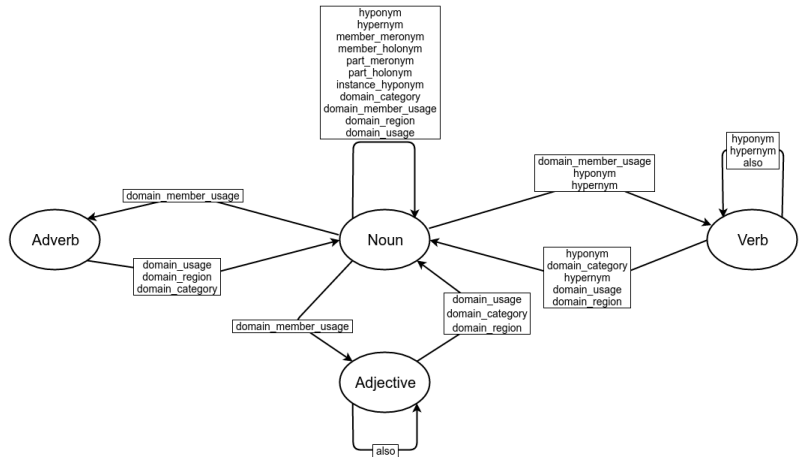}
	\caption{Esquema de datos de WordNet.}
	\label{wn-schema}
\end{figure}

\begin{figure}[htb]

     \begin{center}
        \subfigure[Distribución de nodos por tipo.]{%
            \label{wordnet-nodes}
            \includegraphics[scale=0.15]{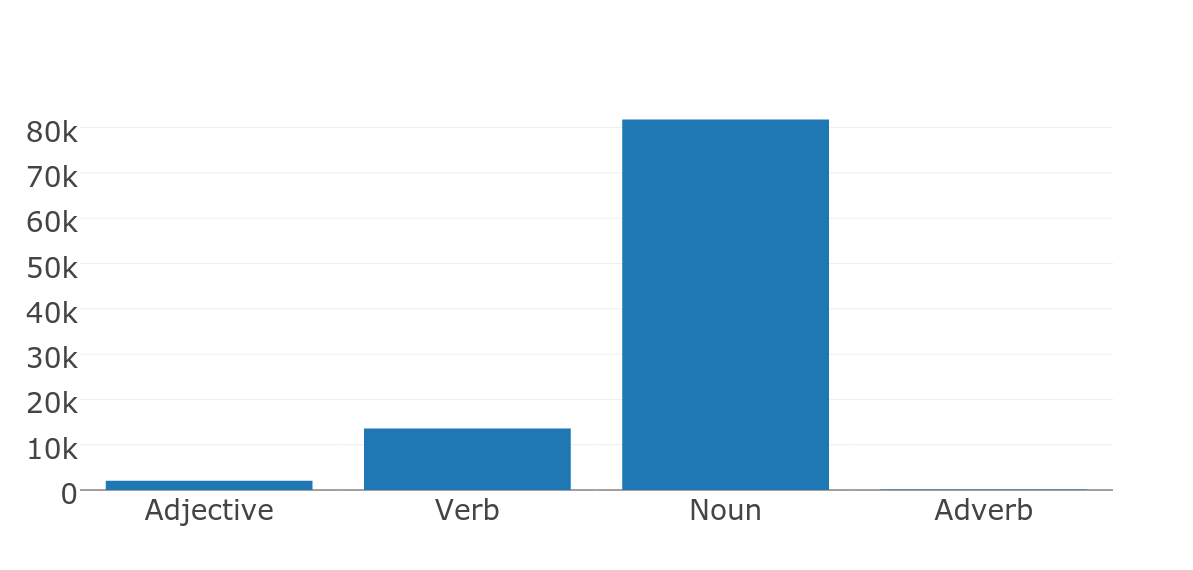}
        }%
        \subfigure[Distribución de aristas por tipo.]{%
            \label{wordnet-links}
            \includegraphics[scale=0.15]{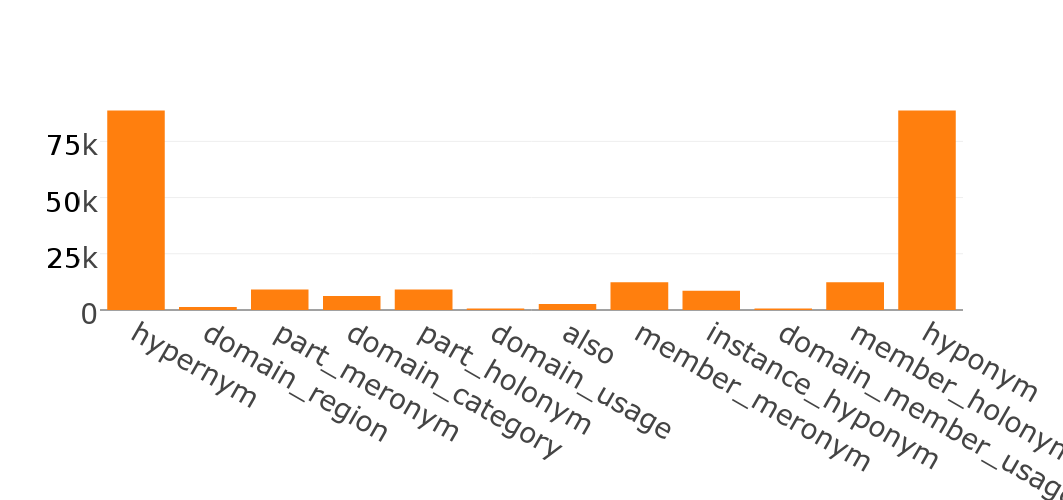}
        }%
    \end{center}
    \caption{%
       Distribución de nodos y aristas en WordNet.
     }%
     \label{wn1}
\end{figure}

Para este trabajo hemos utilizado una sección de la versión 3.0, considerando únicamente las entidades y relaciones que se muestran en la Figura \ref{wn-schema} (de manera similar a \cite{semantic}), obteniendo de esta forma un grafo con 97.593 nodos y 240.485 relaciones, con una distribución de tipos en nodos y aristas tal y como muestra la Figura \ref{wn1}. 

\begin{figure}[htb]
	\centering
	\includegraphics[scale=0.40]{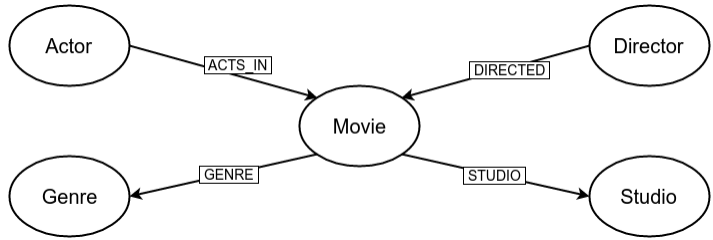}
	\caption{Esquema de datos de TMDb.}
	\label{tmdb-schema}
\end{figure}

\begin{figure}[htb]
	\begin{center}
		\subfigure[Distribución de nodos por tipo.]{%
			\label{cine-nodes}
			\includegraphics[scale=0.14]{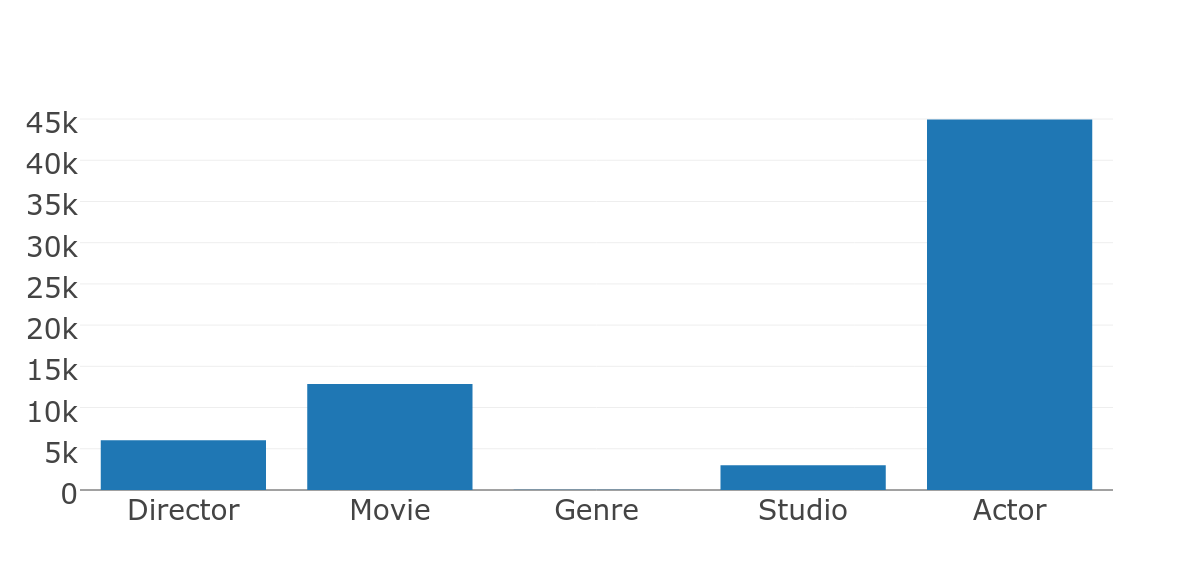}
		}%
		\subfigure[Distribución de aristas por tipo.]{%
			\label{cine-links}
			\includegraphics[scale=0.14]{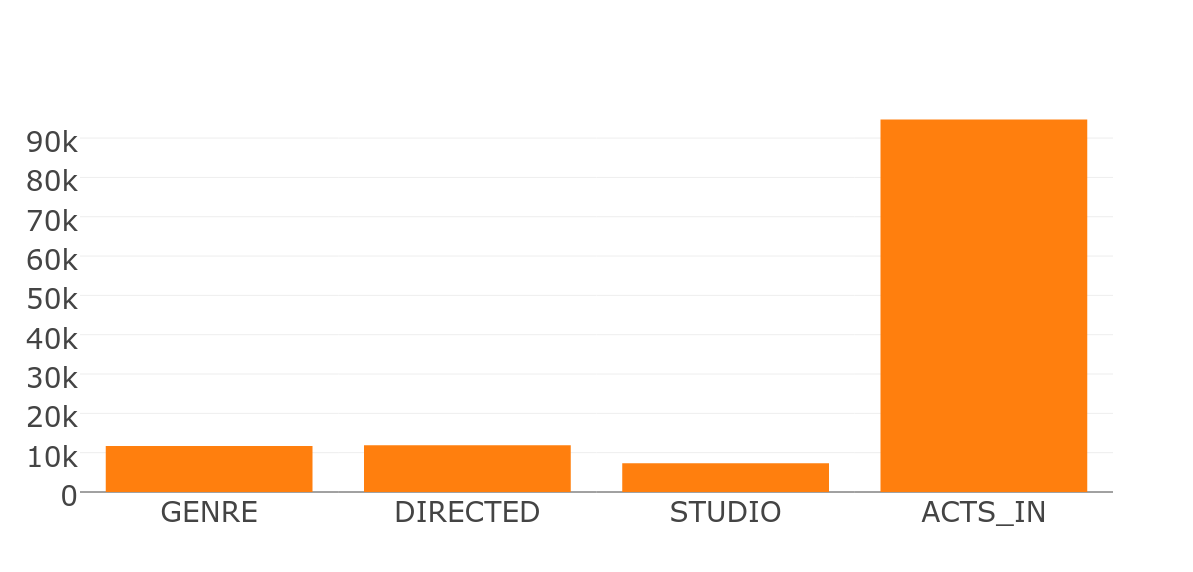}
		}%
	\end{center}
	\caption{%
		Distribución de nodos y aristas en TMDb.
	}%
	\label{cine2}
\end{figure}

\textbf{TheMovieDB} (TMDb)\footnote{\url{https://www.themoviedb.org}} es un conjunto de datos que contiene información sobre actores, películas y contenidos de televisión. Para nuestros experimentos hemos considerado todas las entidades de TMDb que están conectadas por relaciones pertenecientes a los tipos \texttt{acts\textunderscore in, directed, genre} y \texttt{studio}, obteniendo un grafo con 66.020 nodos y 125.624 relaciones. La Figura \ref{tmdb-schema} muestra una representación gráfica del esquema de datos presente en este dataset, y en la Figura \ref{cine2} se muestra la distribución por tipos de nodos y aristas en el subconjunto de TMDb considerado.

Cabe destacar que los tipos \texttt{Actor} y \texttt{Director} están solapados, concretamente, en nuestro conjunto existen 44.097 nodos que sólo tienen asignado el tipo \texttt{Actor}, 5.191 nodos que sólo tienen asignado el tipo \texttt{Director}, y 846 nodos que poseen los tipos \texttt{Actor} y \texttt{Director} al mismo tiempo (nodos multi-tipo). 

\begin{figure}[htb]
	\centering
	\includegraphics[scale=0.30]{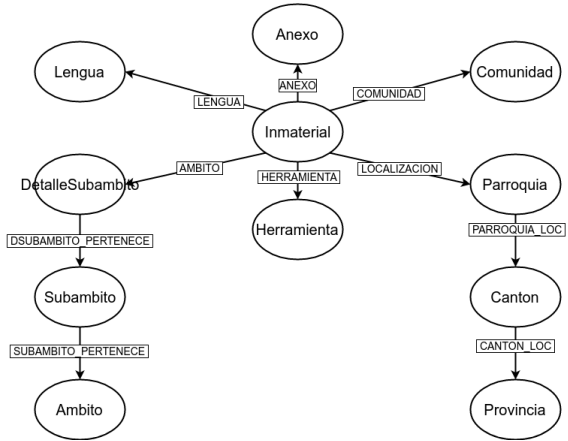}
	\caption{Esquema de datos de EICH.}
	\label{eich-schema}
\end{figure}

\begin{figure}[htb]
	\begin{center}
		\subfigure[Distribución de nodos por tipo.]{%
			\label{eich-nodes}
			\includegraphics[scale=0.14]{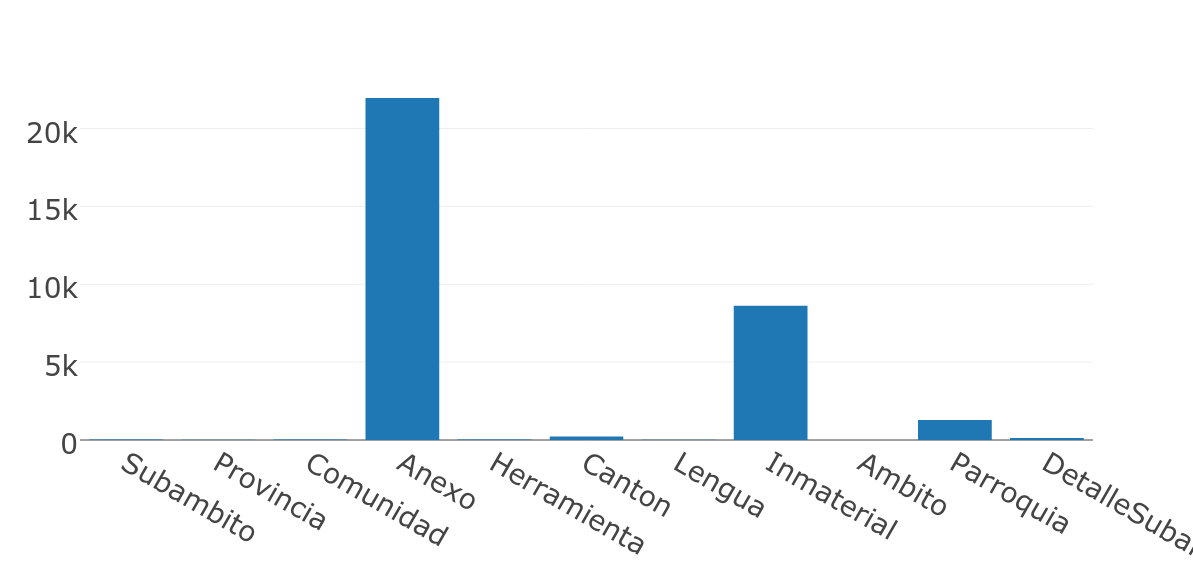}
		}%
		\subfigure[Distribución de aristas por tipo.]{%
			\label{eich-links}
			\includegraphics[scale=0.24]{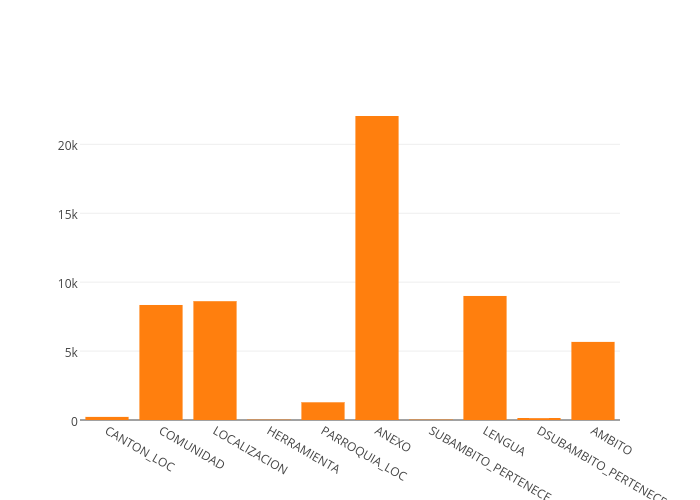}
		}%
	\end{center}
	\caption{%
		Distribución de nodos y aristas en EICH.
	}%
	\label{eich2}
\end{figure}

\textbf{Ecuadorian Intangible Cultural Heritage} (EICH o Base de Datos del Patrimonio Cultural Inmaterial del Ecuador) corresponde a una sección de la base de datos del Instituto Nacional de Patrimonio Cultural Ecuatoriano \footnote{Accesible desde \url{http://www.inpc.gob.ec}} que contiene 38.990 nodos y 55.358 relaciones distribuidas a través de 11 tipos de nodos y 10 tipos de aristas, con información sobre el patrimonio cultural inmaterial del Ecuador. Esta base de datos es la más heterogénea de las 3 analizadas, presentando mayor tipología tanto en nodos como en aristas, y además sus elementos poseen más propiedades que las de los elementos de las otras dos bases consideradas. Las Figuras \ref{eich-schema} y \ref{eich2} muestran respectivamente el esquema  y distribución de nodos y aristas en este grafo.  
%Ejemplos de relaciones en EICH son:
%$$\texttt{quito}\overset{\texttt{localizacion}}{\rightarrow}\texttt{pichincha}$$
%$$\texttt{canciones\textunderscore tradicionales\textunderscore pastaza}\overset{\texttt{lengua}}{\rightarrow}\texttt{achuar}$$

\begin{figure}[htb]
     \begin{center}
        \subfigure[WordNet]{%
            \label{ntp-ns}
            \includegraphics[scale=0.17]{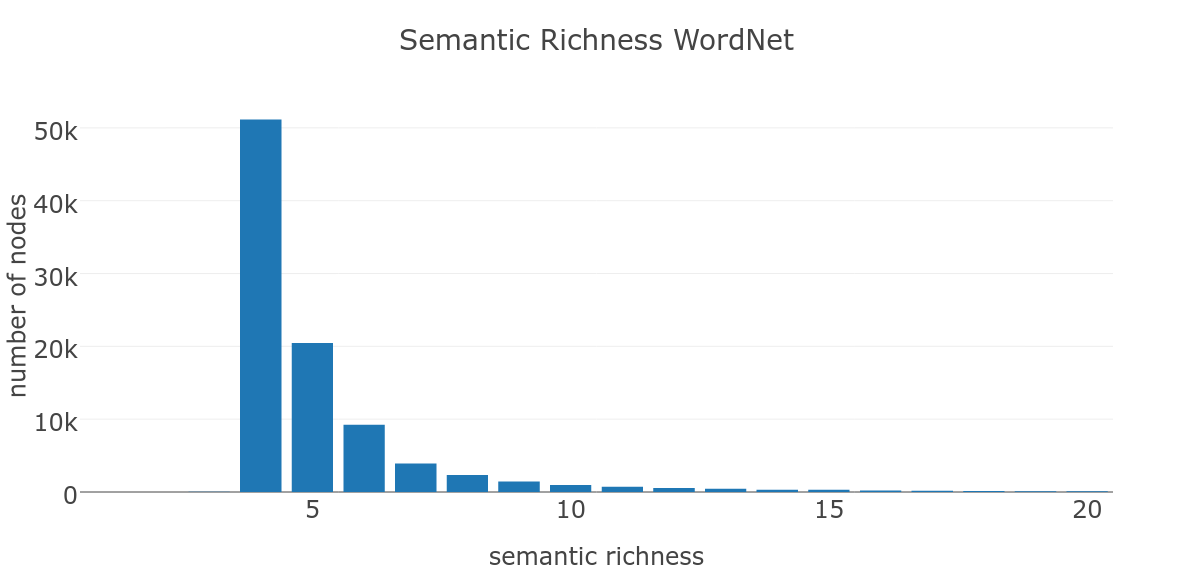}
        }%
     
        \subfigure[TMDb]{%
            \label{ntp-ndim}
            \includegraphics[scale=0.17]{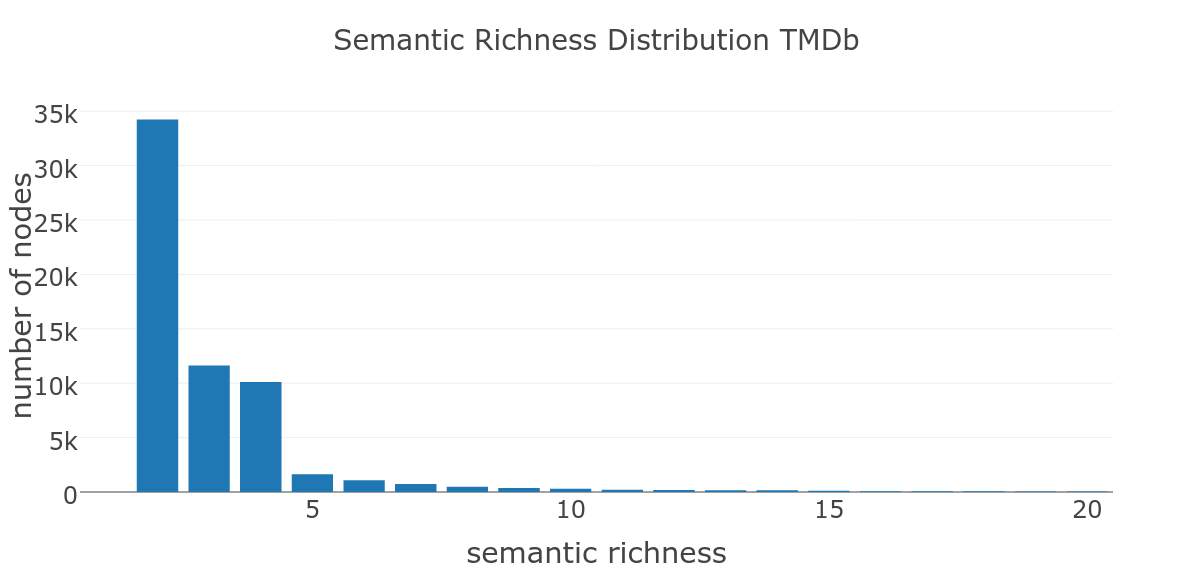}
        }\\ %  ------- End of the first row ----------------------%
     
        \subfigure[EICH]{%
            \label{ntp-l}
		    \includegraphics[scale=0.17]{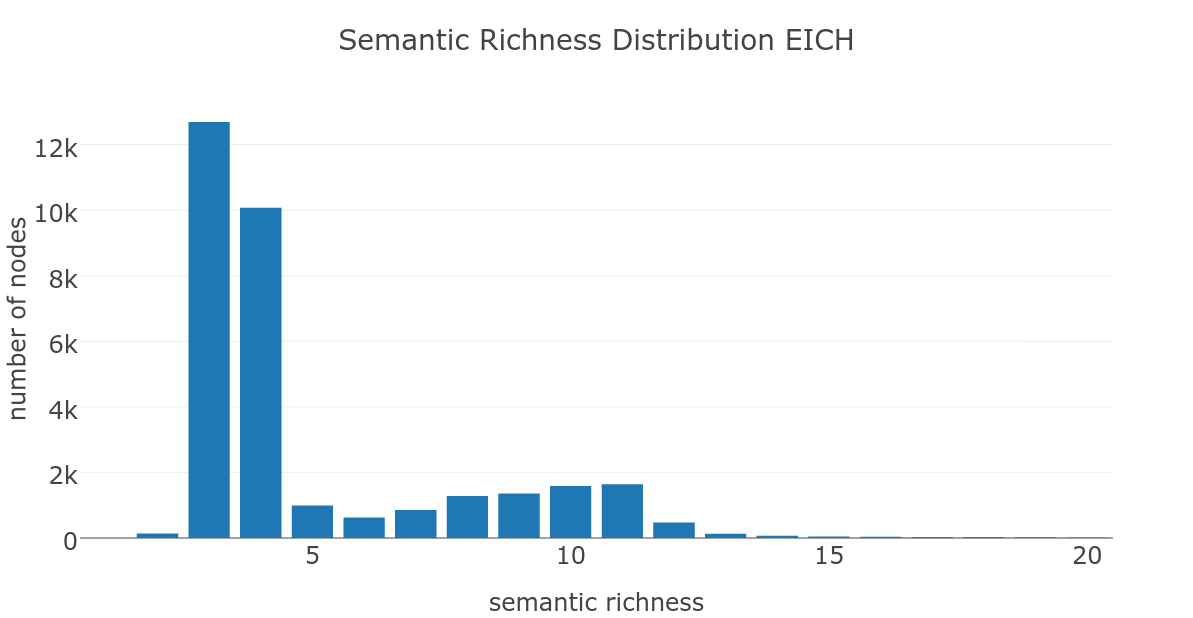}
        }%
    \end{center}
    \caption{%
       Riqueza semántica (inferiores a 20).
     }%
   \label{rsd}
\end{figure}

Si medimos, de forma muy primitiva, la \emph{riqueza semántica} de un nodo como la suma del número de relaciones en las que participa más el número de propiedades que posee, podemos construir el histograma de riqueza semántica para cada uno de los conjuntos de datos anteriores (Fig. \ref{rsd}). En el caso de WordNet, el promedio de riqueza semántica es $5,56$, en el caso de TMDb es $3,21$ y en el caso de EICH es $7,86$. El distinto comportamiento que muestra esta distribución en los casos estudiados puede ayudarnos a entender e interpretar los resultados obtenidos.

\subsection{Predicción de Tipos de Nodos} 

Nuestro primer experimento tiene como objetivo predecir la función $\tau$ que asocia tipos a los nodos del grafo y que, como comentamos anteriormente, no se proporciona durante el proceso de entrenamiento. Por supuesto, de forma similar, podríamos intentar predecir cualquier propiedad de $\mu$, manteniendo siempre la precaución de que la característica evaluada no haya sido utilizada durante el entrenamiento. 

Una primera intuición acerca de que las inmersiones conseguidas mantienen las estructuras semánticas (concretamente, el tipo de cada nodo) la podemos obtener analizando cómo se distribuyen los diversos tipos en el espacio vectorial sobre el que se ha hecho la inmersión. 

\begin{figure}[htb]
	\begin{center}
		\subfigure{%
			\label{cine-nsample1}
			\includegraphics[scale=0.28]{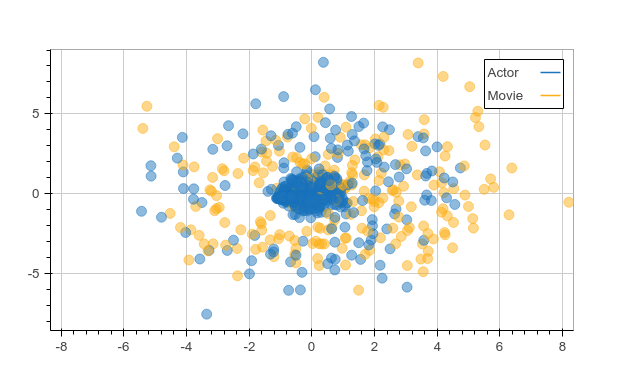}
		}%
		\subfigure{%
			\label{cine-nsample2}
			\includegraphics[scale=0.28]{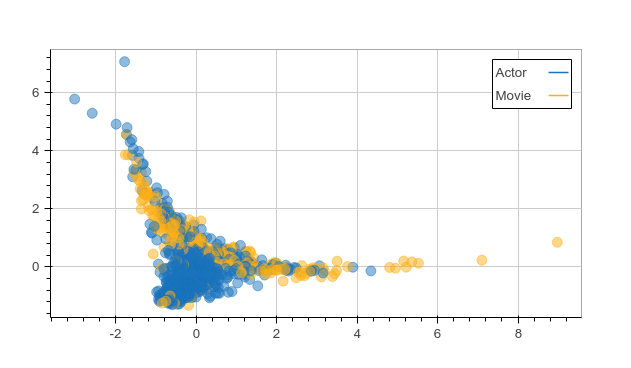}
		}%
	\end{center}
	\caption{%
		Representaciones 2D de nodos de tipo \texttt{Movie} y \texttt{Actor} en TMDb.
	}%
	\label{cine-nsample}
\end{figure}

La figura \ref{cine-nsample} muestra dos proyecciones de una sección de la inmersión obtenida para el grafo TMDb. La representación de la izquierda muestra una selección aleatoria de los vectores asociados a los nodos de tipo \texttt{Movie} y \texttt{Actor} haciendo uso de una inmersión en un espacio de dimensión 200 (que ha sido proyectada posteriormente sobre un espacio bidimensional haciendo uso de la técnica Multi-Dimensional Scaling \cite{mds} para facilitar su visualización), mientras que la representación de la derecha muestra la misma sección del grafo haciendo uso de una inmersión sobre un espacio de dimensión 2 directamente. A pesar de que la reducción de dimensionalidad considerada es a todas luces excesiva (pero necesaria para poder visualizar estos conjuntos en estas páginas), ambas representaciones muestran que las inmersiones de los nodos del grafo TMDb no se distribuyen aleatoriamente respecto del tipo, sino que siguen un patrón, lo que evidencia que, en efecto, la inmersión obtenida mantiene información relativa al tipo de los nodos a pesar de que, como comentamos, la función $\tau$ nunca ha sido utilizada en su construcción.

Además de la libertad de elección en los parámetros que intervienen en la codificación, encontramos algunos grados de libertad adicionales al decidir qué máquina de aprendizaje se usará posteriormente para clasificar la inmersión de los nodos del grafo. Como primera aproximación, y a pesar de la alta carga computacional que demanda, se ha realizado un estudio exhaustivo de los parámetros libres del modelo haciendo uso del método de clasificación $k$-NN para explorar la dependencia que muestra la inmersión respecto del parámetro $k$. La razón por la que se ha elegido este modelo se centra en dos aspectos fundamentales: solo depende de un parámetro propio (el valor de $k$, que se sabe que funciona relativamente bien para $k=3$ de forma general) y, a pesar de su simplicidad, proporciona resultados robustos que sirven de base comparativa para otros modelos de clasificación más sofisticados. 

\begin{table}[tb]
	\centering
	\caption{Parámetros de inmersión}
	\label{ntp-optimos}
	\begin{tabular}{lcccc}
		& \meg{ $N$ } & \meg{ $D$}   & \meg{ Tam. ventana}  &\meg{Predicción} \\ \hline
		\meg{ TMDb} & 400.000    & 150         & 3       &$\simeq 72\%$  \\ \hline
		\meg{ WordNet}      & 1.000.000            & 50 & 8        &$\simeq 96\%$  \\ \hline
		\meg{ EICH}      & 300.000             & 20         & 2 & $\simeq 83\%$         \\ \hline
	\end{tabular}
\end{table}

\begin{figure}[htb]
	\begin{center}
		\subfigure[En función del tamaño del conjunto de entrenamiento.]{%
			\label{ntp-ns2}
			\includegraphics[scale=0.35]{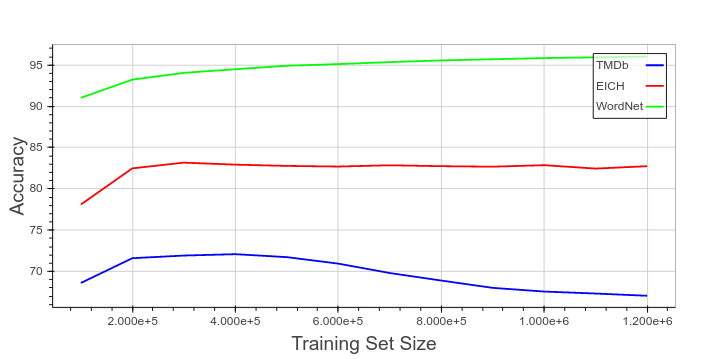}
		}\\ %
		\subfigure[En función del número de dimensiones.]{%
			\label{ntp-ndim2}
			\includegraphics[scale=0.35]{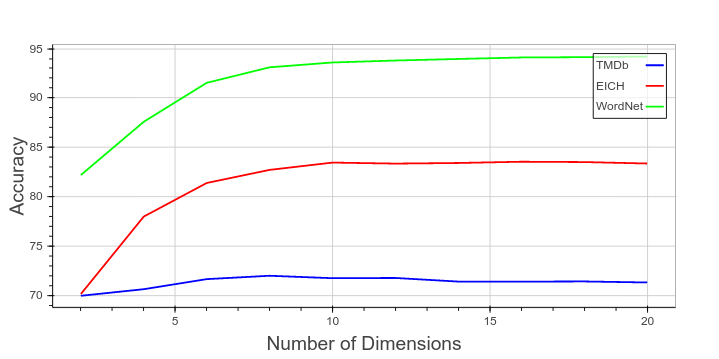}
		}\\ %  ------- End of the first row ----------------------%
		\subfigure[En función del \textit{tamaño de la ventana de selección.}]{%
			\label{ntp-l2}
			\includegraphics[scale=0.35]{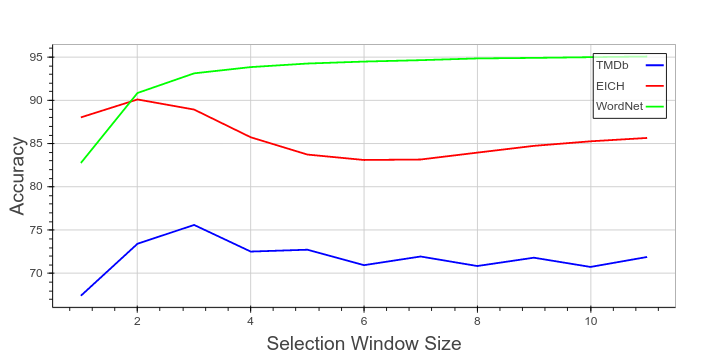}
		}%
	\end{center}
	\caption{%
		Análisis de la inmersión (predicción de tipos de nodo).
	}%
	\label{ntp}
\end{figure}

La Tabla \ref{ntp-optimos} muestra valores de los parámetros en los que se han obtenido resultados de clasificación buenos usando $k$-NN como modelo posterior de aprendizaje (con $k=3$). La Figura \ref{ntp} muestra los resultados de este análisis para los tres datasets considerados, donde se consiguen tasas de predicción superiores al 70\% para todos ellos.

En todos los casos se cumple que el tamaño del conjunto de entrenamiento, $N$, óptimo para realizar la predicción automática de los tipos de nodo es proporcional al número de nodos en el grafo. Tanto EICH como TMDb (para WordNet no conocemos el valor óptimo, porque es creciente en el rango analizado) muestran una reducción en la tasa de predicción a partir del valor óptimo, esto puede deberse a un sobreajuste relacionado con la existencia de nodos de diferentes tipos que poseen la misma etiqueta.

Respecto a la dimensión del espacio vectorial, se pueden observar leves cambios cuando aumentamos $D$ por encima de 10-15, una dimensión relativamente baja, pero es casi imperceptible.

El estudio del parámetro $ws$ muestra que se requieren valores pequeños de este parámetro para obtener buenos resultados en la predicción del tipo asociado a los nodos. Es importante señalar que la mejor predicción no se consigue en ningún caso con $ws=1$, ya que esto supondría que el sistema no necesita recibir pares nodo-contexto como elementos del conjunto de entrenamiento sino que bastaría con mostrarle instancias de las relaciones/propiedades presentes en cada nodo.

\subsubsection{Comparación con otros modelos de predicción}

Una vez fijados los parámetros de la inmersión que proporciona la Tabla \ref{ntp-optimos}, procedemos a comparar la capacidad predictiva con algunos métodos automáticos de clasificación sobre la misma representación. Concretamente, compararemos estos resultados con los obtenidos a través de redes neuronales feedforward y Random Forest.

\begin{figure}[htb]
     \begin{center}
        \subfigure[k-Nearest Neighbor.]{%
            \label{ntp-ns3}
            \includegraphics[scale=0.35]{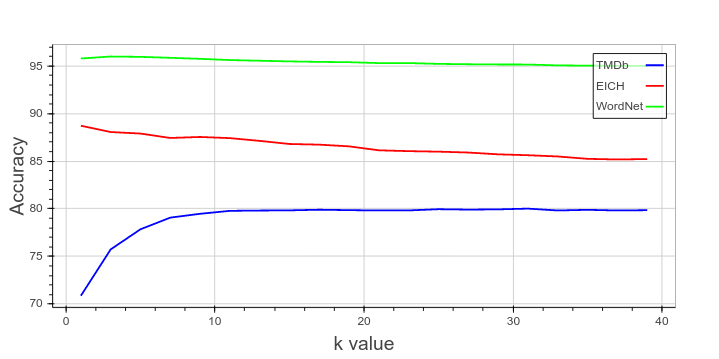}
        }\\ %
        \subfigure[Random Forest.]{%
            \label{ntp-ndim3}
            \includegraphics[scale=0.35]{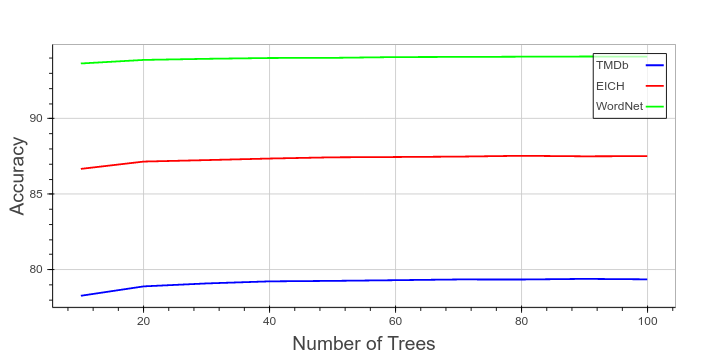}
        }\\ %  ------- End of the first row ----------------------%
        \subfigure[Red Neuronal Artificial.]{%
            \label{ntp-l3}
		    \includegraphics[scale=0.35]{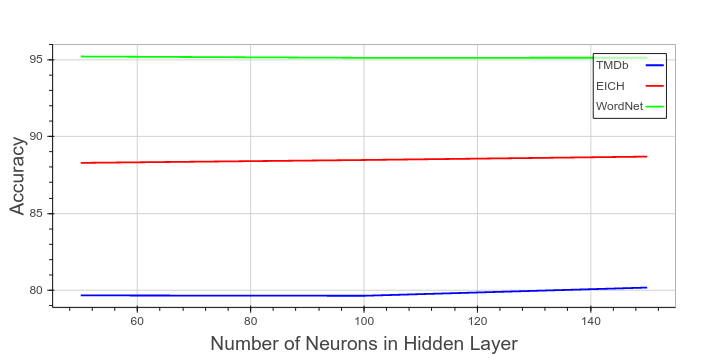}
        }%

    \end{center}
    \caption{%
       Análisis clasificación tipos de nodo por diferentes métodos.
     }%
   \label{ntp-methods}
\end{figure}

En la Figura \ref{ntp-methods} se muestran los resultados obtenidos. La gráfica (a) muestra la variación de los resultados proporcionados por k-NN cuando varía el valor $k$; en (b) se muestran los resultados arrojados por Random Forest cuando se modifica el número de árboles; finalmente, en (c) se muestran los resultados de la red neuronal cuando se modifica el número de neuronas en la capa oculta. En cualquier caso, los valores de los parámetros de estos modelos se mantienen relativamente bajos para la tarea que se está llevando a cabo.

También presentamos las matrices de confusión promediadas tras realizar 10 experimentos utilizando los parámetros óptimos indicados anteriormente: WordNet (Tabla \ref{wordnet_ntcm}), TMDb (Tabla \ref{tmdb_ntcm}), y EICH (Tabla \ref{eich_ntcm}). Estas matrices capturan las similitudes semánticas  entre los tipos de nodo. En EICH, por ejemplo, \texttt{Canton}, \texttt{Parroquia} y \texttt{Provincia} muestran un comportamiento solapado debido a que todos ellos representan información geoespacial altamente correlacionada. En TMDb ocurre algo similar, \texttt{ACTOR} y \texttt{DIRECTOR} aparecen relacionados debido a que, como comentamos, existen numerosos nodos en esta base de datos que tienen ambos tipos.

\begin{table}[tb]
\centering
\caption{Matriz de confusión: Predicción de tipos de nodos (WordNet)}
\label{wordnet_ntcm}
\begin{tabular}{lcccc}
                  & \meg{ adjective} & \meg{ verb}   & \meg{ noun}   & \meg{ adverb} \\ \hline
 \meg{ adjective} & \meg{ 90.01\%}    & 1.75\%         & 8.2\%          & 0.05\%         \\ \hline
 \meg{ verb}      & 0.44\%            & \meg{ 88.36\%} & 11.19\%        & 0.0\%          \\ \hline
 \meg{ noun}      & 0.2\%             & 1.56\%         & \meg{ 98.23\%} & 0.01\%         \\ \hline
 \meg{ adverb}    & 10.16\%           & 1.63\%         & 29.27\%        & \meg{ 58.94\%} \\ \hline
\end{tabular}

\end{table}

\begin{table}[tb]
\centering
\caption{Matriz de confusión: Predicción de tipos de nodos (TMDb)}
\label{tmdb_ntcm}
\begin{tabular}{lccccc}
                 & \meg{ Director} & \meg{ Movie} & \meg{ Genre} & \meg{ Studio} & \meg{ Actor}  \\ \hline
 \meg{ Director} & \meg{ 11.45\%}   & 9.54\%        & 0.02\%        & 1.48\%         & 77.51\%        \\ \hline
 \meg{ Movie}    & 6.51\%           & \meg{ 65.8\%} & 0.02\%        & 0.29\%         & 27.37\%        \\ \hline
 \meg{ Genre}    & 9.66\%           & 33.79\%       & \meg{ 2.07\%} & 3.45\%         & 51.03\%        \\ \hline
 \meg{ Studio}   & 9.66\%           & 8.49\%        & 0.01\%        & \meg{ 1.23\%}  & 80.61\%        \\ \hline
 \meg{ Actor}    & 5.77\%           & 7.87\%        & 0.0\%         & 0.7\%          & \meg{ 85.66\%} \\ \hline

\end{tabular}

\end{table}

\subsection{Predicción de Tipos de Aristas} 

El segundo experimento tiene como objetivo determinar la bondad que presentan las inmersiones en la predicción de los tipos de aristas, que tampoco han sido usados en el proceso de entrenamiento del codificador neuronal. 

\begin{figure}[H]
	\centering
	\includegraphics[scale=0.3]{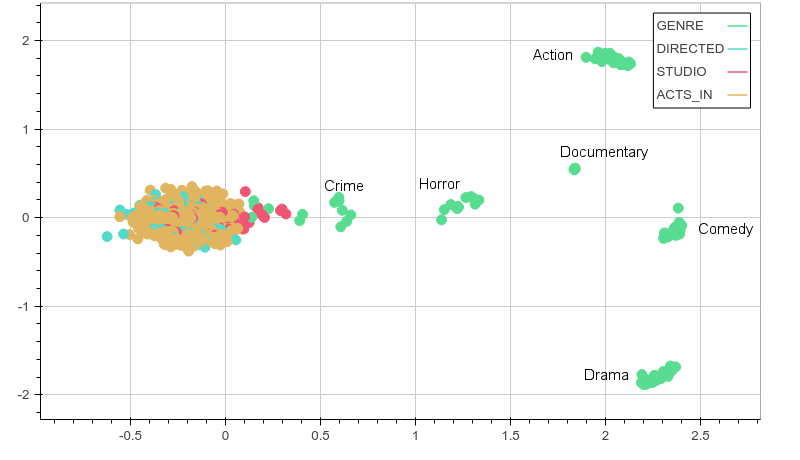}
	\caption{Representación 2D aristas de TMDb.}
	\label{cine-lsample}
\end{figure}

La figura \ref{cine-lsample} muestra una proyección bidimensional (obtenida también aplicando el método MDS) de un conjunto de aristas seleccionadas aleatoriamente del conjunto de datos TMDb. Se puede observar que las aristas de tipo \texttt{Genre} no forman un único cluster, sino un conjunto de clústers periféricos que corresponden a los valores \texttt{Action, Comedy, Drama, Documentary, Horror} y \texttt{Crime}, mostrando un comportamiento semántico diferente al del resto de tipos. Esto podría indicarnos que \texttt{Genre} quizás no forma un tipo único desde el punto de vista semántico, y que su comportamiento refleja información acerca de cierta heterogeneidad que ha sido incluida en las decisiones de diseño al construir la base de datos original. Es aquí donde este tipo de análisis muestra características que lo pueden hacer adecuado para ser usado como normalizador adicional en bases de datos que cubre información semántica y no solo estructural.

\begin{figure}[htb]
     \begin{center}
        \subfigure[En función del tamaño del conjunto de entrenamiento.]{%
            \label{ltp-ns}
            \includegraphics[scale=0.35]{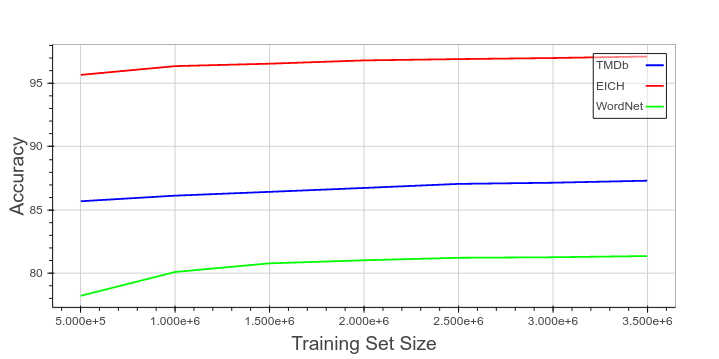}
        }\\ %
        \subfigure[En función del número de dimensiones.]{%
            \label{ltp-ndim}
            \includegraphics[scale=0.35]{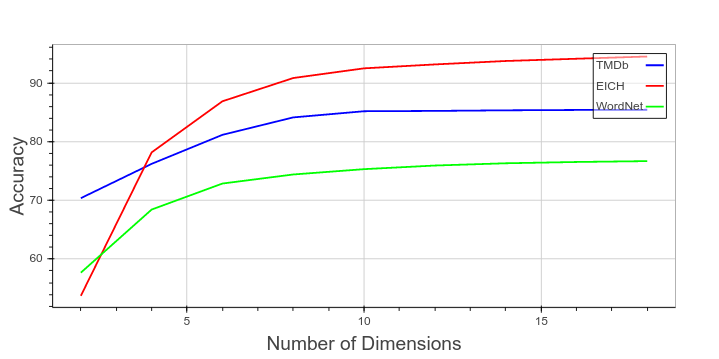}
        } \\ %  ------- End of the first row ----------------------%
        \subfigure[En función del \textit{tamaño de la ventana de selección.}]{%
            \label{ltp-l}
		    \includegraphics[scale=0.35]{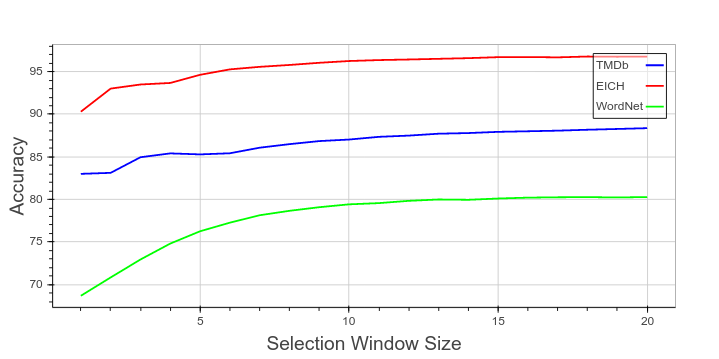}
        }%
    \end{center}
    \caption{%
       Análisis de la inmersión (predicción de tipos de arista).
     }%
   \label{ltp}
\end{figure}

En la Figura \ref{ltp} se muestran los resultados de la eficiencia de k-NN respecto a cambios en los parámetros de la inmersión. Teniendo en cuenta que el porcentaje de acierto está por encima del 80\% en todos los conjuntos de datos estudiados (incluso superando el 95\% en alguno de ellos), podemos concluir que la metodología seguida para la inmersión mantiene las propiedades semánticas también respecto de las aristas.

En general, podemos observar que el tamaño del conjunto de entrenamiento necesario para obtener buenos resultados a la hora de predecir los tipos de las aristas es superior al requerido para realizar una buena predicción de los tipos de nodo para los tres datasets analizados. Además, observando los resultados descubrimos que, a pesar de que WordNet es el grafo con propiedades que mejores resultados ofrecía en la predicción del tipo de los nodos, en el caso de la predicción de aristas se consiguen mejores resultados para EICH, logrando tasas con un valor $\simeq 97\%$ para los parámetros de la inmersión estudiados. 

Al igual que en el caso anterior, las inmersiones requieren una dimensión relativamente baja, con pequeños cambios a partir de $D=15$.

El comportamiento en la predicción de tipos de aristas según $ws$ muestra valores más elevados que los requeridos para la predicción en los tipos de nodos (superior a 20). En cualquier caso, es importante señalar que, de nuevo, la mejor predicción no se consigue en ningún caso con $ws=1$, lo que evidencia que muestrear los contextos locales de un nodo consigue mejores resultados que el que se podría conseguir capturando sólo las relaciones binarias entre los nodos que ofrecen las aristas.

\subsubsection{Comparación con otros modelos de predicción}

\begin{figure}[htb]
     \begin{center}
        \subfigure[k-Nearest Neighbor.]{%
            \label{ntp-ns4}
            \includegraphics[scale=0.35]{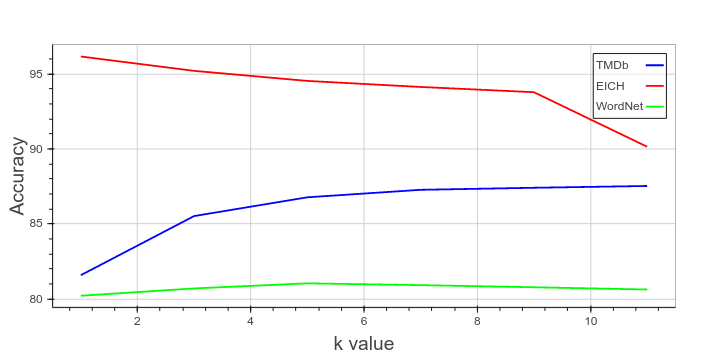}
        }\\%
        \subfigure[Random Forest.]{%
            \label{ntp-ndim4}
            \includegraphics[scale=0.35]{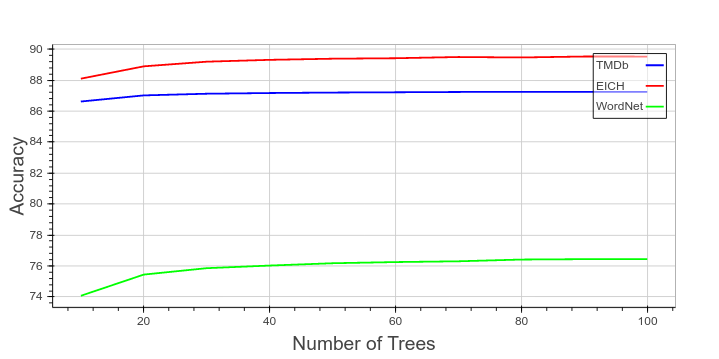}
        }\\ %  ------- End of the first row ----------------------%
        \subfigure[Red Neuronal Artificial.]{%
            \label{ntp-l4}
		    \includegraphics[scale=0.5]{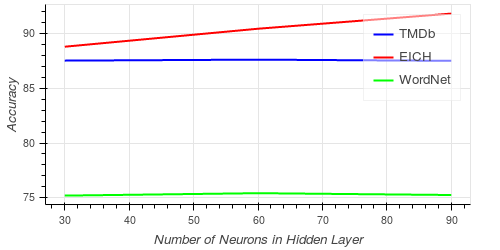}
        }%
    \end{center}
    \caption{%
       Análisis clasificación tipos de arista por diferentes métodos.
     }%
   \label{ltp-methods}
\end{figure}

Siguiendo la misma metodología que para los tipos de nodos, la Figura \ref{ltp-methods} muestra los resultados obtenidos por los tres métodos de clasificación automática utilizados en el apartado anterior. Estas tareas de clasificación se han realizado con inmersiones que hacen uso de los parámetros presentados en la tabla \ref{ntp-optimos}, y se analizan modificando los mismos hiperparámetros de cada modelo concreto.

Las matrices de confusión tras promediar 10 experimentos para cada grafo se muestran en las Tablas \ref{tmdb_ltcm}, \ref{eich_ltcm}, y \ref{wordnet_ltcm}, poniendo en evidencia que las inmersiones capturan similitudes entre diversos tipos de aristas. En el caso de EICH los tipos de aristas relacionados con información geoespacial muestran un comportamiento solapado con las aristas de tipo \texttt{LENGUA}, debido a que existe una correlación entre las lenguas habladas y los territorios en los que se habla. WordNet muestra un comportamiento similar entre tipos de aristas \texttt{hypernym} y \texttt{hyponim}, debido a que tienen un comportamiento semántico similar. En el caso de TMDb, como era de esperar, las aristas de tipo \texttt{DIRECTED} se confunden con las aristas de tipo \texttt{ACTED\_IN} debido al solapamiento entre los tipos de nodo \texttt{ACTOR} y \texttt{DIRECTOR} que intervienen en las mismas.

\begin{table}[tb]
\centering
\caption{Matriz de confusión: Predicción de tipos de aristas (TMDb)}
\label{tmdb_ltcm}
\begin{tabular}{lcccc}

                 & \meg{ GENRE}  & \meg{ DIRECTED} & \meg{ STUDIO} & \meg{ ACTS\_IN} \\ \hline
 \meg{ GENRE}    & \meg{ 99.51\%} & 0.02\%           & 0.21\%         & 0.26\%          \\ \hline
 \meg{ DIRECTED} & 0.01\%         & \meg{ 15.28\%}   & 2.04\%         & 82.67\%         \\ \hline
 \meg{ STUDIO}   & 0.13\%         & 7.22\%           & \meg{ 62.87\%} & 29.79\%         \\ \hline
 \meg{ ACTS\_IN}  & 0.01\%         & 4.75\%           & 0.94\%         & \meg{ 94.3\%}   \\ \hline

\end{tabular}

\end{table}

%\newgeometry{top=0cm,bottom=1cm}

Los resultados experimentales correspondientes a la clasificación automática muestran que, para los conjuntos de datos analizados, la inmersión obtenida con la metodología propuesta conserva la semántica asociada a los tipos de aristas y es capaz de detectar similitudes semánticas entre los tipos de aristas.

Cabe destacar que, debido a que pueden existir aristas de diferentes tipos entre el mismo par de nodos, el resultado en la predicción en el tipo de arista entre dos nodos puede haberse visto afectado. Si estamos tratando de predecir el tipo de una arista entre dos nodos sólo a partir de las posiciones vectoriales de los mismos, y entre ellos existen aristas de diferentes tipos, la solución no es única, ya que cualquiera de estos tipos sería una solución válida. Este hecho no se ha tenido en cuenta a la hora de llevar a cabo los experimentos, por lo que los resultados en la predicción de tipos de aristas suponen un límite inferior y podrían ser mejorados teniendo en cuenta que la respuesta correcta no es única. Sin lugar a dudas, por este método nunca podremos obtener una fiabilidad absoluta acerca de los resultados, y como sistema de predicción puro para aristas presenta limitaciones fundamentales, pero puede ser tenido en cuenta para tareas adicionales de normalización de datos o como método de filtrado para otras operaciones que trabajen sobre la semántica de las aristas.

\subsection{Entity Retrieval}

Para seguir poniendo a prueba la bondad de las inmersiones que conseguimos respecto a la semántica interna de los grafos, y para poner de manifiesto su utilidad en otras tareas de predicción, vamos a evaluar hasta qué punto somos capaces de predecir \textit{relaciones faltantes} haciendo uso de las inmersiones. 

Para ello, consideraremos un subconjunto de aristas, $E' \in E$, que pertenecen al grafo original $G = (V,E,\tau,\mu)$ y que posteriormente eliminaremos, consiguiendo un subgrafo del anterior, $G'=(V,E \setminus E', \tau, \mu)$, sobre el que entrenaremos la inmersión. Posteriormente, trataremos de obtener el nodo destino asociado a cada arista en $E'$ usando únicamente su nodo origen y $\pi(G')$.

Formalmente, dada una arista $e = (s,t) \in E'$ de tipo $\tau(e)$, que ha sido eliminada de manera previa a la inmersión del grafo, trataremos de obtener $t$ a partir de $\pi(G')$, $\tau(e)$ y $s$. Esta tarea de obtener el target de una relación dado el nodo origen y el tipo de la misma es conocida como \textit{Entity Retrieval} \cite{typed}.

Vamos a utilizar el \textit{vector representante} de los tipos para obtener el nodo destino de las relaciones faltantes, que definimos como:

\begin{definition}
Dado un grafo con propiedades $G=(V,E,\tau,\mu)$, el \emph{vector representante}, $\pi(\omega)$, asociado a un tipo de arista $\omega\in \tau(E)$,  es el vector promedio de todos los vectores que representan a aristas de tipo $\omega$. 

Si denotamos $E_\omega= \tau^{-1}(\{\omega\}) =\{e\in E:\ \tau(e)=\omega\}$, entonces:

$$\pi(\omega) = \frac{1}{\#(E_\omega)} \sum_{e\in E_\omega} \pi(e)$$
\end{definition}

A partir de la extensión de $\pi$ que hemos dado para las aristas, si queremos obtener un candidato del destino de una relación $e$ a partir del origen haciendo uso del \textit{vector representante} de la relación y del vector asociado al nodo origen, basta hacer:

$$	\pi(t_e) = \pi(s) + \pi(\tau(e))$$

El vector $\pi(t_e)$ representa la posición a la que \textit{apunta} el \textit{vector representante} de $\tau(e)$ desde el vector que representa el nodo origen $\pi(s)$ de la relación $e$. Una vez obtenido el vector $\pi(t_e)$ podemos obtener un \textit{ranking} para los nodos del grafo, que se puede construir a partir de las distancias a $\pi(t_e)$ de cada vector asociado a los nodos del grafo original, de tal manera que los nodos que más cerca se encuentren del vector $\pi(t_e)$ ocuparán las primeras posiciones de dicho ranking.

En la tabla \ref{ER} se muestran los diez primeros resultados del ranking obtenido tras aplicar \textit{Entity Retrieval} a través del vector representante, $\pi(hypernym)$, de las relaciones de tipo \texttt{hypernym} a diferentes nodos origen del grafo WordNet, los resultados están filtrados de tal manera que sólo se muestran los nodos de tipo \texttt{NOUN}.

\begin{table}[tb]
\setlength{\tabcolsep}{1pt}
\centering 
\caption{Ranking de Entity Retrieval usando la relación \texttt{hypernym}}
\label{ER}
\small
\begin{tabular}{lcccc}
\hline
            & \multicolumn{1}{c}{\textit{foam}} & \multicolumn{1}{c}{\textit{spasm}} & \multicolumn{1}{c}{\textit{justification}} & \multicolumn{1}{c}{\textit{neconservatism}} \\ \hline
\textbf{1}    & hydrazine                                          & ejection & reading                             & pruritus \\ \hline
\textbf{2} & pasteboard                                          & rescue                                     & explanation                             & conservatism                                \\ \hline
\textbf{3}        & silicon dioxide                                          & putting to death                                      & analysis                           & sight                               \\ \hline
\textbf{4}        & humate                                          & sexual activity                                      & proposition                             & hawkishness                                \\ \hline
\textbf{5}        & cellulose ester                                          & behavior modification                                     & religious doctrine                             & coma                                \\ \hline
\textbf{6}        & synthetic substance                                         & disturbance & accusation                               & scientific method                               \\ \hline
\textbf{7}        & silver nitrate                                          & mastectomy                                     & assay                             & autocracy                             \\ \hline
\textbf{8}        & cast iron                                          & sales event                                     & confession                             & judiciousness                                \\ \hline
\textbf{9}        & sulfide                                          & instruction                                      & research                             & reverie                                \\ \hline
\textbf{10}        & antihemorrhagic factor & debasement                            & discouragement  & racism\\ \hline

\end{tabular}
\end{table}

Como nuestra metodología para construir la inmersión se realiza a partir de muestras aleatorias de diferentes \textit{contextos locales} del grafo, es posible que en algunos casos el nodo origen de la relación no haya sido considerado en ningún momento y, por tanto, no podamos construir su representación vectorial. Para que este hecho 
no afecte a los resultados, en estos casos la arista no podrá ser evaluada y no influirá en el resultado experimental obtenido.

Para evaluar la bondad de la inmersión respecto de esta tarea haremos uso de la métrica \textit{Mean Reciprocal Rank}, una métrica habitual en el área de \textit{Information Retrieval}, y que ha sido utilizada en varios estudios de este tipo \cite{typed,learning}.

\begin{definition}
El \emph{Reciprocal Rank} asociado a un resultado concreto, en una lista de posibles respuestas dada una consulta, es el inverso de la posición que ocupa ese resultado en dicha lista. El \emph{Mean Reciprocal Rank} (MRR) es el promedio de los Reciprocal Ranks de una lista de consultas determinada:
$$MRR = \frac{1}{|Q|} \sum_{i=1}^{|Q|} \frac{1}{\text{rank}_i}$$
donde $Q$ representa el conjunto de consultas a evaluar, y $rank_i$ la posición que ocupa en cada ranking la respuesta correcta.
\end{definition}

\begin{figure}[H]
	\centering
	\includegraphics[scale=0.35]{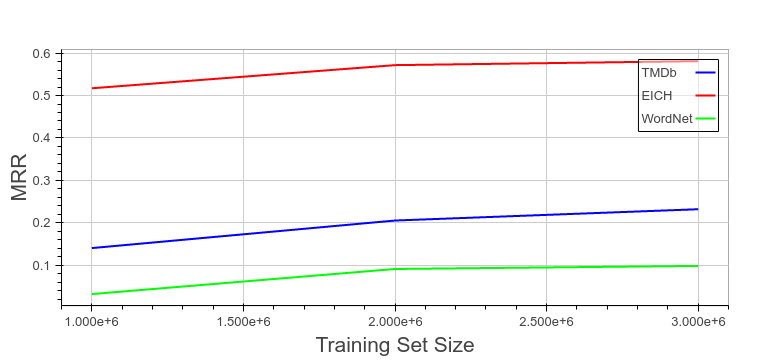}
	\caption{%
		Análisis MRR en \textit{Entity Retrieval}.
	}%
	\label{ntp-ns5}
\end{figure}

En la Figura \ref{ntp-ns5} se muestran los resultados obtenidos usando esta métrica sobre los datasets EICH, TMDb y WordNet en función del tamaño del conjunto de entrenamiento utilizado para realizar la inmersión, y eliminando del ranking aquellos nodos que no son del tipo que indica el nodo destino del tipo de relación evaluada. Como se puede observar en la gráfica, el método propuesto para llevar a cabo este tipo de tareas produce unos excelentes resultados que mejoran cuanto mayor es el conjunto de entrenamiento (en este caso, es un problema de multi-clasificación, por lo que no se pueden esperar resultados que se acerquen al $100\%$).

Este buen comportamiento nos permite obtener ciertas conclusiones sobre la estructura que los diferentes vectores (asociados a nodos y aristas) forman en la nueva representación: si el vector representante sirve para obtener el nodo destino de una arista significa que existe poca desviación entre las aristas del mismo tipo. Por otro lado, los nodos origen y destino del tipo de arista que se usa deben estar lo suficientemente dispersos para que, utilizando el vector representante, se consigan buenos resultados en cuanto a tareas relacionadas con \textit{Entity Retrieval}.

\subsection{Inmersión de caminos tipados}
\label{ctipados}

Por último, y solo a modo de demostración de las posibilidades que abre el tener una buena representación vectorial de los elementos de un grafo con propiedades, presentamos una técnica basada en la inmersión para obtener el nodo destino de un \textit{camino tipado} dado el tipo del camino y el nodo origen del mismo. 

Un camino tipado no es más que la sucesión de tipos en nodos y aristas que corresponde a un camino dentro del grafo (en algunos contextos a estos caminos tipados se les conoce como \emph{traversals}, pero preferimos no usar esta nomenclatura debido al solapamiento que produce con los traversals como métodos de consulta en determinados lenguajes de consulta sobre grafos). Formalmente:

\begin{definition}
Un \emph{camino tipado} de un grafo con propiedades $G=(V,E,\tau,\mu)$ es una sucesión 
$$T=t_1\overset{r_1}{\rightarrow}t_2\overset{r_2}{\rightarrow} \ldots \overset{r_q}{\rightarrow}t_{q+1}$$
donde $t_i \in \tau(V)$ (es un tipo válido para los nodos) y $r_i \in \tau(E)$ (es un tipo válido para las aristas). Denotaremos por $Tp(G)$ el conjunto de posibles caminos tipados de $G$.
\end{definition}

\begin{definition}
Podemos definir la aplicación, $Tp$, que asocia a cada posible camino tipado de $G$ el conjunto de caminos que verifican el patrón de tipos especificado por él, tal que si $T=t_1\overset{r_1}{\rightarrow}t_2\overset{r_2}{\rightarrow} \ldots \overset{r_q}{\rightarrow}t_{q+1}$, entonces para cada camino $\rho\in Tp(T)$ se verifica que $\tau(sop_V(\rho))=(t_1,\dots,t_{q+1})$ y $\tau(sop_E(\rho))=(r_1\dots,r_q )$ (donde $sop_V(\rho)$ representa la secuencia ordenada de nodos en $\rho$, y $sop_E(\rho)$ la secuencia de aristas).
\end{definition}

Nuestro objetivo es obtener el nodo destino de un camino existente dado el nodo origen del mismo y el camino tipado que verifica. En este caso no eliminamos los caminos antes de realizar la inmersión, pues no tratamos de hacer predicción sino de ofrecer un nuevo mecanismo para la obtención (o, al menos, su estimación) del nodo destino de un camino que permita mejorar los tiempos que requieren este tipo de consultas, ya que en los sistemas actuales tienen un coste computacional muy elevado. 

Para ello, definimos el vector representante de un camino de forma similar a como lo hicimos en la tarea anterior (que realmente se puede considerar un caso particular de camino tipado para caminos de longitud 1).

\begin{definition}
El \emph{vector representante} de un camino, $\conecta{n_1}{\rho}{n_k}$, en un grafo con propiedades, es el vector que separa la representación vectorial del nodo origen del camino, $\pi(n_1)$,  y la representación vectorial del nodo destino del mismo, $\pi(n_k)$. Es decir:
$$\pi(\rho)=\overrightarrow{\pi(n_1) \pi(n_k)} = \pi(n_k)-\pi(n_1)$$

El \emph{vector representante} asociado a un camino tipado, $T$, es el vector promedio de todos los vectores que verifican el patrón de tipos especificado por $T$, es decir:
$$\pi(T)=\frac{1}{|Tp(T)|}\sum_{\rho\in Tp(T)}\pi(\rho)$$
\end{definition}

Como ocurría con las aristas, es posible que en algunos casos el nodo origen del camino no haya sido tomado en la muestra de la inmersión y, por tanto, no exista su representación vectorial. En estos casos, dicho camino no podrá ser evaluado y no influirá en el resultado experimental obtenido.

A partir de esta definición se han realizado experimentos para evaluar la tarea de obtener el nodo destino de un camino dado el nodo origen y el vector representante del camino tipado asociado. Para ello, hemos filtrado los nodos destino según el tipo indicado por el último elemento de la secuencia que define el camino tipado y hemos utilizando de nuevo la métrica MRR presentada en el apartado anterior. Los experimentos han sido realizados sobre el dataset EICH debido a que presenta una estructura más compleja en sus tipos que el resto de datasets y permite la construcción de caminos tipados más complejos.

\begin{figure}[H]
	\centering
	\includegraphics[scale=0.5]{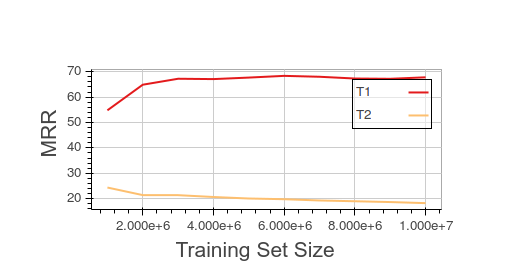}
	\caption{Análisis MRR en caminos tipados.}
	\label{traversals_EICH}
\end{figure}

En  la figura \ref{traversals_EICH} se muestran los resultados obtenidos en los experimentos haciendo uso de los siguientes caminos tipados (los tipos de los nodos se representan en minúsculas, y los de las aristas se omiten porque representan el único tipo de arista que permite el esquema mostrado en la figura \ref{eich-schema}):

\begin{enumerate}

\item $T1=(Inmaterial\overset{r_1}{\rightarrow} DetSubambito\overset{r_2}{\rightarrow}Subambito\overset{r_3}{\rightarrow}Ambito)$

Está asociado a caminos de longitud 3 y contiene información sobre a qué Ambito (existen 5 ámbitos diferentes en EICH) pertenece cada elemento del patrimonio inmaterial almacenado en el grafo. 

\item $T2=(Inmaterial\overset{r_1}{\rightarrow}Parroquia\overset{r_2}{\rightarrow}Canton\overset{r_3}{\rightarrow}Provincia)$

Está asociado a caminos de longitud 3 y contiene información sobre a qué Provincia (existen 24 provincias diferentes en EICH) pertenece cada elemento del patrimonio inmaterial almacenado en el grafo.
\end{enumerate}

En los resultados se aprecia el buen desempeño de la tarea propuesta para caminos tipados de tipo $T1$, cercana al $70\%$, para un conjunto de entrenamiento de tamaño superior a 3 millones. En el caso del camino tipado $T2$, sin embargo, su resultado tiende a empeorar cuando aumentamos el tamaño del conjunto de entrenamiento por encima de 1 millón, llegando a estar por debajo del $20\%$. En cualquier caso, el problema asociado a $T_2$ es considerablemente más complejo, ya que hay más de 20 provincias, frente a los 5 posibles ámbitos para el primer caso, y además pudimos ver anteriormente que se obtenía una alta confusión entre los tipos de nodos involucrados. 

Aunque, por supuesto, harían falta más pruebas para validar esta metodología, esta aplicación muestra que un sistema como éste podría utilizarse para aproximar el resultado de consultas \textit{a larga distancia} en bases de datos, que son especialmente ineficientes en el caso de los sistemas clásicos de persistencia, por lo que la inmersión se presenta como una alternativa interesante que, aunque reduciendo la fiabilidad del resultado, permite  agilizar enormemente la carga computacional requerida en algunas tareas relacionadas.

\section{Conclusiones y Trabajo Futuro}

El objetivo de este trabajo ha sido el de ofrecer la posibilidad de llevar a cabo tareas de aprendizaje automático relacional a través de algoritmos tradicionales haciendo una selección automática de atributos (\emph{feature extraction}) a través de inmersiones vectoriales, manteniendo las estructuras semánticas mediante la generación de un conjunto de entrenamiento adecuado. De esta forma buscamos analizar qué opciones ofrecen los algoritmos tradicionales cuando deseamos no perder las estructuras enriquecidas propias de la información relacional. 

Si existe un elemento (un subgrafo) que está inmerso en una base de datos (un grafo generalizado, o un grafo con propiedades, en nuestro contexto) la tarea de construir atributos para el aprendizaje a partir de las relaciones que presenta en la estructura global puede ser muy complicada. La aproximación que se presenta en este trabajo pasa por construir una representación vectorial de cada elemento en el sistema a partir de un muestreo de la información presente en la red. De esta manera evitamos, por un lado, el trabajo manual de selección de los atributos a tener en cuenta y, por otro, conseguimos que el algoritmo de aprendizaje a utilizar se alimente de una representación obtenida a partir de información global disponible.

En comparación con otras tareas de aprendizaje automático, hay pocos trabajos que hayan utilizado codificadores neuronales para realizar inmersiones de grafos con propiedades, o estructuras similares, en espacios vectoriales. Nuestra metodología ha buscado usar arquitecturas simples para obtener representaciones vectoriales que mantienen las características semánticas y topológicas del grafo original. Además, se ha demostrado experimentalmente que con las inmersiones obtenidas se pueden obtener conexiones semánticas que no aparecen explícitamente en el grafo original (debido a incompletitud en los datos almacenados, o a incoherencias en los mismos), o incluso ayudar en la optimización de algunas tareas complejas de consulta en bases de datos.

Hemos comprobado que las características geométricas de las estructuras formadas por las inmersiones de nodos y aristas en el nuevo espacio vectorial pueden ayudar a asignar tipos o propiedades faltantes a los elementos del grafo original (usando medidas relacionadas con distancia, linealidad, o agrupación, entre otras), o pueden incluso ayudar a identificar nuevas relaciones entre elementos que no están presentes explícitamente. Esta funcionalidad puede ser de gran utilidad en procesos que trabajan con grandes conjuntos de datos relacionales, donde la incompletitud de los datos es inherente al problema.

Además, y por encima de la competitividad que ofrece esta metodología frente a otras existentes, como se ha observado a partir de las pruebas de evaluación, el rendimiento y la precisión de las tareas de aprendizaje automático sobre estas representaciones vectoriales pueden proporcionar información sobre la estructura semántica del conjunto de datos en sí, y no sólo sobre los algoritmos en uso. Por ejemplo, la confusión de algunos nodos / aristas en tareas de clasificación puede darnos información sobre la necesidad de realizar un ajuste en el esquema de datos para reflejar las características semánticas correctamente. Un informe detallado sobre cómo los diferentes tipos, propiedades, y clusters se superponen y confunden en la inmersión sería de utilidad para tomar decisiones relacionadas con la normalización de los esquemas de datos, algo de lo que carecen casi todas las propuestas actuales de análisis y que es prácticamente inexistente en los sistemas de datos noSQL. 

Es evidente que el tamaño del conjunto de entrenamiento y de la ventana de selección influyen positivamente en la capacidad de aplicación de la inmersión resultante, pero estas influencias deben ser estudiadas a mayor profundidad, ya que pueden arrojar claves para la automatización de los parámetros de la inmersión.

Además, en este trabajo se ha explorado cómo las estructuras vectoriales pueden usarse para recuperar información de grafos con propiedades, como muestran los experimentos de \textit{Entity Retrieval} y de \emph{caminos tipados}. Es probable que buscar estructuras complejas en el espacio proyectado sea más sencillo que en el espacio original. De hecho, el uso de una segunda capa de modelos de aprendizaje tras la codificación neuronal puede mejorar los resultados de varias tareas relacionadas con la recuperación de información en grafos semánticos. Los resultados en este trabajo muestran que esta es una línea de investigación que vale la pena ser considerada. A pesar de que no se han llevado a cabo suficientes experimentos en cuanto a consultas a larga distancia a través de los vectores representantes en el nuevo espacio, los resultados obtenidos muestran que los tiempos de consulta pueden ser reducidos dramáticamente sacrificando la optimalidad. Este tipo de consultas son muy costosas en las bases de datos, y a pesar de que las bases de datos en grafo han ayudado a reducir su coste computacional siguen presentando grandes problemas de eficiencia cuando el camino de búsqueda tiene más de 3 aristas.

Frente a otras aproximaciones en la misma dirección, este trabajo presenta la novedad de trabajar con contextos semánticos más generales, y no solo con caminos aleatorios, que suponen una linealización de la estructura del grafo original. Pero estas no son las únicas opciones para llevar a cabo codificaciones de grafos con propiedades por medio de redes neuronales. Podemos conseguir codificaciones vectoriales de grafos con propiedades haciendo uso de autocodificadores neuronales, de tal manera que el codificador neuronal aprenderá la función identidad para los elementos del grafo, desligando la codificación de la función que relaciona a los elementos con su contexto.

Con este trabajo hemos dado un marco inicial para realizar tareas de aprendizaje automático a partir de grafos con propiedades en las que se tiene en cuenta información del grafo completo para codificar cada elemento. Esta nueva representación de los grafos con propiedades permite trabajar con datos relacionales almacenados en casi cualquier sistema de persistencia de manera vectorial, aprovechando la potencia que tienen actualmente las CPUs y las GPUs para trabajar con este tipo de estructuras.

Debe señalarse que durante la revisión de este documento se han publicado nuevas herramientas basadas en las arquitecturas de \textit{Word2Vec} que optimizan el proceso de aprendizaje de semánticas latentes a partir de lenguaje natural \cite{bojanowski2016enriching}. A pesar de la probable mejora que estas herramientas supondrían en nuestra metodología, hemos decidido no tenerlas en cuenta ya que no modifican la esencia de nuestra propuesta, aunque sí aligeraría, posiblemente, la carga de cálculo asociada a los experimentos realizados.

Las mejoras en eficiencia en consultas a larga distancia planteadas en el apartado \ref{ctipados} merecen ser evaluadas a mayor profundidad y comparadas con otros métodos similares. Algunos resultados relacionados con el análisis semántico de grafos con propiedades no han sido llevados a cabo en profundidad y no se han presentado en este artículo aunque se prevé sean presentados en trabajos posteriores. Opciones como muestrear el contexto de las aristas, realizar una inmersión de las mismas y a partir de ésta inferir una inmersión para los nodos no han sido tenidas en cuenta y pueden ofrecer resultados interesantes.

Durante la concepción, implementación y experimentación de este trabajo han ido abriéndose nuevas vías que pueden ser consideradas para analizar las características de las inmersiones obtenidas. 

Una primera consideración a tener en cuenta está relacionada con la manera de construir el conjunto de entrenamiento que es consumido por el codificador neuronal. En los experimentos realizados la construcción del conjunto de entrenamiento ha sido totalmente aleatoria, es decir, todos los nodos tienen la misma probabilidad de ser muestreados, al igual que todas sus propiedades y vecinos. Ésta puede no ser la manera más adecuada dependiendo del tipo de actividad que se desee realizar con la inmersión resultante. Por ejemplo, puede ser favorable construir el conjunto de entrenamiento de manera que aquellos nodos que posean una mayor riqueza semántica tengan más probabilidad de entrar en el conjunto de entrenamiento, lo que puede contribuir a que regiones inicialmente menos probables de ser consideradas compensen este hecho.

Otra línea a tener en cuenta es la de construir una red neuronal que trabaje con los contextos de un elemento como entrada (en formato one-hot) y aprenda a devolver una propiedad determinada de éste como salida de la misma, es decir, conectar el clasificador/regresor neuronal directamente con el codificador, para de esta forma aprender la codificación adecuada y la clasificación/regresión a partir de ésta de manera simultánea. De igual forma, sería interesante pensar en codificadores neuronales que hacen uso de redes neuronales recurrentes para poder analizar el comportamiento de información relacional dinámica, un terreno prácticamente inexplorado en la actualidad.

También cabe destacar que queda abierta la posibilidad de trabajar con propiedades continuas en nodos y aristas, una característica no presente en los datasets utilizados, pero que debe ser considerada para ampliar la capacidad de la metodología presentada. En este caso hay mecanismos directos para incluir la presencia de propiedades continuas, queda como trabajo comenzar probando con estos mecanismos más evidentes y medir posteriormente hasta qué punto se pueden tener en cuenta otras aproximaciones.

Del mismo modo, sería interesante pensar en codificadores neuronales que hacen uso de redes neuronales recurrentes para analizar el comportamiento de información relacional dinámica, un área prácticamente inexplorada hoy en día.

\section*{Agradecimientos}

Agradecemos al Instituto Nacional de Patrimonio Cultural (INPC) del Ecuador por la información relacionada con el Patrimonio Cultural Inmaterial del Ecuador (EICH). Este trabajo ha sido apoyado parcialmente por el Proyecto de Excelencia TIC-6064 de la Junta de Andalucía (España), por el proyecto TIN2013-41086-P del Ministerio Español de Economía y Competitividad (cofinanciado con fondos FEDER) y por el Departamento de Investigación y Postgrado de la Universidad Central del Ecuador.

\bibliographystyle{plain}	% or "siam", or "alpha", etc.
\bibliography{biblio}

\begin{sidewaystable}
	\setlength{\tabcolsep}{1pt}
	
	\tiny
	\caption{Matriz de confusión: Predicción de tipos de aristas (WordNet)}
	\label{wordnet_ltcm}
	\begin{tabular}{lcccccccccccc}
		
		& \meg{ hyper} & \meg{ dom\_reg} & \meg{ part\_mero} & \meg{ dom\_cat} & \meg{ part\_holo} & \meg{ dom\_usage} & \meg{ also}  & \meg{ memb\_mero} & \meg{ inst\_hypo} & \meg{ dom\_memb\_usage} & \meg{ memb\_holo} & \meg{ hypo} \\ \hline
		\meg{ hyper}            & \meg{ 83.67\%}   & 0.01\%                & 1.14\%               & 0.76\%                  & 0.43\%               & 0.02\%               & 0.56\%        & 1.96\%                 & 0.13\%                   & 0.02\%                      & 2.77\%                 & 8.52\%          \\ \hline
		\meg{ dom\_reg}       & 2.08\%           & \meg{ 65.58\%}        & 25.0\%               & 0.33\%                  & 0.93\%               & 0.0\%                & 0.21\%        & 3.27\%                 & 0.24\%                   & 0.0\%                       & 1.12\%                 & 1.25\%          \\ \hline
		\meg{ part\_mero}        & 29.21\%          & 0.74\%                & \meg{ 44.47\%}       & 1.1\%                   & 3.54\%               & 0.01\%               & 0.79\%        & 4.12\%                 & 0.34\%                   & 0.0\%                       & 4.47\%                 & 11.2\%          \\ \hline
		\meg{ dom\_cat}     & 14.88\%          & 0.02\%                & 1.19\%               & \meg{ 78.91\%}          & 0.11\%               & 0.01\%               & 0.44\%        & 0.18\%                 & 0.04\%                   & 0.0\%                       & 0.25\%                 & 3.97\%          \\ \hline
		\meg{ part\_holo}        & 15.53\%          & 0.09\%                & 3.28\%               & 0.34\%                  & \meg{ 45.36\%}       & 0.01\%               & 0.78\%        & 2.25\%                 & 1.24\%                   & 0.0\%                       & 6.67\%                 & 24.45\%         \\ \hline
		\meg{ dom\_usage}        & 4.28\%           & 0.0\%                 & 0.06\%               & 0.39\%                  & 0.03\%               & \meg{ 93.41\%}       & 0.99\%        & 0.06\%                 & 0.0\%                    & 0.0\%                       & 0.11\%                 & 0.68\%          \\ \hline
		\meg{ also}                & 8.0\%            & 0.0\%                 & 0.38\%               & 0.43\%                  & 0.19\%               & 0.03\%               & \meg{ 78.0\%} & 1.92\%                 & 0.07\%                   & 0.01\%                      & 4.64\%                 & 6.35\%          \\ \hline
		\meg{ memb\_mero}      & 11.7\%           & 0.18\%                & 1.76\%               & 0.2\%                   & 0.77\%               & 0.0\%                & 0.93\%        & \meg{ 50.72\%}         & 0.21\%                   & 0.0\%                       & 24.67\%                & 8.87\%          \\ \hline
		\meg{ inst\_hypo}    & 2.47\%           & 0.06\%                & 0.87\%               & 0.05\%                  & 1.81\%               & 0.0\%                & 0.2\%         & 0.53\%                 & \meg{ 80.42\%}           & 0.0\%                       & 1.56\%                 & 12.02\%         \\ \hline
		\meg{ dom\_memb\_usage} & 1.1\%            & 0.0\%                 & 0.03\%               & 0.03\%                  & 0.14\%               & 0.0\%                & 1.13\%        & 0.09\%                 & 0.06\%                   & \meg{ 93.62\%}              & 0.09\%                 & 3.73\%          \\ \hline
		\meg{ memb\_holo}      & 11.24\%          & 0.05\%                & 0.67\%               & 0.09\%                  & 1.83\%               & 0.0\%                & 0.91\%        & 14.01\%                & 0.21\%                   & 0.0\%                       & \meg{ 62.37\%}         & 8.61\%          \\ \hline
		\meg{ hypo}             & 10.52\%          & 0.01\%                & 0.42\%               & 0.31\%                  & 1.18\%               & 0.02\%               & 0.57\%        & 1.23\%                 & 1.18\%                   & 0.01\%                      & 3.67\%                 & \meg{ 80.88\%}  \\ \hline
		
	\end{tabular}
	
	\tiny
	\caption{Matriz de confusión: Predicción de tipos de nodos (EICH)}
	\label{eich_ntcm}
	\begin{tabular}{lccccccccccc}
		
		& \meg{ Subambito} & \meg{ Provincia} & \meg{ Comunidad} & \meg{ Anexos} & \meg{ Herramienta} & \meg{ Canton} & \meg{ Lengua} & \meg{ Inmaterial} & \meg{ Ambito} & \meg{ Parroquia} & \meg{ DetalleSubambito} \\ \hline
		\meg{ Subambito}        & \meg{ 14.53\%}    & 0.0\%             & 0.0\%             & 2.56\%         & 0.0\%               & 0.0\%          & 0.0\%          & 47.86\%            & 0.0\%          & 0.85\%            & 34.19\%                  \\ \hline
		\meg{ Provincia}        & 0.0\%             & \meg{ 7.14\%}     & 2.04\%            & 0.0\%          & 0.0\%               & 69.39\%        & 5.1\%          & 1.02\%             & 0.0\%          & 15.31\%           & 0.0\%                    \\ \hline
		\meg{ Comunidad}        & 0.0\%             & 0.0\%             & \meg{ 0.0\%}      & 7.91\%         & 0.0\%               & 1.44\%         & 0.0\%          & 25.18\%            & 0.0\%          & 65.47\%           & 0.0\%                    \\ \hline
		\meg{ Anexos}           & 0.0\%             & 0.0\%             & 0.0\%             & \meg{ 81.16\%} & 0.0\%               & 0.0\%          & 0.0\%          & 18.63\%            & 0.0\%          & 0.21\%            & 0.0\%                    \\ \hline
		\meg{ Herramienta}      & 0.0\%             & 0.0\%             & 0.0\%             & 0.68\%         & \meg{ 36.99\%}      & 0.0\%          & 0.0\%          & 62.33\%            & 0.0\%          & 0.0\%             & 0.0\%                    \\ \hline
		\meg{ Canton}           & 0.0\%             & 3.74\%            & 0.1\%             & 5.27\%         & 0.0\%               & \meg{ 12.18\%} & 0.0\%          & 24.26\%            & 0.0\%          & 54.27\%           & 0.19\%                    \\ \hline
		\meg{ Lengua}           & 0.0\%             & 0.0\%             & 0.0\%             & 6.25\%         & 0.0\%               & 0.0\%          & \meg{ 0.0\%}   & 21.25\%            & 0.0\%          & 72.5\%            & 0.0\%                    \\ \hline
		\meg{ Inmaterial}       & 0.01\%            & 0.0\%             & 0.0\%             & 9.44\%         & 0.19\%              & 0.0\%          & 0.01\%         & \meg{ 89.77\%}     & 0.0\%          & 0.56\%            & 0.01\%                   \\ \hline
		\meg{ Ambito}           & 44.0\%            & 0.0\%             & 0.0\%             & 0.0\%          & 0.0\%               & 0.0\%          & 0.0\%          & 28.0\%             & \meg{ 0.0\%}   & 0.0\%             & 28.0\%                   \\ \hline
		\meg{ Parroquia}        & 0.02\%            & 0.42\%            & 0.08\%            & 2.34\%         & 0.02\%              & 2.11\%         & 0.35\%         & 29.67\%            & 0.0\%          & \meg{ 64.82\%}    & 0.18\%                   \\ \hline
		\meg{ DetalleSubambito} & 1.63\%            & 0.0\%             & 0.0\%             & 5.42\%         & 0.0\%               & 0.18\%         & 0.0\%          & 49.37\%            & 0.0\%          & 4.52\%            & \meg{ 38.88\%}           \\ \hline
		
	\end{tabular}

	\tiny
	\caption{Matriz de confusión: Predicción de tipos de aristas (EICH)}
	\label{eich_ltcm}
	\begin{tabular}{lcccccccccc}
		
		& \meg{ CANTON\_L} & \meg{ COM} & \meg{ LOC} & \meg{ HERRAM} & \meg{ PARROQ\_L} & \meg{ ANEXO}  & \meg{ SUBAMBITO\_P} & \meg{ LENGUA} & \meg{ AMBITO} & \meg{ DSUBAMBITO\_P} \\ \hline
		\meg{ CANTON\_L}           & \meg{ 25.05\%}     & 1.94\%            & 12.62\%              & 0.0\%               & 26.8\%                & 6.99\%         & 0.0\%                       & 22.91\%        & 3.69\%         & 0.0\%                        \\ \hline
		\meg{ COM}            & 0.0\%              & \meg{ 97.92\%}    & 0.18\%               & 0.0\%               & 0.09\%                & 0.37\%         & 0.0\%                       & 1.4\%          & 0.04\%         & 0.0\%                        \\ \hline
		\meg{ LOC}         & 0.0\%              & 0.12\%            & \meg{ 96.77\%}       & 0.04\%              & 1.08\%                & 1.51\%         & 0.0\%                       & 0.33\%         & 0.15\%         & 0.0\%                        \\ \hline
		\meg{ HERRAM}          & 0.0\%              & 0.0\%             & 1.49\%               & \meg{ 44.03\%}      & 2.24\%                & 48.51\%        & 0.0\%                       & 3.73\%         & 0.0\%          & 0.0\%                        \\ \hline
		\meg{ PARROQ\_L}        & 0.89\%             & 0.99\%            & 13.95\%              & 0.02\%              & \meg{ 59.11\%}        & 3.34\%         & 0.0\%                       & 19.2\%         & 2.45\%         & 0.05\%                       \\ \hline
		\meg{ ANEXO}                & 0.14\%             & 0.14\%            & 0.49\%               & 0.02\%              & 2.46\%                & \meg{ 95.87\%} & 0.0\%                       & 0.73\%         & 0.11\%         & 0.05\%                       \\ \hline
		\meg{ SUBAMBITO\_P}  & 0.81\%             & 0.0\%             & 10.57\%              & 0.0\%               & 8.13\%                & 9.76\%         & \meg{ 30.89\%}              & 14.63\%        & 5.69\%         & 19.51\%                      \\ \hline
		\meg{ LENGUA}               & 0.0\%              & 1.06\%            & 0.09\%               & 0.0\%               & 0.01\%                & 0.39\%         & 0.0\%                       & \meg{ 98.34\%} & 0.1\%          & 0.0\%                        \\ \hline
		\meg{ AMBITO}               & 0.01\%             & 0.04\%            & 0.28\%               & 0.01\%              & 0.04\%                & 1.9\%          & 0.0\%                       & 2.38\%         & \meg{ 95.33\%} & 0.02\%                       \\ \hline
		\meg{ DSUBAMBITO\_P} & 0.18\%             & 0.18\%            & 1.96\%               & 0.0\%               & 2.67\%                & 3.21\%         & 0.89\%                      & 22.46\%        & 7.66\%         & \meg{ 60.78\%}               \\ \hline
		
	\end{tabular}

\end{sidewaystable}